%% file: main.tex
\documentclass[10pt,twocolumn,letterpaper]{article}

\usepackage[pagenumbers]{cvpr}      %

\input{preamble}

\usepackage{array}
\usepackage{placeins}
\usepackage{tikz}
\usepackage{pgfplots}
\pgfplotsset{compat=1.17}
\definecolor{xlayout}{HTML}{0077BB}
\definecolor{xtiming}{HTML}{EE3377}
\definecolor{xrelay}{HTML}{33BBEE}
\definecolor{xswap}{HTML}{EE7733}
\definecolor{xord}{HTML}{8C8C8C}
\definecolor{xlr}{HTML}{4D4D4D}
\definecolor{xreweight}{HTML}{009988}
\usetikzlibrary{arrows.meta,calc,positioning}
\newcommand{\synclane}{synchrony}
\newcommand{\Synclane}{Synchrony}
\newcommand{\synchead}{Sync}
\renewcommand{\dbltopfraction}{0.85}
\renewcommand{\textfraction}{0.10}

\definecolor{cvprblue}{rgb}{0.21,0.49,0.74}
\usepackage[pagebackref,breaklinks,colorlinks,allcolors=cvprblue]{hyperref}

\def\paperID{11368} %
\def\confName{CVPR}
\def\confYear{2027}

\title{CrossSwitch: Can Video Generators Combine\\Evidence Split Across Videos?}

\author{Sara Ghazanfari \qquad Farshad Khorrami \qquad Prashanth Krishnamurthy \qquad Siddharth Garg\\
New York University}

\makeatletter\newcommand{\teasercaptype}{\def\@captype{figure}}\newcommand{\settablecaptype}{\def\@captype{table}}\makeatother
\begin{document}
\raggedbottom
\twocolumn[{%
\renewcommand\twocolumn[1][]{#1}%
\maketitle

\begin{center}
  \small\teasercaptype
  \input{figures/teaser}\\[6pt]
  \input{figures/teaser_mechanism}
  \caption{\textbf{Cross-video evidence integration and a noisy-input shortcut.} Top: Video~1 reveals the cubes' yellow and cyan patches; Video~2 identifies the actor. The target must show yellow in A and cyan in B. After 10,000 updates at 256 pixels, ordinary training generates yellow in both, while noisy-input swap switches correctly. This held-out \synclane{} pair was selected for this contrast (Appendix~\ref{app:qual}); random pairs appear in Figure~\ref{fig:randomgrid}. Bottom: the noisy input $y$ retains the patch at $\sigma=1$, where swapping begins (dashed outline: patch). Ordinary training can copy it; noisy-input swap instead constructs $y$ from the Video-1 sibling.}
  \label{fig:teaser}
\end{center}
\vspace{0.6em}
}]
\begin{abstract}
CrossSwitch tests cross-video reasoning in generation: producing a view whose attributes require evidence split across two input videos. Existing generation benchmarks do not directly test this dependence. In our controlled task, one video reveals two cubes' colored patches, another identifies the acting cube, and the output must show the actor's correct patch. Paired interventions test correspondence through shared positions (\emph{layout}) or simultaneous observations (\emph{\synclane}). A two-video adaptation of AnyView learns \synclane{} late under ordinary training. A diagnostic probe shows that diffusion and flow-matching models copy an incorrect answer planted in the noisy input despite conflicting videos. The task then lets us test answer removal through \emph{noisy-input swap}, midpoint erasure and recoloring. Combined with Min-SNR, noisy-input swap reaches 80.3\% paired accuracy after 10,000 updates on fresh scenes at 256 pixels, versus 58.9\% for a learning-rate-tuned baseline. Gains hold at a shared learning rate, where midpoint and recoloring perform similarly with Min-SNR. Longer training narrows the swap's advantage. Two blind raters support the combination, but human checks reveal method-dependent scorer errors. Noisy-input swap also improves a real-video test requiring both inputs. CrossSwitch thus enables measurement and diagnosis of cross-video reasoning in generation; answer removal and loss reweighting speed learning by the automatic score, while copying's causal role remains unresolved.

\end{abstract}

\section{Introduction}
\label{sec:intro}

CrossSwitch tests cross-video reasoning in generation: combining evidence split across two videos to produce an output. One camera reveals two look-alike cubes' colored patches; another hides the patches but shows which cube acts. Generating the actor's correct patch in a third view requires matching across videos (Figure~\ref{fig:teaser}); neither input alone suffices. This \emph{cross-video correspondence} is needed in ego--exo recordings, robot cameras and multi-camera world models, but natural-scene shortcuts obscure whether both videos are used.

Existing generation benchmarks do not directly test this dependence. LiveStre4m~\cite{quesado2026livestre4m}, NoPo4D~\cite{balice2026nopo4d} and AnyView~\cite{vanhoorick2026anyview} combine or extend views; cross-video benchmarks score answers to questions~\cite{ghazanfari2026syncr,ghazanfari2026syncrworkshop,chung2026mvvbench}. CrossSwitch changes one input while keeping the other byte-identical and tests whether the generated attribute changes correctly. Cubes correspond through shared positions (\emph{layout}) or simultaneous observations (\emph{\synclane}). Both outputs in a pair must show their correct, different patches. Crossing interventions establishes that neither source alone suffices; shortcut checks do not identify a unique matching strategy (Section~\ref{sec:task}).

On this benchmark, \synclane{} improves late in ordinary training. After 5,000 fine-tuning updates at 576 pixels, a two-video adaptation of AnyView switches correctly on only 18\% of \synclane{} pairs. Its most common failure ignores the patch evidence in Video~1. Longer training improves accuracy, including on scenes balanced to defeat two classifiers that use positions without temporal matching (Section~\ref{sec:speed}).

The task also exposes a possible shortcut. The denoiser receives a \emph{noisy input} $y=z+\sigma\epsilon$: $z$ is the clean target latent, $\epsilon$ standard Gaussian noise, and $\sigma$ its scale. Our \emph{copy probe} plants an incorrect patch in $y$ while preserving the videos. Ordinary models follow this answer on training and held-out scenes, in flow matching, and in released AnyView on real cars and people. Reading the answer from $y$ can lower loss without consulting the videos. Related shortcuts occur in segmentation~\cite{chen2023pix2seqd,lai2023denoising}; copying's role in slow learning remains unresolved.

The diagnosis motivates comparing answer removal through \emph{noisy-input swap}, \emph{midpoint erasure} and \emph{sibling-free recoloring}. These sample the opposite-patch scene, average paired targets, or permute colors, respectively, while preserving the videos and regression target. The paired construction follows MemCorr-DP~\cite{su2026memcorr}; swapping retains a rendered scene. On fresh scenes, noisy-input swap improves pooled switch accuracy over a learning-rate-tuned baseline by 9.9 points after 10k updates. Longer training narrows the gain, supporting faster learning rather than higher asymptotic accuracy. Gains also occur on ReCamMaster and a real-video test requiring both inputs (Section~\ref{sec:experiments}).

Answer removal and loss reweighting work well together. Min-SNR~\cite{hang2023efficient} improves automatic scores while copying persists. With noisy-input swap, it reaches 80.3\% pooled accuracy after 10k updates, versus 58.9\% for the tuned baseline; the gain holds at a shared learning rate. Midpoint erasure and recoloring perform similarly with Min-SNR, so exact rendered siblings are not shown to be necessary. Two blind raters support the combination, but method-dependent scorer errors limit automatic comparisons (Section~\ref{sec:human}).

Our contributions are:
\begin{itemize}
  \item \textbf{A task for cross-video reasoning in generation:} CrossSwitch requires combining evidence split across two videos, a dependence existing generation benchmarks do not directly test. Paired interventions and shortcut checks make it measurable.
  \item \textbf{Findings enabled by the task:} ordinary training learns \synclane{} late, and a copy probe reveals copying from the noisy input across noise levels, generators and domains. Human checks expose method-dependent scorer misses.
  \item \textbf{Evidence for faster learning:} removing answer information and applying Min-SNR improve early automatic scores. Noisy-input swap, midpoint and recoloring yield similar results with Min-SNR, while copying's causal role remains unresolved.
\end{itemize}

\section{Related Work}
\label{sec:related}

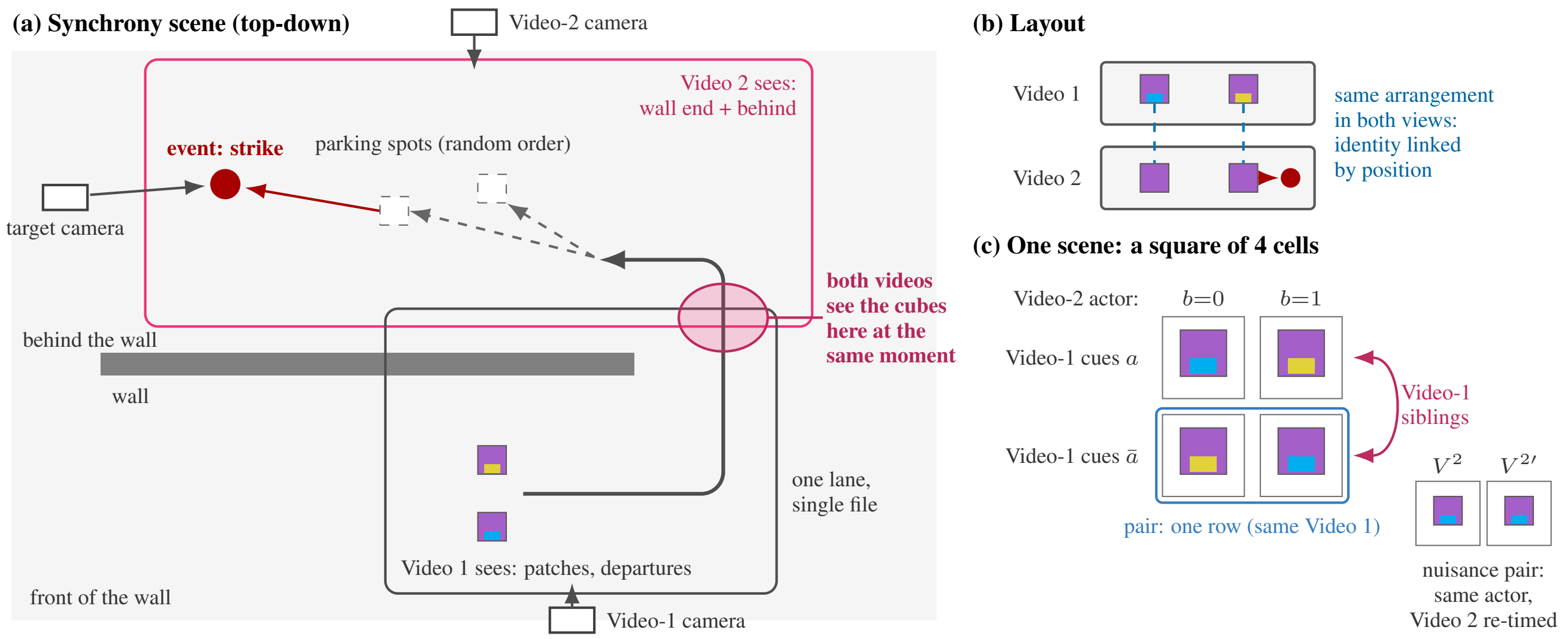
\begin{figure*}[t]
  \centering
  \resizebox{\textwidth}{!}{\input{figures/lane_scene}}
  \caption{\textbf{Scenes and interventions.} (a)~\Synclane{}, viewed from above: both inputs show the cubes rounding the wall at the same moments; parking position is independent of departure order. (b)~Layout preserves spatial arrangement across inputs. (c)~A square crosses patch assignment with actor choice. Rows form pairs; columns, Video-1 siblings. Nuisance pairs re-time Video~2 while preserving the actor.}
  \label{fig:lane}
\end{figure*}

\paragraph{Generating and evaluating views across videos.}
Novel-view and multi-input video generators~\cite{vanhoorick2024gcd,wu2024cat4d,vanhoorick2026anyview,bai2025syncammaster,jiang2025vace} have not been evaluated on whether a generated fact depends on matching evidence across two inputs. Cross-video benchmarks such as SYNCR and MVVBench score answers rather than generated views~\cite{ghazanfari2026syncr,chung2026mvvbench}. CrossSwitch adapts SYNCR's simulator-based single-input controls and evaluates generated views using minimal pairs~\cite{goyal2017making,thrush2022winoground}.

\paragraph{Noise levels and leakage through the noisy input.}
P2 and Min-SNR reweight noise levels~\cite{choi2022perception,hang2023efficient}. Schedule shifts for higher resolutions or more views address residual signal across the noisy input~\cite{esser2024scaling,gao2024cat3d,zhou2025stable}. Other noisy-input shortcuts include residual signal at the highest noise level~\cite{lin2024common,everaert2024exploiting}, conditioning-image leakage in image-to-video models~\cite{zhao2024identifying,ge2025flashi2v}, and noisy masks that diffusion segmenters recover without reading the image~\cite{chen2023pix2seqd,christensen2025ambiguous}; their remedies act on every pixel. Closest to ours, MemCorr-DP observes that noisy actions reveal which behavior to execute and trains diffusion policies on a shared midpoint noisy input~\cite{su2026memcorr} (Section~\ref{sec:method}). We measure noisy-input copying at each noise level in cross-view generation and compare answer corruption with loss reweighting as ways to speed correspondence learning.

\paragraph{Shortcuts and counterfactual training.}
Models take the shortcuts their training signal permits~\cite{geirhos2020shortcut}, including copycat policies in imitation learning~\cite{dehaan2019causal}. Counterfactual and paired samples~\cite{kaushik2020learning,burgert2025motionv2v} add targets or loss terms, and input perturbation and scheduled sampling train on inputs closer to the model's own errors~\cite{ning2023inputperturbation,bengio2015scheduled}; noisy-input swap keeps each branch's target and changes only the noisy input (Appendix~\ref{app:relatedfull}).

\section{Measuring Cross-Video Correspondence}
\label{sec:task}

\subsection{The CrossSwitch benchmark}

\paragraph{Scenes and links.}
Kubric~\cite{greff2022kubric} scenes contain two similar cubes with different side patches. Video~1 reveals the patches; Video~2 hides them but shows which cube strikes a cylinder or reaches a pad first. The target camera shows the actor's patch at the \emph{same time} as Video~2. We define two \emph{links} connecting a Video-1 cube to the Video-2 actor. In \emph{layout}, the cubes' positions correspond across views. In \emph{\synclane}, they do not, but both videos show the cubes passing a wall end at the same moments (Figure~\ref{fig:lane}). \Synclane{} scenes balance departure order and parking position.

\paragraph{Pairs and squares.}
An \emph{item} contains two RGB videos and a target camera, with an empty prompt and no simulator IDs, labels or masks. A \emph{pair}'s two items, or \emph{branches}, share byte-identical Video~1 and the target camera; Video~2 changes which cube acts (Figure~\ref{fig:lane}c). Independently varying Video~1's patch assignment and Video~2's actor gives a four-item \emph{square}; each item is a \emph{cell}. Its \emph{Video-1 sibling} has the other patch assignment but the same Video~2, with a target differing only in the actor's patch. Each input is byte-identical across two cells requiring different answers, so neither input alone suffices. A \emph{nuisance pair} re-times Video~2 while preserving the actor and correct patch. Scene splits are disjoint; model-visible names and prompts carry no labels (Appendix~\ref{app:protocol}; glossary: Appendix Table~\ref{tab:glossary}).

\input{figures/sigma_leak}

\paragraph{Link validity.}
\label{sec:measurement}
The validity check (\emph{link gate}) asks whether the intended evidence identifies the actor and whether specified alternatives fail. On \synclane{}, a rule matching cubes by passage order is 99.3\% accurate; three rules matching positions at different parts of the tracks score 45--51\% (Appendix Table~\ref{tab:links}). Layout and \synclane{} pass their gates; \emph{relay}, a third link in which no cube is visible in both inputs, fails on the original splits (Appendix Table~\ref{tab:gatecounts}) and is tested on a rebuilt set in Section~\ref{sec:reweighting}. These checks exclude the specified rules, not all static cues. Section~\ref{sec:speed} brings together learned position-based classifiers, a classifier-balanced test set and a timing intervention to assess this remaining ambiguity.

\paragraph{Scoring.}
The primary metric, \emph{switch accuracy}, is the fraction of pairs with both correct, different patches. The scorer detects the target event and classifies the actor's patch as correct, the other candidate, an invented cue, or unreadable (an abstention). Abstentions and invented cues count as errors. Its code and thresholds were frozen before independent calibration on clean and edited targets; it passes on every link. We report \emph{pooled} accuracy across layout and \synclane{} pairs.

Independent binary guessing succeeds on 25\% of pairs when both outputs are readable. Multiplying by the share of pairs with two readable outputs gives the \emph{readability-adjusted guessing reference}: 13--19\% on \synclane{} (Appendices~\ref{app:metrics} and~\ref{app:seedlevel}).

Unless stated otherwise, main-text 95\% intervals resample scenes and seeds. Held-out comparisons are pre-specified unless marked secondary, exploratory or descriptive. Calibration on ground truth does not show reliability on generated outputs, which blind raters assess (Section~\ref{sec:human}).

\subsection{Correspondence emerges late in ordinary training}
\label{sec:baseline}

\paragraph{Generators.}
Released generators used zero-shot do not render this camera separation (at most 1 of 88 switches; Appendix~\ref{app:inputhead}). We adapt AnyView~\cite{vanhoorick2026anyview} by adding Video~1 as a third view-token stream, alongside Video~2 and the target. We fine-tune on 1,742 scenes, upsampled from 256 pixels to its native 576 pixels. Standard denoising (\emph{ordinary training}) runs for 5,000 updates (\textbf{Two-video}; about one data pass at batch size 1) or 17,000 (\textbf{Two-video-long}), with three seeds each. The 17,000-update endpoint was fixed before scoring. No run converges (Appendix~\ref{app:models}).

\begin{table}[t]
  \caption{\textbf{Ordinary training at 576 pixels} (held-out, \%; three-seed means). Switch: both branches correct (150 pairs per link); all-four: all cells of a square correct (50 squares); nuisance: both same-actor outputs correct (50 pairs). ---: not evaluated. More: Appendix Table~\ref{tab:corefull}.}
  \label{tab:core}
  \centering\footnotesize
  \setlength{\tabcolsep}{2.2pt}
  \begin{tabular}{@{}l ccc cc cc@{}}
    \toprule
    & \multicolumn{3}{c}{Switch} & \multicolumn{2}{c}{All-four} & \multicolumn{2}{c}{Nuisance} \\
    \cmidrule(lr){2-4}\cmidrule(lr){5-6}\cmidrule(l){7-8}
    Model & Layout & \synchead{} & Pooled & Layout & \synchead{} & Layout & \synchead{} \\
    \midrule
    Two-video & 53.6 & 17.8 & 35.7 & 40.7 & 1.3 & 65.3 & 27.3 \\
    Two-video-long, 10k & 61.3 & 40.9 & 51.1 & --- & --- & --- & --- \\
    Two-video-long & 78.7 & 60.9 & 69.8 & 70.7 & 48.0 & 82.0 & 71.3 \\
    \bottomrule
  \end{tabular}
\end{table}

\paragraph{Held-out results.}
Ordinary training learns layout before \synclane{}. After 5,000 updates, \synclane{} switch accuracy is close to guessing among readable outputs, and the most common square failure ignores Video~1 (Appendix~\ref{app:reversed}). Longer training raises \synclane{} accuracy from 17.8\% to 60.9\%, with every long-run seed beating every short-run seed. Complete squares and nuisance pairs also improve (Table~\ref{tab:core}). Validation curves show the same delayed rise in \synclane{} (Appendix Figure~\ref{fig:emergence}); Section~\ref{sec:speed} examines whether static cues explain it.

\subsection{Models copy the answer from the noisy input}
\label{sec:diagnosis}

\paragraph{Answer information in the noisy input.}
The noisy input usually retains enough information to identify the patch. An ideal observer given both candidate targets can read it in about 95\% of training draws at 576 pixels (Appendix~\ref{app:decision}). AnyView uses $y=z+\sigma\epsilon$, data scale $\sigma_{\rm data}=1$ and loss weight $w(\sigma)=(\sigma^2+1)/\sigma^2$~\cite{karras2022elucidating} (noise distribution: Appendix~\ref{app:models}).

Ordinary models copy the wrong patch planted in the noisy input. The \emph{copy probe} preserves the videos but noises the wrong-patch sibling. We project one denoiser call's clean prediction onto the difference between the target latents; its side of their midpoint identifies the patch it follows.

Copying persists across much of the training noise range: up to $\sigma\approx5.4$--$10.9$, covering 65--84\% of training draws. It also occurs on held-out cells. Longer training shifts predictions toward Video~1 (Figure~\ref{fig:leak}); full sampling from the planted patch agrees (Appendix Figure~\ref{fig:decisionprobetrain}).

\paragraph{Beyond one model and synthetic scenes.}
Copying also occurs on real video. Released AnyView copies a planted color on Argoverse~2 vehicles up to $\sigma\approx2.2$ and on garments in the CMU Panoptic studio~\cite{joo2019panoptic} up to $\sigma\approx5.4$. These cover 33\% and 65\% of training draws.

Other model families also copy. ReCamMaster~\cite{bai2025recammaster}, a flow-matching model, follows a planted patch on nearly all cells up to $\sigma\approx2$; the image editor InstructPix2Pix~\cite{brooks2023instructpix2pix} also copies (Appendix~\ref{app:decision}).

\paragraph{An incentive to ignore conditioning.}
Copying at low noise can be optimal under ordinary denoising: a readable answer leaves little loss for Video~1 to reduce. A binary model of this incentive fails to rank held-out training variants (Appendix~\ref{app:decision}). We therefore treat copying as a diagnostic and possible contributor to slow learning, not an established cause.

\section{Speeding Up Learning}
\label{sec:method}

CrossSwitch enables controlled tests of whether training changes accelerate this task. We compare three forms of answer removal---noisy-input swap, midpoint erasure and recoloring---and the existing Min-SNR loss weighting.

\paragraph{Noisy-input swap.}
Noisy-input swap changes the answer visible in the noisy input while preserving the conditioning videos and correct regression target. CrossSwitch supplies paired scenes that differ in patch assignment. For inputs $(V^1_a,V^2_b)$, target camera $c^\star$ and target latent $z_{a,b}$, the sibling target $z_{\bar a,b}$ retains the actor and camera but changes the patch. At $\sigma\ge1$, with probability $p=0.5$, we construct the noisy input from this sibling and train the denoiser $D_\theta$:
\begin{equation}
\begin{aligned}
y&=z_{\bar a,b}+\sigma\epsilon,\\
\mathcal L&=w(\sigma)\,\bigl\|D_\theta(y;\sigma,V^1_a,V^2_b,c^\star)-z_{a,b}\bigr\|^2.
\end{aligned}
\label{eq:swap}
\end{equation}
On a swapped draw, copying the visible patch raises the regression loss. The validation-selected cutoff $\sigma\ge1$ leaves cleaner inputs unchanged; ordinary pairs and unswapped square samples train as usual. Noisy-input swap adds sibling encoding with the frozen tokenizer, but no denoiser pass or inference change.

Randomizing the swap makes the visible patch uninformative about the correct answer within sibling pairs ($p=0.5$). Always swapping would leave a cue that an ideal reader could invert, although it performs similarly on validation at 10k (Appendix~\ref{app:randomswap}, Table~\ref{tab:mechanismfull}).

\paragraph{Midpoint erasure and recoloring.}
Midpoint erasure and recoloring also remove answer information. Following MemCorr-DP~\cite{su2026memcorr}, our \emph{midpoint} control shares $y=\tfrac12(z_{a,b}+z_{\bar a,b})+\sigma\epsilon$ across branch-specific targets. Noisy-input swap instead samples a rendered target. \emph{Sibling-free recoloring} permutes the target's colors, requiring no paired render but relying on palette symmetry. Section~\ref{sec:experiments} compares them; Appendix~\ref{app:randomswap} compares schedules with MemCorr-DP.

\paragraph{Reweighting the loss.}
The second change leaves the noisy input intact and reweights training draws. We use Min-SNR~\cite{hang2023efficient}, replacing $w(\sigma)$ with $\min(1/\sigma^2,5)$. This removes most of the loss weight at high noise and concentrates the remainder at low noise (Appendix Figure~\ref{fig:effweight}). Min-SNR leaves the answer available, and models keep copying (Section~\ref{sec:reweighting}); the idealized account did not predict its benefit. We test it alone and with noisy-input swap (swap + Min-SNR; in tables, Swap and MS abbreviate noisy-input swap and Min-SNR).

\section{Experiments}
\label{sec:experiments}

\paragraph{Setup.}
We test which training changes improve split-evidence generation. Unless stated otherwise, results are pooled automatic switch accuracy on the fresh split at 256 pixels, the render resolution, after 10k updates of an 18k-update schedule at batch size 1. The baseline, LR$\times$2, doubles ordinary training's base learning rate; higher rates do not help on validation (Appendix~\ref{app:methodstudy}). Comparisons stay within one resolution.

The fresh-split test is pre-specified, with six new seeds per arm and recipes fixed before any output was seen (Table~\ref{tab:methodheldout}). The \emph{first split} is the earlier held-out test (Appendix~\ref{app:confirm}). Figure~\ref{fig:forest} summarizes the contrasts; Section~\ref{sec:human} compares them with human judgments.

\begin{table}[t]
  \caption{\textbf{Fresh-split results at 256 pixels} (automatic switch accuracy, \%; seed means; 10k updates). Rate: multiples of $4\times10^{-6}$. Six-seed rates were pre-specified: base for Min-SNR arms, doubled for others. The shared-rate test uses three of those seeds, hence different baseline means. 17.5k updates: Figure~\ref{fig:forest} and Appendix~\ref{app:heldoutmore}. Human checks: Table~\ref{tab:raterscorer}; earlier split: Appendix Table~\ref{tab:methodheldoutfull}.}
  \label{tab:methodheldout}
  \centering\small
  \setlength{\tabcolsep}{3pt}
  \begin{tabular*}{\linewidth}{@{\extracolsep{\fill}}lcccc@{}}
    \toprule
    Training & Rate & Layout & \synchead{} & Pooled \\
    \midrule
    \multicolumn{5}{@{}l}{\emph{Pre-specified confirmation: six new seeds}} \\
    Ordinary (LR$\times$2) & 2 & 71.8 & 46.0 & 58.9 \\
    Noisy-input swap & 2 & 74.2 & 63.4 & 68.8 \\
    Min-SNR & 1 & 75.4 & 60.3 & 67.9 \\
    Swap + Min-SNR & 1 & 84.2 & 76.4 & \textbf{80.3} \\
    \midrule
    \multicolumn{5}{@{}l}{\emph{Shared-rate comparison: three seeds}} \\
    Ordinary (LR$\times$2) & 2 & 73.3 & 44.2 & 58.8 \\
    Noisy-input swap & 2 & 75.8 & 60.9 & 68.3 \\
    Min-SNR & 2 & 79.3 & 64.2 & 71.8 \\
    Swap + Min-SNR & 2 & 84.9 & 72.7 & 78.8 \\
    Midpoint + Min-SNR & 2 & 83.3 & 75.1 & 79.2 \\
    Recolor + Min-SNR & 2 & 82.7 & 76.2 & 79.4 \\
    \bottomrule
  \end{tabular*}
\end{table}

\begin{figure*}[t]
  \centering
  \input{figures/random_strip_fresh}
  \caption{\textbf{Three random fresh-split \synclane{} pairs.} The draw rule was fixed before viewing outputs and excludes human-check pairs. Each arm: training seed 10, 10k updates, same starting noise. Crops: true actor at contact $+4$ frames. Frozen scorer: \checkmark{} correct, $\times$ other cube's patch, inv.\ invented, -- unreadable. A switch requires both branches correct. All pairs: Table~\ref{tab:methodheldout}; first-split examples: Appendix~\ref{app:qual}.}
  \label{fig:strip}
\end{figure*}

\begin{figure}[t]
  \centering
  \input{figures/forest}
  \caption{\textbf{Learning-speed contrasts} (automatic switch accuracy, points). Swap minus ordinary training at matched updates (256 pixels: LR$\times$2; 576: untuned). Last two rows: swap + Min-SNR minus swap. Bars: 95\% scene-and-seed bootstrap intervals. Filled: pre-specified primary; open: secondary. More: Appendix Tables~\ref{tab:methodheldoutfull} and~\ref{tab:contrasts}.}
  \label{fig:forest}
\end{figure}

\subsection{Removing the answer speeds up learning}
\label{sec:speed}

\paragraph{Fresh-split confirmation.}
Noisy-input swap learns faster than the tuned baseline on fresh scenes, mostly on \synclane{}: a 9.9-point lead at 10k ([$+2.9$, $+16.7$]; Table~\ref{tab:methodheldout}). The lead narrows at 17.5k (secondary; Figure~\ref{fig:forest}). Both arms share the learning rate. Figure~\ref{fig:strip} shows three random pairs. The first split shows the same early advantage (Appendices~\ref{app:confirm} and~\ref{app:seedlevel}).

\paragraph{Robustness and limits.}
The early gain also appears at AnyView's native 576 pixels and on ReCamMaster. The former uses an untuned baseline ($+22.9$ points; Appendix Table~\ref{tab:corefull}); the latter uses the flow-matching generator's own recipe ($+19.2$ points; Appendix~\ref{app:recam}).

Longer training narrows the gain. On a 40k-update schedule, noisy-input swap leads at 20k, but at 40k the difference is $+0.4$ [$-4.0$, $+4.8$], too wide to establish equivalence (Figure~\ref{fig:forest}).

\paragraph{Static cues and timing.}
Static cues remain a benchmark limitation, but do not fully explain the results. Two classifiers using positions without temporal matching identify the actor on 79--87\% of the original \synclane{} items. We therefore test a fresh set constructed before generation to put both classifiers at chance.

Late learning persists on this classifier-balanced set. Longer ordinary training raises accuracy from 18.0\% to 59.8\% at 576 pixels. At 256 pixels, noisy-input swap's lead over LR$\times$2 is unresolved after resampling seeds; adding Min-SNR gives a resolved gain over noisy-input swap (Appendix~\ref{app:worstgroup}).

Models are sensitive to temporal alignment. Delaying Video~1 removes correct same-moment matches while keeping departure order and static cues. Every tested arm loses accuracy mainly on \synclane{}; for swap + Min-SNR, the drop exceeds layout's by 40.3 points ([$+32.0$, $+48.7$]; two seeds). The delay also changes the input distribution and was tested only at 256 pixels, so it does not identify a unique matching strategy (Appendix~\ref{app:tshift}).

\subsection{Reweighting complements answer removal}
\label{sec:reweighting}

\paragraph{Min-SNR improves scores while copying persists.}
Loss reweighting also speeds learning, without removing the answer. Min-SNR scores close to noisy-input swap on the fresh split (67.9 versus 68.8\%) and above it at a shared learning rate (71.8 versus 68.3\%; Table~\ref{tab:methodheldout}; neither difference was tested), yet still copies below $\sigma\approx3$ (Appendix Table~\ref{tab:copyprobe256}). Human ratings do not resolve its gain over LR$\times$2 (Section~\ref{sec:human}).

\paragraph{Combining Min-SNR and noisy-input swap.}
Combining Min-SNR and noisy-input swap gives the highest automatic accuracy: 80.3\% at 10k (Table~\ref{tab:methodheldout}). Every combination seed exceeds every seed of either component and of LR$\times$2. Its lead over noisy-input swap is 11.5 points ([$+5.3$, $+18.1$]) and persists at 17.5k (Figure~\ref{fig:forest}).

The lead survives a shared learning rate: three seeds per arm at LR$\times$2 give 10.4 points ([$+2.9$, $+19.0$]; one-sided seed-level $p=0.058$) over noisy-input swap, and the combination also exceeds Min-SNR. The interaction interval includes zero, so this supports combining the changes without establishing synergy (Appendix~\ref{app:heldoutmore}).

\paragraph{Alternative ways to remove the answer.}
Rendered siblings are not needed for these color-task gains. On the first split, midpoint erasure performs similarly to noisy-input swap, and sibling-free recoloring performs better at a matched rate (secondary). With Min-SNR at a shared rate, all three score near 79\% pooled (Table~\ref{tab:methodheldout}). Their seed-resampled intervals slightly exceed the pre-specified equivalence margin. We thus find neither an advantage for replacement over erasure nor evidence that rendered siblings are needed (Appendix~\ref{app:heldoutmore}).

\begin{table}[t]
  \caption{\textbf{Harder variants} (new held-out scenes, no retraining; switch accuracy, \%, at 10k / 17.5k; two seeds per arm). Relay: trained link, rebuilt with position classifiers at chance. Gate event: untrained. Failed gates preclude evaluation. Readability-adjusted guessing: 15--23\%. Seeds and gates: Appendix~\ref{app:difficulty}.}
  \label{tab:difficulty}
  \centering\footnotesize
  \setlength{\tabcolsep}{2.2pt}
  \begin{tabular}{@{}lcccc@{}}
    \toprule
    Variant (link) & LR$\times$2 & Swap & Min-SNR & Swap + MS \\
    \midrule
    Relay, balanced & 46.3 / 63.0 & 57.7 / 68.0 & 55.0 / 68.0 & \textbf{70.3} / \textbf{75.3} \\
    Gate event (layout) & 9.0 / 5.3 & 27.3 / 24.0 & 12.7 / 11.7 & \textbf{51.7} / \textbf{41.3} \\
    Gate event (\synchead{}) & \multicolumn{4}{c}{fails gates: oracle coverage, missing-patch control} \\
    Three cubes (layout) & \multicolumn{4}{c}{fails scorer ceiling: 89.5\% on round trips} \\
    Three cubes (\synchead{}) & \multicolumn{4}{c}{fails link gate: endpoint matcher 75.0\%} \\
    \bottomrule
  \end{tabular}
\end{table}

\paragraph{Harder variants.}
Every arm scores well above guessing on a fresh relay set; swap + Min-SNR leads LR$\times$2 by 24.0 points ([$+14.7$, $+34.0$]; pre-specified). Relay is in every arm's training mix, so this is an in-distribution test. Before any model output, the 150 pairs were built to put both position classifiers at 50\% and the three spatial matchers at 47--55\%. These new held-out variants were built after the main results and evaluated without retraining (Table~\ref{tab:difficulty}).

Accuracy drops for every arm on an untrained event: a cube driving through a colored gate (layout, 150 new pairs). LR$\times$2 and Min-SNR fall below guessing; noisy-input swap reaches 27.3\% and swap + Min-SNR 51.7\% (Table~\ref{tab:difficulty}). The combination's lead over LR$\times$2 persists at 17.5k (Appendix~\ref{app:difficulty}). Unlike transfer to an untrained attribute (Section~\ref{sec:discussion}), adding Min-SNR helps here. The other variants fail their gates and are not evaluated.

\subsection{Human evaluation of the automatic scorer}
\label{sec:human}

\begin{table}[t]
  \caption{\textbf{Human raters and the scorer} (first split, 40 pairs; switch accuracy, \%; 10k). Last two arms rated in a later session; LR$\times$2: 39 pairs for rater 1, 40 for the scorer.}
  \label{tab:raterscorer}
  \centering\small
  \setlength{\tabcolsep}{3pt}
  \begin{tabular*}{\linewidth}{@{\extracolsep{\fill}}lccc@{}}
    \toprule
    Training & Rater 1 & Rater 2 & Scorer \\
    \midrule
    Ordinary & 57.5 & 57.5 & 42.5 \\
    LR$\times$2 & 66.7 & 72.5 & 40.0 \\
    Noisy-input swap & 77.5 & 80.0 & 42.5 \\
    Min-SNR & 70.0 & 70.0 & 57.5 \\
    Swap + Min-SNR & 100 & 100 & 77.5 \\
    \bottomrule
  \end{tabular*}
\end{table}

Human judgments expose method-dependent scorer misses. Table~\ref{tab:raterscorer} compares independent blind raters with the scorer on 40 random first-split pairs from one seed; intervals here resample pairs. The raters agree closely (Cohen's $\kappa=0.91$), but their agreement with the scorer is much lower.

The scorer misses correct outputs at different rates across methods. In this rating set, each rater confirms at least 96\% of scorer-positive outputs pooled across the models they rated. This is precision, not recall. Misses often arise from failed event detection and can distort automatic comparisons (Appendix~\ref{app:judgecheck}). On these pairs the scorer undercounts every arm, most for noisy-input swap (by 35.0--37.5 points) and least for Min-SNR (12.5; Table~\ref{tab:raterscorer}). Against LR$\times$2 (25.7--32.5), it therefore understates noisy-input swap's gain and overstates Min-SNR's and, by 5--10 points, the combination's.

The raters support noisy-input swap over ordinary training, but not a resolved gain over the tuned baseline. Averaged over raters, its lead over LR$\times$2 is 11.5 points ([$-2.6$, $+26.9$]; 39 complete pairs). The difference lies on layout; the \synclane{} contrast is unresolved. The scorer instead places most of the full-set gain on \synclane{}.

Human ratings also leave Min-SNR's gain over LR$\times$2 unresolved. At the later checkpoint, one rater gives ordinary training, LR$\times$2 and noisy-input swap equal scores (Appendix~\ref{app:judgecheck}).

Both raters support the combination over either component. Each judges swap + Min-SNR correct on all 40 pairs, 30.0 points above Min-SNR in the same session ([$+16.2$, $+43.8$]; secondary). Averaged over raters, its lead over LR$\times$2 is 30.8 points ([$+17.9$, $+44.9$]; primary), but that comparison spans sessions. This ceiling result comes from scenes both raters had already rated for other arms.

\begin{figure}[t]
  \centering
  {\footnotesize\textbf{(a)} Split evidence: both inputs needed}\\[2pt]
  {\scriptsize\setlength{\tabcolsep}{1pt}%
  \begin{tabular}{@{}r@{\hspace{2pt}}ccc@{}}
    & Video 1 (color crops) & Video 2 (grayscale) & target \\
    \raisebox{.07\linewidth}{b1} & \includegraphics[width=.31\linewidth]{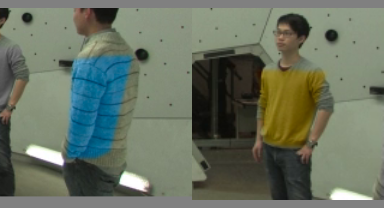} & \includegraphics[width=.31\linewidth]{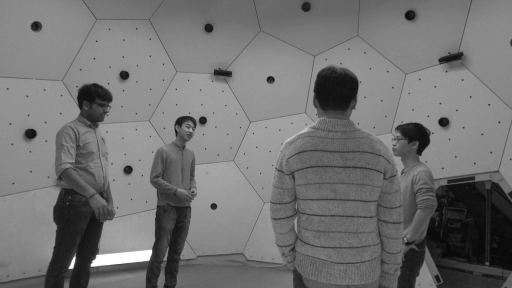} & \includegraphics[width=.31\linewidth]{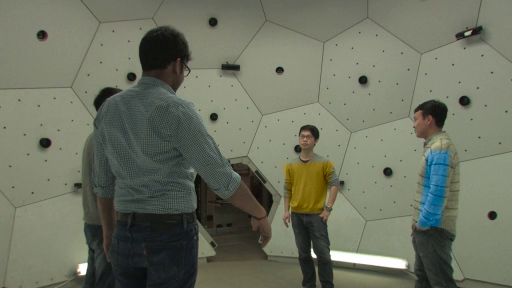} \\
    \raisebox{.07\linewidth}{b2} & \includegraphics[width=.31\linewidth]{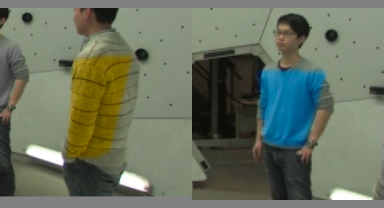} & \includegraphics[width=.31\linewidth]{figures/panoptic_split_example/v2.png} & \includegraphics[width=.31\linewidth]{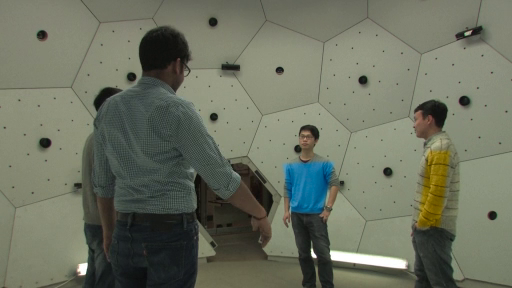} \\
  \end{tabular}}\\[2pt]
  \input{figures/panoptic_split}\\[3pt]
  {\footnotesize\textbf{(b)} Attribute propagation: either input suffices}\\[1pt]
  \input{figures/panoptic_bars}
  \caption{\textbf{Real video} (CMU Panoptic; 5k updates; two seeds per arm). (a) Top: one test item (ground truth, frame 20). Branch b2 exchanges the two colors in Video~1 and the target while Video~2 stays the same, so both branches are right only if colors are bound across videos. Bottom: correct binding on 388 items. (b) Correct binding: edit reaches only the intended person, in place. Position-tolerant: wherever rendered. Natural: unedited garments, in place. $^\dagger$Trained on recolored videos, making the edit familiar; control omits only swapping. Contrasts: Appendix~\ref{app:panopticb}.}
  \label{fig:panoptic}
\end{figure}

\subsection{Real multi-camera video}
\label{sec:realvideo}

Fine-tuning AnyView on CMU Panoptic footage~\cite{joo2019panoptic} tests generation of a third-camera view from two others. We separate correspondence requiring both inputs from attribute propagation that either input could support (Appendix~\ref{app:panopticb}).

\paragraph{Evidence split across both inputs.}
Noisy-input swap improves binding when both real-video inputs are needed. Video~1 shows two people's garment colors in randomly ordered crops; Video~2 shows the grayscale scene. Both people must receive their correct colors, including when Video~1's colors are exchanged.

Correct binding rises from 15.9\% to 25.4\% after 5,000 updates (Figure~\ref{fig:panoptic}a; $+9.5$ [$+4.4$, $+14.8$]; pre-specified; two seeds; Appendix~\ref{app:panopticb}). The gain remains among outputs that place both people, while placement itself does not improve.

Min-SNR alone reaches 35.4\% and the combination 40.7\%. Absolute accuracy remains low. Appearance and motion timing link the inputs; this is not a test of synchrony alone.

\paragraph{Attribute propagation.}
Min-SNR improves binding in a simpler test where either input can carry a garment edit (Figure~\ref{fig:panoptic}b). About half the gain comes from better placement of the person. Noisy-input swap stops copying a planted edit at every tested $\sigma\ge1$, unlike a control trained on the same data (Appendix~\ref{app:panopticb}).

\section{Discussion and Limitations}
\label{sec:discussion}

CrossSwitch makes cross-video reasoning in generation testable through controlled interventions. Ordinary training learns \synclane{} late; answer removal and Min-SNR accelerate automatic scores and work best together on the trained attribute. Midpoint and recoloring perform similarly with Min-SNR, supporting answer removal without favoring rendered replacement. Longer training narrows the swap's lead; higher asymptotic accuracy is not established.

Models across generators and domains copy readable answers planted in the noisy input. Whether this slows learning remains unresolved: Min-SNR helps while copying persists. Both raters support the combination, but human checks do not resolve either component's gain over the tuned baseline. Method-dependent scorer errors limit the measured gain's magnitude and interpretation.

\emph{Established:} ordinary training learns \synclane{} late; models copy answers planted in the noisy input, across generators and on real video; on the trained attribute, answer removal with Min-SNR raises automatic accuracy early, and two raters confirm the combination's lead. \emph{Open:} whether copying causes the slow learning, whether the gains persist at convergence or with larger batches, and how they extend to other attributes.

\paragraph{Limitations.}
\emph{Scope.} CrossSwitch tests attribute correspondence in a controlled family of scenes with two look-alike cubes, mostly at 256 pixels, with additional checks at 576 pixels and on ReCamMaster. The 576-pixel and ReCamMaster baselines were not learning-rate tuned; at 256 pixels, doubling AnyView's rate alone adds 5.7 points. Noisy-input swap requires counterfactual targets, rendered here and recolored on real video. About half of Min-SNR's real-video binding gain reflects placement. Noisy-input swap adds 9.5 points on the real-video split-evidence test at one checkpoint (Section~\ref{sec:realvideo}). \emph{Measurement.} The scorer relies on patch hue and is not robust to global color shifts. Only one human check has two raters, and scorer--rater disagreement varies by method (Section~\ref{sec:human}). \emph{Generality.} Correspondence learned on CrossSwitch does not transfer to real handoffs (Appendix~\ref{app:real}). No training variant transfers from color to previously unseen patch placement on \synclane{} (Appendix Tables~\ref{tab:corefull} and~\ref{tab:methodheldoutfull}). On layout, adding Min-SNR reduces noisy-input swap's transfer accuracy (24.7 versus 42.7\% at 10k; secondary; Appendix~\ref{app:heldoutmore}). The gains may depend on small appearance attributes determined late in sampling; those determining coarse structure at high noise remain untested. Further limitations: Appendix~\ref{app:morelimits}. Data and code will be released.
\section*{Acknowledgments}
This paper is supported in part by the Army Research Office under grant number W911NF-21-1-0155 and by the New York University Abu Dhabi (NYUAD) Center for Artificial Intelligence and Robotics, funded by Tamkeen under the NYUAD Research Institute Award CG010. Additional support was provided by the NYU IT High Performance Computing resources, services, and staff expertise.

\section*{LLM Usage Statement}
The authors take full responsibility for all submitted content, regardless of any LLM or AI-tool use. We gave the manuscript to LLM providers (OpenAI ChatGPT/GPT and Anthropic Claude) to obtain reviewer-style feedback. LLM tools assisted with writing and editing text, related work and citations, figures and visual material, and synthetic data and dataset creation, by editing or refining author-produced content. For code and implementation, the authors reviewed, edited or approved tool-generated work. The authors declare no LLM or AI-tool use for experiment design, execution or analysis of results, or for research ideas, hypotheses, claims or conclusions.

{
    \small
    \bibliographystyle{ieeenat_fullname}
    \bibliography{main}
}
\clearpage
\maketitlesupplementary
\appendix
\let\oldappsection\section\renewcommand{\section}{\FloatBarrier\oldappsection}
\renewcommand{\topfraction}{0.95}\renewcommand{\dbltopfraction}{0.95}\renewcommand{\bottomfraction}{0.6}\renewcommand{\textfraction}{0.05}
\renewcommand{\floatpagefraction}{0.7}\renewcommand{\dblfloatpagefraction}{0.7}\setcounter{topnumber}{4}\setcounter{dbltopnumber}{3}\setcounter{bottomnumber}{2}\setcounter{totalnumber}{6}
\paragraph{Overview.} Appendix~\ref{app:relatedfull} expands the related work; Appendix~\ref{app:protocol} specifies the benchmark; Appendix~\ref{app:metrics} reports scorer calibration and human checks; Appendix~\ref{app:decision} describes the probes; and Appendix~\ref{app:models} details training, inference and compute. Appendix~\ref{app:results} presents results in main-text order. Appendix~\ref{app:other} reports other interventions and negative results, including our first method; Appendix~\ref{app:deviations} records protocol deviations; and Appendix~\ref{app:positioning} discusses further limitations. The \emph{576-pixel study} (Section~\ref{sec:baseline}) and \emph{256-pixel method study} (Section~\ref{sec:experiments}) use the same training scenes, but their results are not directly comparable.

\oldappsection{Extended related work}
\label{app:relatedfull}

This section expands Section~\ref{sec:related}.

\paragraph{Closest evaluations.} Our test combines four features: two observed moving videos as input, both needed for a defined relation, a scored relation in the generated view, and a paired change to one input video that should switch that relation. To our knowledge, no prior work has all four. LiveStre4m generates a target view from two synchronized video streams and reports rendering metrics~\cite{quesado2026livestre4m}; such metrics, like multi-modal perceptual metrics~\cite{ghazanfari2024unisim}, score similarity to a reference rather than a defined relation. The MiniMax-H3 evaluation scores relations inferred across complementary views in generated video, but those views are still images~\cite{zhao2026minimaxh3}. SpotEdit benchmarks editing guided by a reference image, also on stills~\cite{ghazanfari2025spotedit}. DreamRelation scores relations in generated video, with exemplar videos used in training rather than as inputs~\cite{wei2025dreamrelation}. What-If World evaluates paired generations, but the changed variable is a text action~\cite{cai2026whatifworld}. SYNCR and MVVBench test cross-video evidence but score answers~\cite{ghazanfari2026syncr,chung2026mvvbench}, as does Chain-of-Frames, which improves video reasoning in multimodal LLMs~\cite{ghazanfari2026chainofframes}. In concurrent work, XTrack uses a video diffusion transformer to put ego and exo videos in correspondence, but predicts masks rather than generated views~\cite{jin2026xtrack}.

\paragraph{Leakage through the noisy input.}

In image-to-video models the conditioning image leaks into the noisy input, and models over-rely on it~\cite{zhao2024identifying,ge2025flashi2v}; in image restoration, the similarity between the noisy input and a paired condition offers a shortcut at low noise~\cite{liao2024denoising}. Cascaded models corrupt the conditioning~\cite{ho2022cascaded,saharia2022photorealistic}, and masked diffusion transformers drop noisy-input tokens~\cite{gao2023mdt,zheng2024maskdit}. Diffusion segmenters can recover a noisy mask without reading the image and are therefore trained with lower signal-to-noise inputs or noised predictions~\cite{chen2023pix2seqd,lai2023denoising,vangansbeke2024simple,christensen2025ambiguous}, and predictive video diffusion is argued to take a shortcut through the noisy input~\cite{zhang2026foresight}; these remedies act on every pixel.

Closest to ours, MemCorr-DP observes that in diffusion policies the noisy action reveals which behavior to execute, and trains on a shared midpoint noisy input with branch-specific targets at selected high-timestep indices, on one simulated task~\cite{su2026memcorr}. These indices are not equivalent to CrossSwitch's high-noise band (Appendix~\ref{app:randomswap}). We measure the corresponding noisy-input, item-specific leak in cross-view video generation per noise level and relate it to how fast correspondence is learned; noisy-input swap randomizes answer evidence in the noisy input and keeps the regression target, and we compare it with reweighting and shifting noise levels (Section~\ref{sec:experiments}).

Input perturbation corrupts the noisy input with random noise against exposure bias~\cite{ning2023inputperturbation,ning2024elucidating}, and scheduled sampling~\cite{bengio2015scheduled} and correction training~\cite{ahn2026correction} likewise train on inputs closer to the model's own errors; noisy-input swap instead uses a plausible sibling with a different answer, retaining the conditioning and regression target; our decision probe starts sampling from a noised input as in SDEdit~\cite{meng2022sdedit}.

\section{Benchmark Specification and Link Validity}
\label{app:protocol}

\begin{table*}[!tp]
  \caption{Glossary.}
  \label{tab:glossary}
  \centering\footnotesize
  \setlength{\tabcolsep}{4pt}
  \begin{tabular}{@{}p{0.2\textwidth}p{0.76\textwidth}@{}}
    \toprule
    Term & Meaning \\
    \midrule
    Scene base (scene) & One simulated scene: objects, cameras, lighting and pre-event motion. Splits are disjoint in scene bases. \\
    Item / branch & One evaluation case: Video~1, Video~2 and a target camera, with a ground-truth target video. \\
    Pair & Two items with the same Video~1 and target camera; Video~2 changes which cube acts. \\
    Square, cell & Two Video-1 cue assignments crossed with two Video-2 actor branches: four items (cells). \\
    Video-1 sibling & The cell with the same Video~2 and the other cue assignment; its target differs only in the actor's patch. \\
    Nuisance pair & Video~2 re-timed but the actor unchanged; the depicted identity should not change. \\
    Link & What connects a Video-1 cube to the Video-2 actor: \emph{layout} (positions), \emph{\synclane{}} (cubes seen at the same moment), \emph{relay} (identity carried across a gap; fails its gate). \\
    Link gate & Per-selection check that the intended matcher succeeds and spatial-only matchers stay at 40--60\%. \\
    Switch accuracy & Share of pairs whose two outputs both depict their correct, different patches; all-four: all cells of a square correct. \\
    Coverage, answered accuracy & Share of outputs the scorer reads; accuracy among them. \\
    Two-video, Two-video-long & AnyView with Video~1 as a third view-token stream, after 5,000 and 17,000 updates (576 pixels). \\
    LR$\times$2 & Ordinary training at a doubled learning rate (the tuned baseline at 256 pixels). \\
    +CF & Our first, pre-specified method: an auxiliary counterfactual loss on pure-noise targets (failed its criterion). \\
    Noisy-input swap & Noisy input built from the Video-1 sibling's target, regression target unchanged. Midpoint: from the siblings' mean. Sibling-free recolor: from the cell's own target with colors permuted. Both act as noisy-input swap does, with $p=0.5$ at $\sigma\ge1$. \\
    Patch weighting, Min-SNR & Loss up-weighted on the answer region (simulator or MLLM mask); loss weight $\min(1/\sigma^2,5)$ in place of EDM's. \\
    Copy / decision probe & One denoiser call / a full sampling run starting from a noisy input that carries the sibling's patch. \\
    Sampler splicing & Sampling with one checkpoint above a noise level $S$ and another below it. \\
    $\sigma^\ast$ & Noise level up to which an ideal observer tells the two candidate targets apart with 60\% accuracy. \\
    \bottomrule
  \end{tabular}
\end{table*}

This section specifies the data units and splits, rendering requirements, paired interventions, and validity checks used to select CrossSwitch scenes.

\subsection{Units, splits, and sample budget}
\label{app:data}
A \emph{scene base} contains geometry, appearances, three fixed cameras, lighting, and scripted pre-event motion. A \emph{pair} has two actor branches and one target video per branch; both branches receive the same encoded Video-1 file. A \emph{square} crosses two Video-1 cue assignments with two Video-2 actor choices, giving four target videos. A \emph{nuisance pair} changes Video~2 but keeps the actor. Scene bases, not branch clips, are the units for splitting and for uncertainty.

The three links are layout, \synclane{} and relay. Each split is stratified by link and by event family (\texttt{strike}/\texttt{order}), so a final cell holds 75 pairs of each event. Patch colors cycle through the three unordered pairs of yellow, cyan, and pink, excluding confusable body--patch combinations; orbit side and acting cube are balanced. Each ledger row draws seeds from its own disjoint range. Seeds and scene-specification hashes are checked unique across splits, and a square's four cells and a nuisance pair's two cells stay in one split.

\paragraph{Realized training set.}
Every evaluation split meets its quota. Training differs in one way: layout's design pool was exhausted at 296 squares, and a frame-count scan before training removed every square with a truncated target video. All models therefore train on the same 1,742 bases: 900 ordinary pairs (300 per link) and 842 squares (layout 295, \synclane{} 247, relay 300), or 5,168 target clips.

\subsection{Scene design and rendering requirements}
\label{app:render}
Videos are 256$\times$256 RGB, 60 frames at 12\,fps. Each scene has two equal-body cubes of side 0.62--0.78 scene units, a colored side patch of 0.6 times the cube side, a wall occluder, scripted keyframe motion, static cameras, and identical lighting and initial state across branches. The target camera differs from Video~1 by 70--115$^\circ$ in azimuth, at distance 9--11 units with focal length 24--70\,mm. AnyView samples 41 input frames evenly from the 60. The model receives encoded RGB videos and camera matrices, never simulator IDs, masks, labels, or the question.

In layout, identities are linked by persistent spatial arrangement. In \synclane{} scenes, two cubes start beside a wall, merge into one lane, pass its end in turn, and are co-visible to both input cameras for at least three frames. In relay scenes, Video~2 frames only the region behind the wall, and no cube is co-visible in the two inputs.

Admission requires analytic design checks and rendered-pixel checks: each patch is visible in Video~1 on at least eight frames with at least 20 mask pixels; Video~2 contains zero patch pixels; actor and landmark are visible at the queried event in Video~2 and in the target; and the target shows at least 15 actor-patch pixels in three of the seven frames around the event. Lane cells also require their co-visibility and order conditions. Independently rendered Video-1 branches may differ after the branch point by at most four connected pixels above a mean RGB difference of 8/255, and not at all before it. The released input uses one canonical encoded Video-1 file, so the model input is byte-identical across branches.

\subsection{Squares and nuisance pairs}
\label{app:interventions}
To build a square, choose cubes $A,B$, distinct patches $p,q$, and one scene, trajectory set, event family, and target camera. Video-1 mapping $a$ is $A\mapsto p,B\mapsto q$ and mapping $b$ is $A\mapsto q,B\mapsto p$. For each mapping, Video~2 has one branch in which $A$ acts and one in which $B$ acts. The correct labels are $(p,q)$ in row $a$ and $(q,p)$ in row $b$. Both cue mappings receive the same canonical Video-2 file for a fixed actor, and both actor branches receive the same canonical Video-1 file for a fixed mapping. Independent Video-2 renders may differ by at most 32 gray path-tracing pixels and by no patch-colored blob larger than four pixels. We check file hashes, decoded-frame equality, and the absence of labels in model-visible filenames. Within a square, a predictor that sees only one input has identical evidence for two opposite labels, so its expected accuracy is at most 50\% and its all-four accuracy is zero.

A nuisance pair keeps the actor, cue mapping, event type, and target camera fixed and re-times post-branch motion by a factor in $[0.75,0.90]$ or $[1.12,1.35]$. It requires at least five frames with 50 changed foreground pixels in Video~2 and keeps the target event visible. A generator passes only if it depicts the same correct actor in both branches.

\subsection{Link gates}
\label{app:linkgates}
The link gate (Section~\ref{sec:measurement}) runs on the validation split and on every final selection. It uses identity-true within-video tracks from segmentation IDs and has no access to the cross-video match. For a temporal link, an oracle order matcher must answer at least 95\% of items with at least 95\% accuracy, and each of three spatial-only matchers must stay within 40--60\%: median-nearest position (balanced during selection), Video-1 last-seen to Video-2 first-seen position (endpoint), and Video-1 start to Video-2 parking position. The last two are held out from selection as independent checks. The oracle order matcher uses the scorer's tracker. For Video-2 strikes whose short post-contact interval misses the push rule, the oracle alone retries with a relaxed onset (0.3~m, sustained above 0.25~m); the video scorer is unchanged. Items the oracle cannot answer count against coverage. A cell that fails is quarantined whole, and final items are never dropped after model outcomes are seen.

Table~\ref{tab:gatecounts} reports the validity checks on all twelve evaluation selections. Layout and \synclane{} pass every selection. On the fresh held-out split (Appendix~\ref{app:confirm}), pairs, layout squares and nuisance pairs pass; \synclane{} nuisance pairs reach the 60\% endpoint bound exactly. Fresh \synclane{} squares fail (endpoint matcher 66.0\%) and were excluded before any model output. Relay also fails: the endpoint matcher reaches 62.0\% on both validation and final pairs, and order coverage is 75--94\%. Rebalancing endpoint outcomes after this failure would compromise the check's independence. Relay is therefore excluded from confirmatory pooling and reported as secondary evidence. This decision followed the failed checks and preceded any model outcome (Appendix~\ref{app:deviations}).

\begin{table}[!htbp]
  \caption{Link gates on the evaluation selections. Spatial: identity-true median-nearest / endpoint / start-to-parking accuracy on all $N$ branch items. Order: correct / answered / $N$, with coverage over all $N$. For layout, spatial correspondence is the intended link. Gate: order accuracy and coverage $\ge$95\%; every spatial $\max(a,1-a)\le$60\%.}
  \label{tab:gatecounts}
  \centering\scriptsize
  \setlength{\tabcolsep}{2.5pt}
  \begin{tabular}{@{}lrcrrc@{}}
    \toprule
    Selection & $N$ & Spatial (\%) & Order & Cov. & Gate \\
    \midrule
    Layout val. & 100 & 100 / 98.0 / 87.0 & --- & --- & pass \\
    Layout nuisance & 100 & 100 / 98.0 / 86.0 & --- & --- & pass \\
    Layout final pairs & 300 & 100 / 99.3 / 84.0 & --- & --- & pass \\
    Layout final squares & 200 & 100 / 98.0 / 82.0 & --- & --- & pass \\
    \midrule
    \synchead{} val. & 100 & 52.0 / 50.0 / 50.0 & 95/95/100 & 95\% & pass \\
    \synchead{} nuisance & 100 & 54.0 / 44.0 / 40.0 & 96/98/100 & 98\% & pass \\
    \synchead{} final pairs & 300 & 50.3 / 44.7 / 50.7 & 292/294/300 & 98\% & pass \\
    \synchead{} final squares & 200 & 55.0 / 42.0 / 50.0 & 196/198/200 & 99\% & pass \\
    \midrule
    Relay val. & 100 & 49.0 / \textbf{62.0} / 50.0 & 75/75/100 & \textbf{75\%} & fail \\
    Relay nuisance & 100 & 52.0 / 56.0 / 48.0 & 92/94/100 & \textbf{94\%} & fail \\
    Relay final pairs & 300 & 52.7 / \textbf{62.0} / 50.3 & 278/280/300 & \textbf{93\%} & fail \\
    Relay final squares & 200 & 51.0 / 56.0 / 50.0 & 188/188/200 & \textbf{94\%} & fail \\
    \bottomrule
  \end{tabular}
\end{table}

\begin{table}[!htbp]
  \caption{\textbf{Shortcut audit} on final pair items (accuracy, \%). For synchrony and relay, the gate checks the three spatial rules above the line and the order oracle's accuracy and coverage. Learned classifiers are additional audits. Order: accuracy (coverage); classifier: logistic (boosted). $^\ast$Added after the confirmatory runs.}
  \label{tab:links}
  \centering\footnotesize
  \setlength{\tabcolsep}{3pt}
  \begin{tabular}{@{}lccc@{}}
    \toprule
    Matcher & Layout & \Synclane{} & Relay \\
    \midrule
    Median position & 100 & 50.3 & 52.7 \\
    Endpoint & 99.3 & 44.7 & \textbf{62.0} \\
    Start $\to$ parking & 84.0 & 50.7 & 50.3 \\
    \midrule
    Order oracle & --- & 99.3 (98) & 99.3 (\textbf{93}) \\
    Time-synchronous$^\ast$ & 100 & 100 & no answer \\
    \midrule
    Link gate & pass & pass & fail \\
    Learned static classifier & --- & 79 (87) & 71 (84) \\
    \bottomrule
  \end{tabular}
\end{table}

\paragraph{Time-synchronous matcher.}
After the confirmatory runs, we added a fourth matcher. At each frame, each visible Video-2 cube votes for the nearest Video-1 cube visible at that frame (within 0.75~m). Using ground-truth identity tracks, it answers all 300/300 final \synclane{} items and all layout items correctly. It answers no relay item because no frame shows the same cube in both inputs. On \synclane{}, the same cube is visible in both videos for a median of six frames (maximum ten). Thus, \synclane{} permits matching at the same moment; it does not require tracking movement order. The time-shift test supports models' use of this evidence (Appendix~\ref{app:tshift}). Only relay requires carrying identity across a visibility gap.

\subsection{Held-out attribute: patch placement}
\label{app:placement}
A second carried attribute replaces patch color with vertical patch placement. Both cubes carry the same colored band, one high and one low, and cue-swapped squares exchange their placements. Apart from one direct-training test (Appendix~\ref{app:heldoutmore}), no model is trained on this attribute. Placement uses the layout and \synclane{} designs, the same selection rules, and its own frozen scorer (Appendix~\ref{app:metrics}). Its selections hold 150 ordinary pairs and 50 squares per link: 400 scene bases and 1,000 ground-truth targets. All four selections pass the link gate (spatial matchers 85.7--100\% on layout, 49.0--52.0\% on \synclane{}). \Synclane{} order accuracy is 295/297 (99.3\%) on pair items and 196/196 on square items, with coverage of 99.0\% and 98.0\%.

\section{Scorer Calibration and Human Checks}
\label{app:metrics}

\subsection{Calibration protocol}
Scorer code, color model, and thresholds are frozen and hashed before each independent calibration round, and any repair requires a new round on fresh, disjoint pairs. Earlier versions of both scorers were superseded during development; all reported scores use the final frozen color scorer (v8 in the main tables) and the final frozen placement scorer (placement v2). Each round renders fresh calibration pairs (about 50--60 per link), and each pair supplies two clean targets. Each clean target yields three invalid controls, with visibility rechecked after each edit: a wrong-actor control (the opposite branch's target scored against this branch's label), a missing-cue control (actor patch removed through its segmentation mask), and an invented-cue control (patch recolored to an unused color). Table~\ref{tab:calib3} reports the frozen color scorer on its final calibration round.

\begin{table}[!htbp]
  \caption{Independent calibration of the frozen color scorer (percent of $N$ targets). Clean: correct / abstain. False-correct: share of invalid controls scored as the original answer. Swap wrong: swapped-actor targets identified as the other candidate.}
  \label{tab:calib3}
  \centering\scriptsize
  \setlength{\tabcolsep}{3pt}
  \begin{tabular}{@{}lrccccc@{}}
    \toprule
    & & Clean & \multicolumn{3}{c}{False-correct} & Swap \\
    Link & $N$ & corr./abst. & swap & missing & invented & wrong \\
    \midrule
    Layout & 116 & 98.3 / 1.7 & 0.0 & 0.0 & 0.9 & 98.3 \\
    \Synclane{} & 106 & 95.3 / 4.7 & 0.0 & 2.8 & 0.0 & 95.3 \\
    Relay & 110 & 95.5 / 3.6 & 0.0 & 2.7 & 0.0 & 95.5 \\
    \midrule
    Gate & & $\ge$90 & $\le$5 & $\le$5 & $\le$5 & $\ge$90 \\
    \bottomrule
  \end{tabular}
\end{table}

\paragraph{Degradations.}
These are reported separately from the gate. On the same targets, correctness under RGB noise, blur, JPEG quality 35, downsampling, and color jitter is 97.4\%, 96.6\%, 90.5\%, 98.3\%, and 49.1\% for layout, and 95.3\%, 94.3\%, 85.8\%, 95.3\%, and 33.0\% for \synclane{}. The scorer is not robust to color jitter, so generated outputs with global hue shifts may be misscored.

\paragraph{Placement scorer.}
The placement scorer is a separate frozen snapshot that reads which band, upper or lower, the actor carries; ambiguous middle-band frames cast no vote, and the color scorer is unchanged. Its first frozen version passed independent calibration on layout but failed on \synclane{} (82.5\% clean correctness, 83.3\% coverage on 120 targets). In 12 of 13 patch-related abstentions the answer was on the lower band, which a $5\times5$ silhouette erosion removed in the small \synclane{} target views. Removing the erosion in placement mode only, and recalibrating on fresh pairs, passes on both links (clean correctness 99.5\% on layout and 94.0\% on \synclane{}; false-correct rates at most 4.6\%). No placement evaluation output had been inspected when this repair was made. The repair raises missing-cue false-correct rates toward the 5\% ceiling. Under the same degradations as above, layout / \synclane{} correctness is 97.7/95.0\%, 96.7/89.0\%, 90.7/88.5\%, 98.1/92.7\%, and 62.6/50.9\%.

\subsection{Metrics}
\label{app:switchdef}
Let $\hat Y_{i,b}$ be the generated target of branch $b$ of ordinary pair $i$, $r_{i,b}$ its correct patch ($r_{i,0}\ne r_{i,1}$), and $q(\cdot)$ the frozen scorer's reading. Switch accuracy over $N$ pairs is $A_{\rm switch}=\frac1N\sum_{i}\mathbf 1[q(\hat Y_{i,0})=r_{i,0}\land q(\hat Y_{i,1})=r_{i,1}]$, and item accuracy uses all $2N$ branch outputs, $A_{\rm item}=\frac{1}{2N}\sum_{i,b}\mathbf1[q(\hat Y_{i,b})=r_{i,b}]$. Coverage is the share of outputs the scorer answers, and answered accuracy is accuracy among those outputs. Actor-region PSNR uses a fixed 16-pixel dilation of the ground-truth actor and event mask, for evaluation only. Intervals are 95\% paired bootstrap intervals over scene bases (10,000 resamples, link-stratified, branches kept together) unless stated otherwise.

\subsection{Human checks}
\label{app:judgecheck}
\emph{Rating sets.} Two human studies were specified with at least three raters (Appendix~\ref{app:deviations}). The second rating set, fixed before any rating, covers the method arms on the same 40 held-out first-split pairs (20 per link, both branches, seed~0 of each model), with repeated ground-truth checks; a part is excluded if fewer than 8 of its 10 checks are right, and none was. Raters are blind and independent of the authors, and answer options are shuffled per item. Table~\ref{tab:raterscorer} gives switch accuracy at 10,000 updates for both raters; Min-SNR and swap + Min-SNR were rated in a later session than the other three arms. The 40 pairs are a random draw, and on them the scorer happens to show no noisy-input swap gain at 10,000 updates (42.5\% for both arms), whereas on all 300 held-out pairs it gives seed~0 $+18.3$ [$+12.0$, $+24.3$].

\emph{Agreement.} The raters give the same four-way label on 97.5\% of the 239 ordinary, LR$\times$2 and noisy-input swap outputs both answered (Cohen's $\kappa$ 0.91) and on 95.6\% of the 160 Min-SNR and swap + Min-SNR outputs ($\kappa$ 0.69; all 80 swap + Min-SNR outputs identical). Against the scorer, $\kappa$ is 0.19--0.47 per model at 10,000 updates (uninformative for swap + Min-SNR, which both raters call right on every output); where both decide, each rater agrees with the scorer on 88--95\% of outputs, and each confirms 273 of 283 scorer-right outputs (96.5\%) pooled over the five models.

\emph{Contrasts} (pair bootstrap stratified by link; averaged over raters at pair level). Noisy-input swap leads ordinary training by $+21.2$ [$+7.5$, $+36.2$] (scorer $0.0$) and LR$\times$2 by $+11.5$ [$-2.6$, $+26.9$] (39 pairs; first rater $+12.8$ [$+0.0$, $+28.2$], second $+7.5$ [$-7.5$, $+22.5$]). Descriptively, the latter is $+20.0$ [$+5.0$, $+40.0$] on layout and $+2.6$ [$-21.1$, $+26.3$] on \synclane{} (19 pairs), so on these pairs the raters do not show the \synclane{} gain over LR$\times$2 that the scorer shows on all held-out pairs. The pre-specified primary for the combination packet, swap + Min-SNR minus LR$\times$2, is $+30.8$ [$+17.9$, $+44.9$] (39 pairs; scorer $+38.5$; first rater $+33.3$ [$+17.9$, $+48.7$], second $+27.5$ [$+15.0$, $+42.5$]). Swap + Min-SNR leads noisy-input swap by $+21.2$ [$+10.0$, $+32.5$] and Min-SNR by $+30.0$ [$+16.2$, $+43.8$]; only the last compares arms rated in one session. Min-SNR minus LR$\times$2 is $0.0$ [$-14.1$, $+14.1$] (scorer $+17.9$; descriptive, across sessions). Both raters had rated these scenes for other arms before the combination packet, and swap + Min-SNR sits at a ceiling (40 of 40 pairs for each rater; exact 95\% lower bound 91.2\%).

\emph{Later checkpoint.} One rater rated the 17,500-update parts (410 items, all checks passed). Human-scored switch accuracy is 47.5\% for Two-video (5,000 updates), 77.5\% for Two-video-long, and 80.0\% for ordinary training, LR$\times$2 and noisy-input swap, against 25.0, 57.5, 62.5, 62.5 and 52.5\% for the scorer (Cohen's $\kappa$ 0.27--0.49); Two-video-long leads the 5,000-update model by $+30.0$ [$+12.5$, $+47.5$] (scorer $+32.5$). Over all of this rater's parts, agreement where both answer is 91.7\% of 518 outputs and 96.1\% of 434 scorer-right outputs are confirmed; all 20 checks and 10 repeats pass.

\emph{Where the scorer under-credits noisy-input swap.} On the 40 pairs at 10,000 updates, the outputs the rater calls right and the scorer does not number 14, 17 and 21 for ordinary training, the doubled learning rate and noisy-input swap. Half or more are event-detection abstentions, where the scorer finds no contact, no push or an ambiguous event (7, 13 and 14); the others are read as another palette color (0, 2 and 5), as an invented cue (4, 1 and 2) or as an unseen patch (3, 1 and 0). Across all held-out pair outputs, event detection does not differ between these arms (Table~\ref{tab:methodheldoutfull}: 89.2\% for noisy-input swap, 89.3\% for LR$\times$2), so these counts describe the rated pairs and do not explain the gap in general.

\paragraph{Scorer check on generated videos.}
The first rating set holds 396 generated ordinary-pair outputs from six earlier model conditions (Two-video seeds 0--2, +InputHead, ReCamMaster and the modular reference), sampled by scorer category, with ground-truth attention checks; the rater sees only the target video and the question, with the two candidate cues, ``different patch color'' and ``cannot tell'' as options. On the 192 outputs rated (8 of 8 checks passed), agreement where both answer is 104 of 126 (82.5\%; 75--89\%) and the rater confirms 57 of 68 scorer-right outputs (83.8\%). The scorer is conservative: of 97 outputs the rater judges right, it calls 57 right and abstains on 26. Some scorer credits are rejected as well (11 of 68), so automatic switch rates are not strict lower bounds on human judgments. No scorer threshold or model score was revised after the human check.

\section{Probes: When Is Identity Decided?}
\label{app:decision}
The probes compare evidence from the noisy input with evidence from the conditioning inputs, across released models and fine-tuned checkpoints. Synthetic probes use training and held-out squares; real-video and image-editing probes use the paired edits specified below. A copy probe uses one denoiser call, while a decision probe runs the sampler to completion from a planted answer.

\paragraph{Method-study checkpoints.} Table~\ref{tab:copyprobe256} gives the copy probe for every 256-pixel arm probed after 10,000 updates. Below $\sigma=5$, only the two changes that corrupt the answer at low noise, noisy-input swap and the midpoint, stop the models from copying the planted patch (the midpoint still copies at $\sigma=1$, where its input is nearly the average of the two candidates). Schedule shifts, extra pure-noise draws, swapping only at $\sigma\ge20$ and Min-SNR copy as often as ordinary training; patch weighting copies slightly less at $\sigma=2$--$3.3$ and as often at $\sigma\le1$. Above $\sigma\approx8$ every arm follows Video~1 on 33--42 cells.

\begin{table}[!htbp]
  \caption{Copy probe for the 256-pixel method-study checkpoints after 10,000 updates (one run of the probe for all of them): training cells (of 42) whose prediction follows Video~1 when the noisy input carries the sibling's patch, by noise level.}
  \label{tab:copyprobe256}
  \centering\scriptsize
  \setlength{\tabcolsep}{1.8pt}
  \begin{tabular}{@{}lc ccccccccccc@{}}
    \toprule
    Training & Seed & 0.5 & 1 & 2 & 3.3 & 5 & 8 & 13 & 20 & 40 & 80 & 200 \\
    \midrule
    Ordinary & 0 & 0 & 0 & 12 & 29 & 38 & 39 & 39 & 40 & 39 & 38 & 37 \\
     & 1 & 0 & 0 & 13 & 29 & 36 & 40 & 42 & 42 & 40 & 36 & 36 \\
    Ordinary, LR$\times$2 & 0 & 0 & 0 & 11 & 31 & 38 & 39 & 40 & 40 & 39 & 37 & 37 \\
     & 1 & 0 & 0 & 17 & 35 & 38 & 39 & 40 & 40 & 37 & 35 & 34 \\
    Noisy-input swap & 0 & 0 & 40 & 41 & 41 & 41 & 42 & 42 & 40 & 41 & 39 & 39 \\
     & 1 & 1 & 42 & 42 & 41 & 41 & 41 & 41 & 42 & 41 & 40 & 39 \\
    Swap only at $\sigma\ge20$ & 0 & 0 & 0 & 10 & 27 & 34 & 38 & 39 & 39 & 40 & 35 & 35 \\
     & 1 & 0 & 0 & 10 & 27 & 35 & 39 & 40 & 40 & 38 & 39 & 38 \\
    Midpoint & 0 & 0 & 2 & 37 & 40 & 41 & 42 & 42 & 40 & 39 & 37 & 36 \\
     & 1 & 0 & 11 & 38 & 41 & 41 & 41 & 41 & 41 & 39 & 36 & 34 \\
    Schedule shift $\sigma\times2$ & 0 & 0 & 0 & 12 & 29 & 36 & 39 & 40 & 39 & 39 & 38 & 38 \\
     & 1 & 0 & 0 & 11 & 27 & 34 & 38 & 38 & 40 & 36 & 35 & 34 \\
    Schedule shift $\sigma\times4$ & 0 & 0 & 0 & 14 & 32 & 39 & 40 & 40 & 40 & 39 & 37 & 37 \\
     & 1 & 0 & 0 & 9 & 27 & 35 & 39 & 40 & 39 & 36 & 33 & 33 \\
    5$\times$ more pure-noise draws & 0 & 0 & 0 & 12 & 25 & 36 & 40 & 40 & 38 & 36 & 35 & 34 \\
     & 1 & 0 & 0 & 16 & 35 & 40 & 41 & 42 & 41 & 40 & 39 & 40 \\
    Min-SNR-5 & 0 & 0 & 0 & 16 & 30 & 39 & 40 & 40 & 41 & 41 & 39 & 39 \\
     & 1 & 0 & 0 & 16 & 31 & 38 & 39 & 40 & 41 & 41 & 37 & 36 \\
    Patch loss weight & 0 & 0 & 0 & 20 & 37 & 41 & 42 & 42 & 42 & 41 & 41 & 40 \\
     & 1 & 0 & 0 & 23 & 37 & 40 & 42 & 42 & 41 & 39 & 40 & 40 \\
    \bottomrule
  \end{tabular}
\end{table}

\paragraph{Real video.} On 40 Argoverse~2 vehicles and 40 people from CMU Panoptic social games~\cite{joo2019panoptic}, each clip selected by a rule fixed before scoring, we rotate the object's hue by 150$^\circ$ and add 40 chroma units inside a feathered box (luminance kept), and plant this edit in the noisy input of the released models, conditioned on the real source video. Released AnyView copies the edit up to $\sigma\approx2.25$ on vehicles and $\sigma\approx5.35$ on people (above 65\% of its training draws); ReCamMaster copies it up to $\sigma\approx4.57$ on vehicles. In a reverse control that recolors the source and plants the real color, AnyView copies up to $\sigma\approx6.1$, so at low noise it copies whatever the noisy input shows. Given another clip's source video, AnyView's output changes, so it uses the source; ReCamMaster's does not.

\paragraph{Instruction-guided image editing.} The probe also runs on InstructPix2Pix~\cite{brooks2023instructpix2pix} (Stable Diffusion 1.5, timesteps uniform over 1,000) on $512\times512$ crops of the 40 vehicles, with the instruction ``make the car red'' (or blue) and the other color planted in the noisy latent. The editor copies the planted color up to $t^\ast=777$ ($\sigma\approx4.7$ in EDM units), 78\% of its training draws; halving the answer difference moves the crossover to $t^\ast=617$ ($\sigma\approx2.4$), as the two-hypothesis account below predicts. With an empty instruction the projection never favors the instructed color, so the positive scores at high noise come from the instruction.

\paragraph{Second architecture.} The same copy probe runs on ReCamMaster (Wan2.1 flow matching), with the same 42 cells, scores and per-cell noise seeds. The noisy input is $x_s=(1-s)z+s\epsilon$, mapped to $\sigma=s/(1-s)$; both models encode with the same Wan2.1 autoencoder (AnyView's Cosmos-Predict2 tokenizer is this autoencoder, used with $\sigma_{\rm data}=1$), so their latents and their ideal-observer threshold coincide, $\sigma^\ast\approx54$ (10th--90th percentile over cells 44--63) for ReCamMaster, with 8.5\% of its training draws above it. %
Every checkpoint follows the planted patch on at least 39 of 42 cells up to $\sigma\approx2$, as AnyView does.

\paragraph{How much does the noisy input reveal?}
Let $z$ and $z_{\rm sib}$ be the target latents of a square cell and of its Video-1 sibling, which differ only in the actor's patch. Given the noisy input $y=z+\sigma\epsilon$, an ideal binary classifier that knows both candidates picks the right one with probability $\Phi\bigl(\lVert z-z_{\rm sib}\rVert/2\sigma\bigr)$. The threshold $\sigma^\star$ is the noise level at which this accuracy falls to 60\%: $\sigma^\star=\lVert z-z_{\rm sib}\rVert/(2\cdot0.2533)$. On the 42 probe cells, $\sigma^\star$ has median 158 (10th--90th percentile 132--187) at the 576-pixel training resolution. Under the released noise distribution (Appendix~\ref{app:models}), only the 5\% of draws taken from the high-noise range exceed it. In the remaining 95\% of ordinary updates, the patch is in principle readable from the noisy input.

\paragraph{Copy probe.}
The noise level above which most cells follow Video~1 is no higher for \synclane{} than for layout (Two-video-long: $\sigma\approx3.3$ against 8), so the leak is not specific to \synclane{}.

For 42 training cells (14 per link), each paired with its Video-1 sibling, we keep a cell's own inputs but build its noisy input from the sibling, $y=z_{\text{sib}}+\sigma\epsilon$. We then project the predicted clean latent on the sibling difference. Figure~\ref{fig:leak} (right) counts cells whose prediction leans to the cell's own, Video-1-consistent patch. At low noise most predictions follow the planted patch. The noise level above which predictions follow Video~1 falls with training. Between $\sigma{=}8$ and $20$, the 17,000-update model follows Video~1 on 33--41 of 42 cells, against 16--31 for the 5,000-update model.

On 42 held-out cells from test scenes, built the same way, the crossover of the mean projection lies at $\sigma\approx12.2$ for the 5,000-update model and 8.4 and 6.4 for the long run after 10,000 and 17,000 updates (training cells: 10.9, 6.7 and 5.4), 86, 78 and 71\% of training draws; no held-out cell follows Video~1 at $\sigma\le2$.

At 256 pixels, ordinary training crosses at $\sigma\approx3.4$ and 3.5 on held-out cells (2.7 and 2.8 on training cells) and noisy-input swap at 0.8 and 0.7 (0.8 and 0.7). Every held-out crossover lies within a factor of 1.3 of its training-cell value, so copying is not an artifact of memorized training scenes. The models thus stop using the target's patch information far below $\sigma^\star$.

\begin{figure}[!htbp]
  \centering
  \input{figures/probe_style}
  \input{figures/decision_probe}
  \caption{Decision probe on 86 \synclane{} training cells (576 pixels; Two-video after 5k updates, Two-video-long after 17k): sampling starts from the sibling's target noised to $\sigma_s$; share of answered outputs depicting the correct, Video-1-consistent patch (bars: Wilson 95\% intervals).}
  \label{fig:decisionprobetrain}
\end{figure}

\paragraph{Decision probe.}
Here the sampler starts from the sibling's target noised to $\sigma_s$, so the wrong patch is planted at the start, and runs to completion. The finished outputs of 86 \synclane{} training cells are scored with the frozen scorer. Figure~\ref{fig:decisionprobetrain} reports the share of answered outputs that depict the cell's own patch. In counts, the 5,000-update model gives 6/67, 17/63, 25/60, and 35/64 for $\sigma_s=10, 20, 40, 80$; the 17,000-update model gives 33/62, 56/67, 67/74, and 66/69. The long-trained model overrides a planted wrong identity from moderate noise on. The short model keeps it, and is near chance even from pure noise.

\paragraph{Decision probe on held-out squares.}
On the 200 cells of the held-out \synclane{} squares (256-pixel models, seed~0, 10,000 updates), the share of answered outputs that override the planted patch at $\sigma_s=10/20/40/80$ is 57.6, 65.7, 69.0 and 75.0\% for ordinary training, 92.3, 89.2, 90.4 and 94.0\% for noisy-input swap (whose training inputs resemble this probe, so this shows what it learned without testing the mechanism independently), 85.7, 91.1, 92.0 and 95.2\% for patch weighting and 78.4, 83.4, 87.4 and 90.5\% for Min-SNR. Together with the copy probe, where both reweighting arms still copy below $\sigma\approx3$ (Table~\ref{tab:copyprobe256}), this suggests that reweighting lets the model settle the identity at moderate noise, after which copying the visible patch is harmless, whereas noisy-input swap also stops the copying itself.

\paragraph{Use of Video~1 per noise level.}
A pre-specified probe measures how much each model's prediction depends on Video~1 at each noise level. On the 42 training-square cells of the copy probe, each with its own noisy input and noise draw, Video~1 is replaced by the sibling's, which shows the other patch assignment, and $D(\sigma)=s(\text{own Video 1})-s(\text{sibling's Video 1})$, where $s$ is the copy-probe projection of the predicted patch on the answer difference ($D=1$: the patch follows Video~1 fully; $D=0$: Video~1 is ignored). Table~\ref{tab:v1dep} gives $D$ and the rise of the answer-region error when Video~1 is replaced, for the 256-pixel, 10k checkpoints of ordinary training and noisy-input swap (seeds 0--2). At $\sigma=1$, noisy-input swap's prediction depends on Video~1 four times as much as ordinary training's, and every noisy-input swap seed lies above every ordinary seed up to $\sigma=5$; from $\sigma=13$ on, the arms are close. Ordinary training's dependence peaks at $\sigma=5$ and falls at lower noise, where its models copy.

\begin{table}[!htbp]
  \caption{Use of Video~1 per noise level on 42 training-square cells (256 pixels, 10,000 updates; mean over seeds 0--2, range in brackets). $D$: change of the predicted patch, as a fraction of the answer difference, when Video~1 is replaced by the sibling's. MSE: rise of the answer-region error ($\times10^{3}$).}
  \label{tab:v1dep}
  \centering\footnotesize
  \setlength{\tabcolsep}{4pt}
  \begin{tabular}{@{}r cc cc@{}}
    \toprule
    & \multicolumn{2}{c}{$D$} & \multicolumn{2}{c}{MSE} \\
    \cmidrule(lr){2-3}\cmidrule(lr){4-5}
    $\sigma$ & Ordinary & Noisy-input swap & Ordinary & Noisy-input swap \\
    \midrule
    1 & 0.09 {\scriptsize(0.08--0.11)} & 0.35 {\scriptsize(0.31--0.39)} & 2.6 & 11.9 \\
    2 & 0.17 {\scriptsize(0.15--0.22)} & 0.39 {\scriptsize(0.36--0.44)} & 5.7 & 15.5 \\
    5 & 0.25 {\scriptsize(0.22--0.29)} & 0.35 {\scriptsize(0.33--0.38)} & 9.9 & 14.9 \\
    13 & 0.21 {\scriptsize(0.19--0.24)} & 0.25 {\scriptsize(0.24--0.26)} & 8.9 & 11.1 \\
    40 & 0.13 {\scriptsize(0.11--0.16)} & 0.15 {\scriptsize(0.14--0.15)} & 6.1 & 6.9 \\
    \bottomrule
  \end{tabular}
\end{table}

\paragraph{An incentive to ignore conditioning.}
Copying at low noise can be optimal under ordinary denoising: when $y$ reveals the answer, consulting Video~1 offers little additional loss reduction. A binary model formalizes this incentive, but does not describe learning dynamics. Its predicted benefit from using Video~1 fails to rank held-out training variants (Appendix~\ref{app:decision}). We therefore treat copying as a diagnostic and possible contributor to slow learning, not an established cause.

\paragraph{A two-hypothesis account of the leak.}
Consider equally likely target latents $z_0,z_1$ with $\Delta=z_1-z_0$, and assume an ideal reader knows both candidates. Under ordinary corruption $y=z_A+\sigma\epsilon$, let $q(y)=P(A=1\mid y)$ and $h(d)=\mathbb E[q(1-q)]$, with $d=\|\Delta\|/(2\sigma)$. A reader without Video~1 has answer-component Bayes risk $\|\Delta\|^2h(d)$. Perfect conditioning would remove this risk, giving a potential loss reduction
\begin{equation}
G_{\rm ord}(\sigma)=w(\sigma)\,\|\Delta\|^2 h(d).
\label{eq:reward}
\end{equation}
This compares risks under a binary idealization; it does not describe actual gradients or training speed.

\paragraph{Randomized sibling corruption.}
\label{app:randomswap}
Let $B\sim\mathrm{Bernoulli}(p)$ be independent of the true answer $A$, and construct $y=z_C+\sigma\epsilon$ with $C=A\oplus B$. Equal answer priors give the same marginal mixture for $y$ as ordinary corruption. Writing $q(y)=P(C=1\mid y)$, the posterior of the true answer is $p+(1-2p)q(y)$. Its expected variance is therefore
\begin{equation}
\begin{aligned}
\mathbb E[\mathrm{Var}(A\mid y)]
&=p(1-p)+(1-2p)^2h(d),\\
G_p(\sigma)&=w(\sigma)\|\Delta\|^2\mathbb E[\mathrm{Var}(A\mid y)].
\end{aligned}
\label{eq:randomswap}
\end{equation}
At low noise the Bayes risk without Video~1 approaches $p(1-p)\|\Delta\|^2$. At $p=0.5$, the noisy candidate is independent of the true answer; at $p=1$, an ideal reader can invert it. Thus the error of \emph{copying on a swapped draw}, approximately $\|\Delta\|^2$, is not the same as the optimal risk under the randomized mixture. The experiment applies swapping only to eligible square cells at $\sigma\ge1$ and also trains ordinary samples, so this subproblem does not predict the full recipe's behavior at $p=1$. This clarification was derived after the experiments; the predictive test below used the original heuristic proxy, not Eq.~\ref{eq:randomswap}.

\paragraph{Comparing noise bands with MemCorr-DP.}
MemCorr-DP~\cite{su2026memcorr} samples paired timesteps 80--99 of a 100-step schedule, with $\beta_t=10^{-4}+t(0.02-10^{-4})/99$. Using $\bar\alpha_t=\prod_{j=0}^t(1-\beta_j)$ and $\sigma_t=\sqrt{(1-\bar\alpha_t)/\bar\alpha_t}$ gives $\sigma_{80}\approx0.970$ and $\sigma_{99}\approx1.323$. Its highest timestep ranks are therefore not equivalent to CrossSwitch's $\sigma\ge20$ band. Target normalization and answer separation also differ across action and video representations. Our band ablations characterize the CrossSwitch schedule; they are not a noise-matched reproduction of MemCorr-DP.

\paragraph{A quantitative test of the original reward proxy.}
A pre-specified test asked whether an original heuristic reward proxy, the predicted reward for reading Video~1 computed from each arm's noise distribution, loss weight and shares of draws that contradict or erase the answer, ranks the training variants. It ranks the 18 validation variants of Table~\ref{tab:mechanismfull} (Spearman's $\rho=+0.70$ [$+0.29$, $+0.90$], effective reader) but fails the pre-specified held-out test across 15 arms ($\rho=+0.07$ [$-0.50$, $+0.65$]), mainly because it gives the Min-SNR arms the smallest reward although they are among the most accurate.

\section{Training, Inference, and Compute}
\label{app:models}

\paragraph{Generators.}
We evaluate two released multi-video models, LiveStre4m and NoPo4D, zero-shot, and adapt the released single-video AnyView model into the Two-video generator. Neither multi-video model was trained for our wide-baseline cameras: LiveStre4m is fine-tuned on four Neural3DVideo scenes with nearby forward-facing cameras, and NoPo4D on Ego-Exo4D rigs. AnyView's training mixture includes Kubric-4D and Kubric-5D, which makes it the most in-domain starting point.

\paragraph{Two-video training.}
All Two-video models start from the released AnyView checkpoint and train on the same 1,742 bases (Appendix~\ref{app:data}) with the standard AnyView denoising loss and learning rate $4\times10^{-6}$ unless stated otherwise. Video~1 enters as a third view-token stream. Ordinary training denoises every cell of the training squares, so no method receives extra target videos.

Training noise levels follow the released configuration: $\sigma=\sqrt{11}\,e^{n}$ with $n\sim\mathcal N(0,1)$ (median $\approx$3.3), except that 5\% of samples draw $\sigma$ log-uniformly in $[200,10^5]$; one $\sigma$ is drawn per sample. Each learning-rate schedule spans the run's full length, so a checkpoint inside a longer run is not equivalent to a shorter run's final model.

A run interrupted by the cluster resumes from its last saved checkpoint with the original schedule and update counter and restarted AdamW moments; updates after that checkpoint are discarded. In the method study's primary arms, only seed~0 of ordinary training and of noisy-input swap at the base learning rate were interrupted once each and resumed from a saved checkpoint (at 6,500 and 5,500 updates); all other primary runs, including every doubled-learning-rate run, trained without interruption, and the schedule continued unchanged at each resume. Two-video-long seeds 0, 1 and 2 resumed one, three and two times. Table~\ref{tab:compute} summarizes all trained models.

The models of Section~\ref{sec:baseline} (Table~\ref{tab:core}) train at 576 pixels, AnyView's released resolution, on the 256-pixel renders upsampled bilinearly, so that the diagnosis uses the model in its native regime; the method study instead trains at the data's 256 pixels, about seven times faster per update. Before the method study, a tokenizer round trip of ground-truth validation targets passed a pre-set gate of at most 5 points below 576 pixels (layout 47/50 switches at both resolutions; \synclane{} 45/50 at 256 against 42/50 at 576).

\begin{table*}[!tp]
  \caption{Trained models. Updates: evaluated checkpoint and, in parentheses, the length of its learning-rate schedule. Alternating runs interleave denoising and auxiliary updates in equal number. Seconds per update are logged training speeds (H: H100/H200, A: A100).}
  \label{tab:compute}
  \centering\footnotesize
  \setlength{\tabcolsep}{4pt}
  \resizebox{\textwidth}{!}{\begin{tabular}{@{}llllll@{}}
    \toprule
    Model & Res. & Updates & LR & Seeds & s/update \\
    \midrule
    Two-video & 576 & 5,000 (5,000) & $4\times10^{-6}$ & 0, 1, 2 & 13.2 (H) \\
    +InputHead & 576 & 10,000 alternating (10,000) & $4\times10^{-6}$ & 0 & 21.5 (A) \\
    +CF & 576 & 10,000 alternating (10,000) & $4\times10^{-6}$ & 0 & 38.3 (A), 22.4 (H) \\
    Two-video-long & 576 & 17,000 (18,000) & $4\times10^{-6}$ & 0, 1, 2 & 22.7 (A), 13.2 (H) \\
    Noisy-input swap, 576 & 576 & 10,000 (18,000) & $4\times10^{-6}$ & 0, 1 & 13.6 (H) \\
    ReCamMaster & --- & 5,000 (constant LR) & $10^{-5}$ & 0 & 1.81 (A) \\
    ReCamMaster, longer & --- & 17,000 (constant LR) & $10^{-5}$ & 0, 1 & 1.08--1.19 (H) \\
    ReCamMaster, patch w.\ / noisy-input swap & --- & 17,000 (constant LR) & $10^{-5}$ & 0, 1 & 1.19 / 1.13 (H) \\
    \midrule
    Method study (all arms) & 256 & 10,000 and 17,500 (18,000) & $4\times10^{-6}$; controls $8\times10^{-6}$, $1.2\times10^{-5}$ & 0--2 & 1.98 (H) \\
    \bottomrule
  \end{tabular}}
\end{table*}

\paragraph{Compute matching.}
Two-video-long's evaluated checkpoint, 17,000 updates, was fixed before any output was scored as the save whose training time is closest to that of +CF (Appendix~\ref{app:other}): 107.2 against 107.0 device-hours.

\paragraph{Other training.}
+CF is described in Appendix~\ref{app:other}. ReCamMaster conditions a camera-controlled video DiT on a source video by concatenating clean source latents with the noisy input latents along the frame axis. We extend it to two inputs as for AnyView: the streams are target, Video~2, and Video~1, and each stream's tokens receive their camera pose relative to Video~2 through ReCamMaster's camera encoder, widened with two zero-initialized focal-length inputs. It is fine-tuned with its own recipe (flow matching, target-only loss, trainable self-attention and camera layers). A separate update-matched run trains it for 17,000 updates with the same constant learning rate.

\paragraph{Batch size and passes.} Every Two-video and method-study run uses one GPU and one training item per update (batch size 1, following the AnyView release recipe). The training split contains 5,168 items: 1,800 pair outputs and 3,368 square cells. Thus, 10,000 updates correspond to about two passes over the data and 17,000 to about 3.3 passes. Validation accuracy in the long ordinary run is still rising at its final checkpoint (Figure~\ref{fig:emergence}); no run is trained to convergence.

\paragraph{Inference.}
Two-video generation uses 41 input frames, 35 denoising steps from $\sigma_{\max}=80$, an empty prompt, and seed 0; all outputs of one model share the starting noise. Outputs are deterministic given checkpoint, starting noise and GPU model. The fresh-split, shared-rate, time-shift, answer-size and split-evidence tests are each generated on one GPU model. Other evaluations drew each arm from the same pool of GPUs (first split and 576 pixels: H100 and H200; direct placement, the real-video garment test and the balanced set's 576-pixel arms also A100). Regenerating ten first-split runs on a second GPU model changed per-link switch accuracy by 1.8 points on average and at most 7.3, with no systematic shift; whether H100 and H200 outputs differ was not tested.

ReCamMaster samples with 50 steps, an empty prompt, and seed 0. Each evaluated model generates 900 ordinary-pair outputs, 600 square cells, and 300 nuisance outputs across the three links. A shuffled-Video-2 control covers 50 pairs and 50 squares per link. Because shuffling creates inconsistent scene evidence, its coverage and renderability are reported rather than read as identity use. Placement adds 600 pair outputs and 400 square cells per model.

\section{Results in Main-Text Order}
\label{app:results}
The subsections follow the main text: the baseline (Section~\ref{sec:baseline}), learning speed, reweighting and transfer (Sections~\ref{sec:speed} and~\ref{sec:reweighting}), with the mechanism analyses they summarize, and real video (Section~\ref{sec:realvideo}).

\subsection{Static cues: the classifier-balanced set}
\label{app:worstgroup}
Beyond the gate's spatial matchers, a logistic classifier on 24 positional quantities from the identity-true tracks (positions only, without temporal matching; trained on training scenes, regularization chosen by grouped cross-validation) predicts the actor on 79.3\% [74.4, 83.5] of final \synclane{} pairs, and boosted trees on the same inputs on 87.3\%, mostly because the cube whose Video-1 track passes closest to the actor's Video-2 track is usually not the actor. A test set balanced against both was therefore built on fresh scenes.

\emph{Classifier-balanced \synclane{} set.} Pre-specified on October 26 and committed to the project repository before any scene was designed: fresh \synclane{} pairs from the held-out split's generator and rendering requirements (a new seed block, 640 designs per event family), of which 831 valid pairs recover the actor through the intended link on both branches. The two classifiers above (the same fitted models, which reproduce 79.3 and 87.3\% on the final pairs) pick the actor on 75.8 and 81.4\% of this pool. For each event family we select 75 pairs, with counts per (logistic, boosted) correctness class as close as possible to the pool's proportions, subject to both classifiers being correct on exactly half of the items (integer program; random draw within classes).

On the 150 selected pairs (300 items) both classifiers score 50.0\% [44.4, 55.6], the three spatial matchers 45--55\%, and the intended link 100\%; the frozen scorer switches on 88.7\% of the ground-truth renders.

The 256-pixel arms (three seeds each; two for Min-SNR and its combination with noisy-input swap) are generated on one GPU model (L40S). After 10,000 updates, switch accuracy is 34.7\% for ordinary training (seeds 31.3, 28.0, 44.7), 44.0\% at the doubled learning rate (46.7, 42.7, 42.7), 53.1\% for noisy-input swap (54.7, 47.3, 57.3), 56.7\% for Min-SNR (57.3, 56.0) and 70.7\% for noisy-input swap with Min-SNR (71.3, 70.0), the same order as on the held-out splits, and all above the flat 25\% and above guessing among readable outputs (13.4--19.2\%; both outputs answered on 53.8--77.0\% of pairs).

Both pre-specified 256-pixel readings are positive: noisy-input swap minus the doubled learning rate, $+9.1$ [$+3.6$, $+14.4$] (seeds resampled: [$-0.4$, $+17.8$]), and noisy-input swap with Min-SNR minus noisy-input swap, $+17.6$ [$+11.9$, $+23.2$] ([$+8.9$, $+26.4$]). Over ordinary training, the doubled learning rate gains $+9.3$ [$+4.9$, $+13.8$], Min-SNR $+22.0$ [$+16.7$, $+27.3$] and the combination $+36.0$ [$+30.3$, $+41.6$].

After 17,500 updates noisy-input swap reaches 64.4\% (65.3, 62.0, 66.0) and the doubled learning rate 60.0\% (62.0, 61.3, 56.7), a difference of $+4.4$ [$-0.7$, $+9.8$] (seeds resampled: [$-2.9$, $+11.8$]; evaluated as pre-specified, but not one of the readings): as on the held-out splits, the lead is largest at 10k.

The third pre-specified reading, at 576 pixels, is positive too: Two-video switches on 18.0\% after 5,000 updates (seeds 18.7, 14.7, 20.7; guessing among readable outputs 12.6\%), Two-video-long on 40.9\% after 10,000 (38.0, 44.0, 40.7) and 59.8\% after 17,000 (61.3, 58.0, 60.0); late learning thus replicates where the static classifiers are at chance, $+41.8$ [$+35.6$, $+48.2$] for 17,000 against 5,000 updates (seeds resampled [$+33.8$, $+49.8$]; 10,000 against 5,000: $+22.9$ [$+17.1$, $+28.7$]). All three readings are positive.

\subsection{Time-shift test}
\label{app:tshift}
Pre-specified on October 27 and committed to the project repository before any gate or generation output, and amended once after the gate and before any generation. To test whether \synclane{} switching rests on same-moment correspondence, Video~1 is delayed by seven frames (frame $f$ shows original frame $f-7$, the first seven frames repeating frame 0), applied to the decoded input before the usual resampling; Video~2, the target and the answer are unchanged. The items are the first split's final pairs, 150 per link. On their ground-truth renders under the delay, the time-synchronous matcher is right on none of the 300 \synclane{} items (contradicted on 180, no same-moment evidence on 120), while the departure order (100\%) and all static cues are unchanged and the onset matcher becomes more informative (59.3 to 95.3\%); on layout the space matcher stays at 100\% with full coverage, and in both links every patch remains visible on at least eight frames before the branch.

Five runs after 10,000 updates (256 pixels: seed~0 of ordinary training at LR$\times$2, noisy-input swap and Min-SNR, and two seeds of noisy-input swap with Min-SNR) generate both conditions with seed 0 on one GPU model (L40S). The undelayed outputs are regenerated, because a check of 20 items reproduced the earlier outputs' decisions on only 18 (19 required).

The primary statistic, for noisy-input swap with Min-SNR (pair switches averaged over its two seeds), is $D=\mathrm{Drop}_{\text{sync}}-\mathrm{Drop}_{\text{layout}}$, where a link's drop is its switch accuracy without the delay minus with it; the interval is a paired bootstrap over pairs, resampling each link separately.

\begin{table}[!htbp]
  \caption{Time-shift test (switch accuracy, \%, without the delay $\to$ with Video~1 delayed by seven frames; 150 pairs per link). $D$: \synclane{} drop minus layout drop, 95\% paired bootstrap interval. Swap + Min-SNR: two seeds; other arms: one seed; generated on L40S, so values differ from the multi-seed tables. $^\ast$Pre-specified primary.}
  \label{tab:tshift}
  \centering\scriptsize
  \setlength{\tabcolsep}{2.5pt}
  \begin{tabular}{@{}lccl@{}}
    \toprule
    Arm (10k) & \Synclane{} & Layout & $D$ \\
    \midrule
    Swap + Min-SNR$^\ast$ & 68.3 $\to$ 26.7 & 87.3 $\to$ 86.0 & $+40.3$ [$+32.0$, $+48.7$] \\
    \quad seed 0 only & 70.7 $\to$ 25.3 & 85.3 $\to$ 85.3 & $+45.3$ [$+34.7$, $+56.0$] \\
    Noisy-input swap & 52.0 $\to$ 16.0 & 79.3 $\to$ 76.7 & $+33.3$ [$+23.3$, $+43.3$] \\
    Min-SNR & 55.3 $\to$ 27.3 & 77.3 $\to$ 76.0 & $+26.7$ [$+17.3$, $+36.0$] \\
    Ordinary, LR$\times$2 & 40.0 $\to$ 24.0 & 73.3 $\to$ 72.0 & $+14.7$ [$+6.0$, $+24.0$] \\
    \bottomrule
  \end{tabular}
\end{table}

The primary is positive (Table~\ref{tab:tshift}), and every arm loses accuracy on \synclane{} only: under the delay, \synclane{} switching is 16.0--27.3\%, about the flat 25\% guessing level (guessing among readable outputs: 11.8--15.8\%), while layout changes by at most 2.6 points. Since departure order and the static cues survive the delay and onset timing improves, the drop cannot come from an order, onset or static reader.

Under the delay, correct switches still outnumber reversed switches about two to one (for swap + Min-SNR, reversals rise from 0.6 to 13.9\% on contradicted pairs), so some signal other than same-moment evidence remains; we do not identify it. A delayed input is also a distribution shift; layout, which the delay leaves solvable, is its control.

\subsection{Correct, reversed and same-actor switches}
\label{app:reversed}
A pair whose two outputs are both readable either switches correctly, switches in reverse (each branch shows the other branch's actor), or shows the same patch in both branches. A policy that ignores Video~1 but follows Video~2 consistently would switch correctly and in reverse equally often. Table~\ref{tab:reversed} splits the 150 ordinary pairs per link. On \synclane{}, the 5,000-update seeds reverse two thirds as often as they switch correctly (53 against 80 over three seeds; seed~0: 25 against 24), while the long-trained seeds almost never reverse (274 vs 12 over three seeds). In complete squares, the commonest failure of the 5,000-update models ignores Video~1: the depicted patch follows which cube acts but not which cube carried it (18 of 56 squares, against 12.5\% by chance).

\begin{table}[!htbp]
  \caption{Outcomes of the 150 held-out ordinary pairs per link: correct switch / reversed switch / same patch in both branches / at least one output unreadable.}
  \label{tab:reversed}
  \centering\footnotesize
  \setlength{\tabcolsep}{3pt}
  \begin{tabular}{@{}lcc@{}}
    \toprule
    Model & Layout & \Synclane{} \\
    \midrule
    Two-video, seed 0 & 92 / 3 / 16 / 39 & 24 / 25 / 38 / 63 \\
    Two-video, seed 1 & 61 / 7 / 37 / 45 & 29 / 14 / 34 / 73 \\
    Two-video, seed 2 & 88 / 0 / 21 / 41 & 27 / 14 / 36 / 73 \\
    +InputHead & 77 / 0 / 19 / 54 & 37 / 8 / 33 / 72 \\
    +CF & 104 / 0 / 17 / 29 & 54 / 11 / 26 / 59 \\
    Two-video-long, seed 0 & 116 / 0 / 7 / 27 & 93 / 2 / 15 / 40 \\
    Two-video-long, seed 1 & 113 / 0 / 15 / 22 & 85 / 6 / 20 / 39 \\
    Two-video-long, seed 2 & 125 / 0 / 9 / 16 & 96 / 4 / 16 / 34 \\
    Modular RGB & 118 / 0 / 12 / 20 & 77 / 10 / 12 / 51 \\
    ReCamMaster, 5k & 22 / 10 / 18 / 100 & 9 / 3 / 13 / 125 \\
    \bottomrule
  \end{tabular}
\end{table}

\subsection{Pre-specified confirmation of noisy-input swap}
\label{app:confirm}
Noisy-input swap was selected on the 100 validation pairs (Appendix~\ref{app:methodstudy}), after the pre-specified method, +CF, had failed its compute-matched criterion (Appendix~\ref{app:cf}). The following protocol for the held-out test (Table~\ref{tab:methodheldout}) was fixed before any held-out result of the 256-pixel runs was seen.

\emph{Frozen recipe.} Noisy-input swap with $p=0.5$, swapping only at $\sigma\ge1$; learning rate $4\times10^{-6}$; 18,000-update schedule; 256 pixels; no other change.
\emph{Arms.} Three seeds (0--2) each of noisy-input swap, ordinary training, and ordinary training at a doubled learning rate ($8\times10^{-6}$). Secondary, one seed: noisy-input swap at $8\times10^{-6}$, evaluated at 17,500 updates only (extended afterwards to three seeds and to 10,000 updates; Appendix~\ref{app:deviations}). The midpoint variant is not part of this test; it was evaluated on the held-out split later, as a control (Appendix~\ref{app:heldoutmore}). The seed-0 noisy-input swap run and the seed-0 and seed-1 ordinary runs are the development runs of Section~\ref{sec:experiments}; seed 0 at the doubled rate is the development control.
\emph{Splits and checkpoints.} The confirmatory final pairs (300 pairs), squares (100), and nuisance pairs (100), the relay selections, and the placement selections, all at 17,500 updates; the confirmatory final pairs also at 10,000 updates.
\emph{Primary analysis.} Pooled switch accuracy on the 300 confirmatory final pairs per arm, averaged over its three seeds, with a paired bootstrap over pairs in which each pair's switch outcome is averaged over seeds. The primary contrasts are noisy-input swap minus ordinary training and noisy-input swap minus ordinary training at the doubled learning rate, each at 10,000 updates (speed) and at 17,500 updates (end point).
\emph{Secondary measures.} Per-link switches, relay, all-four squares, nuisance pairs, event detection, placement, and PSNR. All results are reported whatever the outcome.

\emph{Fresh-split confirmation.} Pre-specified on October 16 and committed, with a timestamp, to the project repository before any output on the fresh split (the log will be released with the code): ordinary training against noisy-input swap, both at $8\times10^{-6}$ (the best learning rate known for either arm; the later sweep at $1.6\times10^{-5}$ changed neither, Appendix~\ref{app:methodstudy}), seeds 10--15, 256 pixels, 18,000-update schedule; primary statistic swap minus ordinary pooled switch accuracy on the fresh split's 300 ordinary pairs at 10,000 updates, seed-level Welch test (one-sided, $\alpha=0.05$) and a bootstrap over pairs and seeds; secondary the same at 17,500 (two-sided), Video-1 sensitivity on layout squares (the fresh \synclane{} squares failed their link gate and were quarantined before any model output), shortcut-incongruent items and the pre-defined classifier-balanced subset; all seeds reported.

Both arms were generated on the same GPU model. Pooled switch accuracy per seed (seeds 10--15): ordinary 61.0, 60.7, 54.7, 55.7, 61.7, 59.7 and noisy-input swap 71.7, 73.0, 60.3, 60.0, 70.7, 77.3 at 10,000 updates ($t=3.17$, $p=0.0083$); ordinary 70.3, 73.0, 75.0, 72.7, 73.0, 70.3 and noisy-input swap 79.3, 77.0, 79.0, 75.0, 77.3, 78.0 at 17,500 ($t=5.37$, $p=0.0003$).

The pre-specified secondaries agree in direction: on the 74-pair classifier-balanced subset, noisy-input swap leads by $+18.9$ [$+12.2$, $+25.7$] at 10,000 and $+7.9$ [$+1.6$, $+14.4$] at 17,500 updates, and Video-1 sensitivity on layout squares is near ceiling for both arms (80.2\% each at 10,000; 91.0 against 87.8\% at 17,500).

The earlier split shows the same early advantage: LR$\times$2 needs about 15k updates to reach noisy-input swap's 10k score. Its later contrast is unresolved, and seed-level tests do not survive multiple-comparison correction. These limitations motivated the fresh confirmation (Appendices~\ref{app:confirm} and~\ref{app:seedlevel}).

\emph{First-split outcome.} Table~\ref{tab:methodheldoutfull} reports all primary contrasts, with seed-resampled intervals; Figure~\ref{fig:forest} plots those against LR$\times$2. The protocol fixed no decision threshold; reading a contrast as positive when its 95\% interval excludes zero, three of the four are positive and noisy-input swap minus the doubled learning rate at 17,500 updates is not. Section~\ref{sec:experiments} also gives intervals that resample seeds; they exclude zero for the same three contrasts. After the primary results, the secondary arm was also evaluated at 10,000 updates and extended to three seeds, and a low-noise variant chosen on validation received its own held-out check (Table~\ref{tab:deviations}).

\begin{table*}[!tp]
  \caption{Full held-out comparison at 576 pixels (main-text Table~\ref{tab:core} shows Two-video and Two-video-long) (\%; frozen scorers). Switch: right in both branches (150 ordinary pairs per link); all-four: 50 squares per link; nuisance: 50 same-actor pairs per link; placement: 150 pairs per link on an attribute never seen in training; event: target event detected in the 600 pair outputs. Two-video and Two-video-long: mean over three seeds [range]; $^\dagger$one run. $\Delta$: pooled switch difference from Two-video, 95\% hierarchical bootstrap over scenes and seeds. Guessing among each model's readable outputs gives 13--19\% \synclane{} switches, depending on its coverage. First block: 5,000 denoising updates (+InputHead and +CF add 5,000 auxiliary ones); Two-video-long: 17,000 unless marked 10k; ---: not evaluated or not compared. $^\S$Noisy-input swap at 576 pixels after 10,000 updates of the 18,000-update schedule, two seeds [range]; pre-specified comparison with Two-video-long at 10,000 updates: $+22.9$ [$+13.9$, $+32.6$], resampling scenes and seeds (Section~\ref{sec:speed}). Modular RGB is a perceive--decide--re-render reference, not a generator.}
  \label{tab:corefull}
  \centering\small
  \setlength{\tabcolsep}{2.6pt}
  \resizebox{\textwidth}{!}{\begin{tabular}{@{}l ccc c cc cc cc c@{}}
    \toprule
    & \multicolumn{3}{c}{Switch} & & \multicolumn{2}{c}{All-four} & \multicolumn{2}{c}{Nuisance} & \multicolumn{2}{c}{Placement} & \\
    \cmidrule(lr){2-4}\cmidrule(lr){6-7}\cmidrule(lr){8-9}\cmidrule(lr){10-11}
    Model & Layout & \synchead{} & Pooled & $\Delta$ pooled [95\% CI] & Layout & \synchead{} & Layout & \synchead{} & Layout & \synchead{} & Event \\
    \midrule
    Two-video & 53.6 {\scriptsize[41--61]} & 17.8 {\scriptsize[16--19]} & 35.7 {\scriptsize[30--39]} & --- & 40.7 & 1.3 & 65.3 & 27.3 & 26.7 & 7.6 & 82.0 \\
    +InputHead$^\dagger$ & 51.3 & 24.7 & 38.0 & $+2.3$ [$-4.8$, $+10.0$] & 50 & 16 & 62 & 30 & 41.3 & 7.3 & 73.5 \\
    +CF$^\dagger$ & 69.3 & 36.0 & 52.7 & $+17.0$ [$+10.1$, $+25.0$] & 64 & 20 & 76 & 56 & 42.0 & 9.3 & 87.0 \\
    \midrule
    Two-video-long, 10k & 61.3 & 40.9 & 51.1 & --- & --- & --- & --- & --- & --- & --- & --- \\
    Two-video-long & 78.7 {\scriptsize[75--83]} & 60.9 {\scriptsize[57--64]} & 69.8 {\scriptsize[66--74]} & $+34.1$ [$+26.8$, $+41.9$] & 70.7 & 48.0 & 82.0 & 71.3 & 39.3 & 10.2 & 91.2 \\
    Noisy-input swap, 10k$^\S$ & 83.7 {\scriptsize[83--84]} & 64.3 {\scriptsize[60--69]} & 74.0 {\scriptsize[72--76]} & --- & 74.0 & 62.0 & 78.0 & 71.0 & & & \\
    \midrule
    Modular RGB$^\dagger$ & 78.7 & 51.3 & 65.0 & $+29.3$ [$+21.8$, $+37.4$] & 70 & 52 & 86 & 72 & 60.0 & 44.7 & 83.8 \\
    ReCamMaster, 5k$^\dagger$ & 14.7 & 6.0 & 10.3 & $-25.3$ [$-32.2$, $-17.8$] & 0 & 0 & 14 & 14 & 4.7 & 2.7 & 53.3 \\
    ReCamMaster, 17k$^\dagger$ & 48.7 & 17.3 & 33.0 & $-2.7$ [$-9.1$, $+4.4$] & 42 & 4 & 64 & 30 & 8.0 & 9.3 & 73.5 \\
    \bottomrule
  \end{tabular}}
\end{table*}

\emph{576 pixels.} The gain also appears at AnyView's native resolution. At 10k updates, noisy-input swap improves pooled accuracy by 22.9 points ([$+13.9$, $+32.6$]; two seeds against three). Its mean already exceeds ordinary training's at 17k, with similar per-update cost. This baseline was not learning-rate tuned (Appendix Table~\ref{tab:corefull}).

\subsection{Further held-out contrasts}
\label{app:heldoutmore}
\begin{table*}[!tp]
  \caption{Noisy-input swap on the held-out test, all arms and columns (main-text Table~\ref{tab:methodheldout} repeats the fresh-split rows; 256 pixels; \%; columns as in Table~\ref{tab:corefull}). Mean over three seeds. $\Delta$~ord.: pooled switch difference from ordinary training at the same updates, with 95\% intervals from the pre-specified paired bootstrap over pairs (Appendix~\ref{app:confirm}) and, second, from one that also resamples seeds. $\Delta$~LR: difference from LR$\times$2 at the same updates, with the pairs interval. The last block is the pre-specified confirmation on scenes never used before. Rows below the thin rule in each earlier block were added or selected after the primary results: $^\dagger$secondary arm, pre-specified for one seed at 17.5k and extended to three seeds and to 10k afterwards; $^\ddagger$variants chosen on validation (low-$\sigma$: swap only at $1\le\sigma<5$; sibling-free: color-permuted own target). Seed~0 of ordinary training and of noisy-input swap each resumed once from a checkpoint (Appendix~\ref{app:models}).}
  \label{tab:methodheldoutfull}
  \centering\small
  \setlength{\tabcolsep}{2.5pt}
  \resizebox{\textwidth}{!}{\begin{tabular}{@{}l ccc cc cc cc c@{}}
    \toprule
    & \multicolumn{3}{c}{Switch} & & & \multicolumn{2}{c}{All-four} & \multicolumn{2}{c}{Placement} & \\
    \cmidrule(lr){2-4}\cmidrule(lr){7-8}\cmidrule(lr){9-10}
    Training & Layout & \synchead{} & Pooled & $\Delta$~ord.\ [pairs] [+seeds] & $\Delta$~LR [pairs] & Layout & \synchead{} & Layout & \synchead{} & Event \\
    \midrule
    \multicolumn{11}{@{}l}{\emph{10,000 updates}} \\
    Ordinary & 70.0 & 30.9 & 50.4 & --- & --- & 62.7 & 22.0 &  &  & 87.8 \\
    Ordinary, LR$\times$2 & 74.2 & 38.0 & 56.1 & $+5.7$ [$+2.2$, $+9.0$] [$-0.9$, $+12.3$] & --- & 66.0 & 26.7 &  &  & 89.3 \\
    Noisy-input swap & 78.9 & 52.7 & 65.8 & $+15.3$ [$+11.6$, $+19.0$] [$+8.3$, $+22.8$] & $+9.7$ [$+5.9$, $+13.4$] & 72.7 & 48.0 &  &  & 89.2 \\
    \cmidrule(lr){1-11}
    Noisy-input swap, LR$\times$2$^\dagger$ & 80.9 & 59.1 & 70.0 & $+19.6$ [$+15.9$, $+23.1$] [$+11.7$, $+27.7$] & $+13.9$ [$+10.3$, $+17.4$] & 76.0 & 50.7 &  &  & 89.2 \\
    Low-$\sigma$ swap$^\ddagger$ & 81.1 & 56.9 & 69.0 & $+18.6$ [$+14.8$, $+22.2$] [$+11.2$, $+26.1$] & $+12.9$ [$+9.0$, $+16.8$] &  &  &  &  &  \\
    Sibling-free$^\ddagger$ & 82.0 & 58.7 & 70.3 & $+19.9$ [$+16.2$, $+23.6$] [$+13.2$, $+27.1$] & $+14.2$ [$+10.7$, $+18.0$] &  &  &  &  &  \\
    \midrule
    \multicolumn{11}{@{}l}{\emph{12,500 updates} (added after the primary results; ordinary training not run)} \\
    Ordinary, LR$\times$2 & 70.9 & 50.0 & 60.4 & --- & --- &  &  &  &  &  \\
    Noisy-input swap & 79.3 & 60.2 & 69.8 & --- & $+9.3$ [$+5.6$, $+13.1$] &  &  &  &  &  \\
    \midrule
    \multicolumn{11}{@{}l}{\emph{17,500 updates}} \\
    Ordinary & 80.0 & 52.7 & 66.3 & --- & --- & 74.0 & 36.7 & 21.3 & 12.2 & 92.7 \\
    Ordinary, LR$\times$2 & 84.2 & 60.2 & 72.2 & $+5.9$ [$+3.0$, $+8.8$] [$+0.3$, $+11.3$] & --- & 78.7 & 55.3 & 14.9 & 13.8 & 93.1 \\
    Noisy-input swap & 85.8 & 61.6 & 73.7 & $+7.3$ [$+3.9$, $+10.9$] [$+1.4$, $+13.2$] & $+1.4$ [$-2.0$, $+4.8$] & 84.7 & 52.7 & 39.1 & 14.0 & 91.4 \\
    \cmidrule(lr){1-11}
    Noisy-input swap, LR$\times$2$^\dagger$ & 89.1 & 68.7 & 78.9 & $+12.6$ [$+9.0$, $+16.2$] [$+5.9$, $+19.2$] & $+6.7$ [$+3.3$, $+10.0$] & 82.7 & 59.3 & 27.8 & 15.1 & 93.0 \\
    Low-$\sigma$ swap$^\ddagger$ & 86.7 & 64.9 & 75.8 & $+9.4$ [$+5.8$, $+13.2$] [$+3.6$, $+15.4$] & $+3.6$ [$-0.0$, $+7.2$] &  &  &  &  &  \\
    Sibling-free$^\ddagger$ & 87.6 & 63.1 & 75.3 & $+9.0$ [$+5.7$, $+12.4$] [$+3.7$, $+14.7$] & $+3.1$ [$-0.1$, $+6.4$] &  &  &  &  &  \\
    \midrule
    \multicolumn{11}{@{}l}{\emph{Fresh held-out split, pre-specified confirmation: six new seeds per arm, both at LR$\times$2; $\Delta$ from ordinary training at LR$\times$2}} \\
    Ordinary, LR$\times$2, 10k & 71.8 & 46.0 & 58.9 & --- & --- &  &  &  &  &  \\
    Noisy-input swap, LR$\times$2, 10k & 74.2 & 63.4 & 68.8 & $+9.9$ [$+7.1$, $+12.9$] [$+2.9$, $+16.7$] &  &  &  &  &  &  \\
    Ordinary, LR$\times$2, 17.5k & 81.8 & 63.0 & 72.4 & --- & --- &  &  &  &  &  \\
    Noisy-input swap, LR$\times$2, 17.5k & 84.9 & 70.3 & 77.6 & $+5.2$ [$+2.5$, $+7.9$] [$+1.4$, $+9.1$] &  &  &  &  &  &  \\
    \bottomrule
  \end{tabular}}
\end{table*}

These complement the held-out comparison of Section~\ref{sec:experiments} (Tables~\ref{tab:methodheldout} and~\ref{tab:methodheldoutfull}). Table~\ref{tab:contrasts} collects the further pooled contrasts. In the pre-specified band test, patch weighting restricted to $1\le\sigma<5$ is similar to weighting everywhere, and restricted to $\sigma\ge20$ does not help; these bands cover different shares of draws (52\% and 8\%), so the comparison does not match total added gradient weight or establish copying as the sole mechanism.

Per-seed pooled switches after 10,000 updates (generated from the same pool of GPUs as the reference arms): patch weighting ($1{+}4\times$mask) 66.0, 72.0 and 59.3; Min-SNR-5 65.3 and 64.7; noisy-input swap with Min-SNR 80.3 and 78.0; noisy-input swap with patch weighting 69.3 and 70.7; the midpoint 65.7, 62.3 and 64.7 (layout 79.6, \synclane{} 48.9); swapping only at $\sigma\ge20$ 46.3 and 51.0; MLLM-box weighting 64.0 and 65.0; MLLM-box weighting with Min-SNR 73.0 and 68.7 (pre-specified October 22; better single arm Min-SNR).

Patch weighting restricted to one noise band (pre-specified October 22, two seeds each): only at $1\le\sigma<5$ 67.3 and 68.0, only at $\sigma\ge5$ 54.7 and 62.0, only at $\sigma\ge20$ 42.0 and 54.7.

On a 40,000-update schedule (pre-specified October 21; LR$\times$2, three seeds each), ordinary training reaches 70.7, 70.0 and 73.3 at 20k and 79.0, 82.3 and 79.7 at 40k, noisy-input swap 77.3, 79.3 and 74.3 at 20k and 82.7, 79.7 and 80.0 at 40k; under the pre-specified rule the 40k end point is inconclusive (interval not within $\pm3$) and noisy-input swap at 20k does not reach the ordinary 40k end point ($-3.3$ [$-6.7$, $-0.1$]). Under the pre-specified rule (combination minus the better single arm, pairs interval above zero), all three combinations improve on their better individual component.

\emph{Fresh-split combination test.} Pre-specified on October 22 and committed to the project repository before any output of these runs: noisy-input swap with Min-SNR and Min-SNR alone, with the recipe of the first split (base learning rate $4\times10^{-6}$, Min-SNR-5, swap probability 0.5 at $\sigma\ge1$) and six new seeds each (10--15), against the fresh split's ordinary and noisy-input swap arms at LR$\times$2 (Appendix~\ref{app:confirm}); all four arms generated on the same GPU model. Primary statistic: the combination minus the better single arm by seed mean, pooled switch accuracy at 10,000 updates; the combination improves on the better single arm if the one-sided seed-level Welch test rejects the null ($\alpha=0.05$) and the paired-bootstrap interval over pairs has a lower bound above zero.

Per-seed pooled switches: Min-SNR 67.0, 67.3, 64.3, 65.7, 75.0, 68.0; noisy-input swap with Min-SNR 80.3, 81.7, 78.3, 81.0, 81.7, 79.0. The better single arm is noisy-input swap (68.8 against 67.9\%); the combination exceeds it by $+11.5$ [$+8.7$, $+14.4$] ($t=3.90$, $p=0.0049$), meeting the criterion on the fresh split (Table~\ref{tab:contrasts}).

On \synclane{} it leads noisy-input swap by $+13.0$ [$+8.7$, $+17.2$] and ordinary training by $+30.4$ [$+25.1$, $+35.8$]; on the 74-pair classifier-balanced subset, by $+13.7$ [$+8.3$, $+19.1$] and $+32.7$ [$+25.5$, $+39.9$].

\emph{Past 10,000 updates.} Pre-specified on October 27 and committed before any output of these continuations: the six combination runs continue to 17,500 updates on the same schedule, against noisy-input swap at LR$\times$2. Per-seed pooled switches are 81.3, 83.0, 83.0, 84.0, 81.3 and 83.0, a mean of 82.6\% (layout 89.8, \synclane{} 75.4). The primary, the combination minus noisy-input swap, is $+5.0$ [$+2.4$, $+7.5$] (seeds resampled: [$+1.6$, $+8.4$]), and the difference from LR$\times$2 is $+10.2$ [$+7.2$, $+13.4$] (secondary). The combination's own change from 10,000 to 17,500 updates is $+2.3$ [$+0.2$, $+4.4$] (seeds resampled: [$-0.8$, $+5.2$]), with layout rising from 84.2 to 89.8 and \synclane{} flat (76.4 to 75.4). Min-SNR alone reaches 75.9\% (layout 84.0, \synclane{} 67.9), up $+8.1$ [$+5.3$, $+10.8$] (seeds resampled: [$+3.4$, $+12.4$]) from 10,000 updates; the combination leads it by $+6.7$ [$+3.9$, $+9.4$] ([$+3.3$, $+10.2$]), against $+12.4$ at 10,000. All arms were generated on the same GPU model.

\emph{Shared-rate combination test.} Pre-specified on October 26 and committed before any output: Min-SNR and noisy-input swap with Min-SNR retrained at LR$\times$2 ($8\times10^{-6}$) with seeds 10--12, against the fresh split's ordinary and noisy-input swap arms at LR$\times$2 with the same seeds, all four generated on the same GPU model; primary statistic swap+Min-SNR minus noisy-input swap, positive if the lower bound of the paired bootstrap over pairs exceeds zero. Pooled switches (layout / \synclane{}): ordinary 58.8\% (73.3 / 44.2; seeds 61.0, 60.7, 54.7), noisy-input swap 68.3\% (75.8 / 60.9; 71.7, 73.0, 60.3), Min-SNR 71.8\% (79.3 / 64.2; 68.0, 71.0, 76.3), noisy-input swap with Min-SNR 78.8\% (84.9 / 72.7; 80.3, 78.7, 77.3).

Primary: $+10.4$ [$+6.7$, $+14.3$], positive (seeds resampled [$+2.9$, $+19.0$]; the seed-level Welch test, not part of this rule, gives one-sided $p=0.058$).

Secondary contrasts are in Table~\ref{tab:contrasts}; on \synclane{}, Min-SNR leads ordinary training by $+20.0$ [$+14.0$, $+25.8$] and the combination leads swap by $+11.8$ [$+6.2$, $+17.3$].

Not pre-specified: the combination leads Min-SNR alone by $+7.0$ [$+3.7$, $+10.3$] (seeds resampled [$+0.9$, $+13.0$]; one-sided Welch $p=0.045$), on \synclane{} by $+8.4$ [$+3.8$, $+13.1$] ([$-1.1$, $+18.2$]) and on layout by $+5.6$ [$+1.1$, $+10.4$] ([$-2.0$, $+12.7$]).

With all six seeds of the ordinary and noisy-input swap arms, the combination leads swap by $+9.9$ [$+6.5$, $+13.4$] and Min-SNR leads ordinary training by $+12.9$ [$+9.5$, $+16.3$]. Against the base-rate arms of the combination test, doubling the rate changes the combination by $-1.6$ [$-4.1$, $+0.9$] and Min-SNR by $+3.9$ [$+1.1$, $+6.7$] ([$-2.3$, $+10.1$]).

\emph{Midpoint + Min-SNR and recolor + Min-SNR at the shared rate.} Each comparison was pre-specified before its outputs were generated (seeds 10--12, the same recipe, two-sided reading against swap + Min-SNR). Pooled switches (layout / \synclane{}): midpoint + Min-SNR 79.2\% (83.3 / 75.1; seeds 79.7, 83.0, 75.0) and recolor + Min-SNR 79.4\% (82.7 / 76.2; 77.0, 79.7, 81.7), against 78.8\% for swap + Min-SNR (84.9 / 72.7; 80.3, 78.7, 77.3). Swap + Min-SNR minus midpoint + Min-SNR is $-0.4$ [$-3.3$, $+2.6$] (seeds resampled [$-6.0$, $+5.2$]; \synclane{} $-2.4$, layout $+1.6$) and minus recolor + Min-SNR $-0.7$ [$-3.7$, $+2.3$] ([$-5.2$, $+4.1$]; \synclane{} $-3.6$, layout $+2.2$); recolor minus midpoint is $+0.2$ [$-3.1$, $+3.4$]. None is distinguished: the pair intervals lie inside the pre-specified $\pm5$ range for practical equivalence, the seed-resampled intervals do not. The midpoint uses the same rendered siblings as noisy-input swap; the recolor needs none but relies on this palette's symmetry.

Complete squares provide a stricter, less conclusive check. The combination has the highest \synclane{} all-four score, but its margin over noisy-input swap includes zero ($+14.0$ [$-4.0$, $+32.0$]). Nuisance-pair accuracy also improves (first split, two seeds per arm; comparisons specified before scoring existing outputs; Appendix~\ref{app:heldoutmore}).

\emph{Swap + Min-SNR on complete squares.} Pre-specified before these scores were computed, on outputs generated earlier, and reported whatever they show: all-four squares, nuisance pairs and Video-1 sensitivity on the first split after 10,000 updates, seeds 0 and 1 of every arm (Table~\ref{tab:headlinearm}). The fresh \synclane{} squares failed their gate and are not used. These are matched two-seed means, so they differ from the three-seed rows of Table~\ref{tab:methodheldoutfull}; ordinary training's \synclane{} all-four is 13.0\% here (seeds 6.0 and 20.0) and 22.0\% there.

The primary, swap + Min-SNR minus noisy-input swap on \synclane{} all-four, is $+14.0$ [$+0.0$, $+27.0$] (seeds resampled [$-4.0$, $+32.0$]), not positive by the rule that the lower bound exceed zero; with 50 squares and two seeds it is not resolved, which is not equivalence.

The combination scores highest on every measure of Table~\ref{tab:headlinearm}; its secondary contrasts on these measures will be released with the analysis log.

\emph{Transfer to an untrained attribute.} Noisy-input swap transfers to an untrained attribute on layout only. After color training, its layout score on patch placement exceeds ordinary training by 25.0 points at 10k ([$+16.3$, $+34.0$]; first split, two seeds, base rate; secondary), whereas Min-SNR does not help. No arm transfers on \synclane{}, and adding Min-SNR reduces noisy-input swap's layout transfer (Section~\ref{sec:discussion}).

\emph{Placement} (pre-specified secondary; scorer v2, 150 pairs and 50 squares per link; models trained on color, tested on placement; at 10,000 updates, so not comparable with the 17,500-update placement rows of Table~\ref{tab:methodheldoutfull}). Layout switch is 17.7\% for ordinary training at the same rate, 42.7\% for noisy-input swap, 12.7\% for Min-SNR and 24.7\% for swap + Min-SNR; \synclane{} switch is 11.0, 14.0, 7.7 and 13.7\%, near the floor for every arm. All-four is 3.0, 24.0, 0.0 and 7.0\% on layout and 0.0, 5.0, 1.0 and 1.0\% on \synclane{}.

On layout switch, swap + Min-SNR minus noisy-input swap is $-18.0$ [$-24.7$, $-11.3$] [$-27.3$, $-9.3$] and noisy-input swap minus ordinary training $+25.0$ [$+18.7$, $+31.3$] [$+16.3$, $+34.0$]; on \synclane{} switch the former is $-0.3$ [$-5.3$, $+5.0$]. Adding Min-SNR thus lowers noisy-input swap's layout gain on this untrained attribute, the one measure so far where noisy-input swap alone exceeds the combination.

\begin{table}[!htbp]
  \caption{Swap + Min-SNR on complete squares and nuisance pairs (first split, 10,000 updates, seeds 0 and 1 of every arm; \%; 50 squares and 50 pairs per link). Matched two-seed means; Table~\ref{tab:methodheldoutfull} gives three seeds for the first three arms.}
  \label{tab:headlinearm}
  \centering\footnotesize
  \setlength{\tabcolsep}{2.5pt}
  \begin{tabular}{@{}lccccc@{}}
    \toprule
    Measure & Ord. & LR$\times$2 & Swap & Min-SNR & Swap + Min-SNR \\
    \midrule
    \synchead{} all-four & 13.0 & 29.0 & 45.0 & 47.0 & \textbf{59.0} \\
    Layout all-four & 64.0 & 62.0 & 71.0 & 71.0 & \textbf{77.0} \\
    \synchead{} nuisance & 43.0 & 51.0 & 64.0 & 70.0 & \textbf{84.0} \\
    Layout nuisance & 70.0 & 76.0 & 80.0 & 84.0 & \textbf{86.0} \\
    \synchead{} Video-1 sens. & 31.0 & 53.5 & 68.5 & 62.5 & \textbf{72.5} \\
    Layout Video-1 sens. & 77.0 & 76.5 & 83.0 & 83.5 & \textbf{89.5} \\
    \bottomrule
  \end{tabular}
\end{table}

\begin{table}[!htbp]
  \caption{Further held-out contrasts (swap: noisy-input swap; ord.: ordinary training): pooled switch differences (points) with 95\% paired bootstrap intervals over pairs and, where given, over pairs and seeds; MS: Min-SNR. Patch weighting, the midpoint and the two variants chosen on validation (low-$\sigma$ swap, sibling-free recolor): three seeds; Min-SNR, the combinations, swapping only at $\sigma\ge20$ and MLLM boxes: two; fresh split: six per arm.}
  \label{tab:contrasts}
  \centering\footnotesize
  \setlength{\tabcolsep}{2pt}\renewcommand{\arraystretch}{0.90}
  \begin{tabular}{@{}lrcc@{}}
    \toprule
    Contrast & $\Delta$ & Pairs & +Seeds \\
    \midrule
    \multicolumn{4}{@{}l}{\emph{10,000 updates}} \\
    Min-SNR $-$ ord. & $+14.6$ & {\scriptsize[$+10.5$, $+18.6$]} & \\
    Patch weight $-$ ord. & $+15.3$ & {\scriptsize[$+11.8$, $+18.7$]} & \\
    Swap + Min-SNR $-$ ord. & $+28.7$ & {\scriptsize[$+24.3$, $+33.2$]} & \\
    Swap $-$ ord., 2nd sampling seed & $+12.3$ & {\scriptsize[$+8.8$, $+15.9$]} & \\
    Low-$\sigma$ swap $-$ swap & $+3.2$ & {\scriptsize[$-0.1$, $+6.4$]} & \\
    Sibling-free $-$ swap & $+4.6$ & {\scriptsize[$+1.2$, $+8.0$]} & \\
    Patch weight $-$ swap & $0.0$ & {\scriptsize[$-3.6$, $+3.6$]} & {\scriptsize[$-8.2$, $+8.2$]} \\
    Patch weight $-$ swap at LR$\times$2 & $-4.2$ & {\scriptsize[$-7.9$, $-0.7$]} & \\
    Min-SNR $-$ swap & $-0.8$ & {\scriptsize[$-5.1$, $+3.6$]} & {\scriptsize[$-7.2$, $+5.4$]} \\
    Swap + Min-SNR $-$ swap & $+13.4$ & {\scriptsize[$+9.4$, $+17.3$]} & {\scriptsize[$+7.2$, $+19.3$]} \\
    Swap + Min-SNR $-$ Min-SNR & $+14.2$ & {\scriptsize[$+10.0$, $+18.5$]} & {\scriptsize[$+9.0$, $+19.3$]} \\
    Midpoint $-$ ord. & $+13.8$ & {\scriptsize[$+10.2$, $+17.3$]} & \\
    Midpoint $-$ LR$\times$2 & $+8.1$ & {\scriptsize[$+4.7$, $+11.6$]} & \\
    Midpoint $-$ swap & $-1.6$ & {\scriptsize[$-4.9$, $+1.8$]} & \\
    Swap at $\sigma\ge20$ only $-$ ord. & $-1.8$ & {\scriptsize[$-5.6$, $+2.2$]} & \\
    Swap at $\sigma\ge20$ only $-$ swap & $-17.1$ & {\scriptsize[$-21.7$, $-12.6$]} & \\
    Patch w.\ at $1\le\sigma<5$ $-$ ord. & $+17.2$ & {\scriptsize[$+13.3$, $+21.2$]} & {\scriptsize[$+10.6$, $+24.0$]} \\
    Patch w.\ at $\sigma\ge5$ $-$ ord. & $+7.9$ & {\scriptsize[$+4.1$, $+11.7$]} & {\scriptsize[$-0.4$, $+16.3$]} \\
    Patch w.\ at $\sigma\ge20$ $-$ ord. & $-2.1$ & {\scriptsize[$-5.7$, $+1.5$]} & {\scriptsize[$-12.8$, $+8.7$]} \\
    Patch w.: $1\le\sigma<5$ $-$ $\sigma\ge20$ & $+19.3$ & {\scriptsize[$+15.2$, $+23.5$]} & {\scriptsize[$+9.2$, $+29.3$]} \\
    Patch w.\ at $1\le\sigma<5$ $-$ all $\sigma$ & $+1.9$ & {\scriptsize[$-1.8$, $+5.5$]} & {\scriptsize[$-5.6$, $+9.7$]} \\
    \midrule
    \multicolumn{4}{@{}l}{\emph{Fresh split, 10,000 updates (six seeds; swap at LR$\times$2)}} \\
    Swap + Min-SNR $-$ swap & $+11.5$ & {\scriptsize[$+8.7$, $+14.4$]} & {\scriptsize[$+5.3$, $+18.1$]} \\
    Swap + Min-SNR $-$ Min-SNR & $+12.4$ & {\scriptsize[$+9.6$, $+15.5$]} & {\scriptsize[$+7.7$, $+17.1$]} \\
    Swap + Min-SNR $-$ LR$\times$2 & $+21.4$ & {\scriptsize[$+18.3$, $+24.8$]} & {\scriptsize[$+16.8$, $+26.0$]} \\
    Min-SNR $-$ LR$\times$2 & $+9.0$ & {\scriptsize[$+6.2$, $+11.9$]} & {\scriptsize[$+3.9$, $+14.3$]} \\
    \midrule
    \multicolumn{4}{@{}l}{\emph{Fresh split, 17,500 updates (six seeds; swap at LR$\times$2)}} \\
    Swap + Min-SNR $-$ swap & $+5.0$ & {\scriptsize[$+2.4$, $+7.5$]} & {\scriptsize[$+1.6$, $+8.4$]} \\
    Swap + Min-SNR $-$ Min-SNR & $+6.7$ & {\scriptsize[$+3.9$, $+9.4$]} & {\scriptsize[$+3.3$, $+10.2$]} \\
    Swap + Min-SNR $-$ LR$\times$2 & $+10.2$ & {\scriptsize[$+7.2$, $+13.4$]} & {\scriptsize[$+6.3$, $+14.2$]} \\
    \midrule
    \multicolumn{4}{@{}l}{\emph{Fresh split, shared rate LR$\times$2 (three seeds, 10--12)}} \\
    Swap + Min-SNR $-$ swap & $+10.4$ & {\scriptsize[$+6.7$, $+14.3$]} & {\scriptsize[$+2.9$, $+19.0$]} \\
    Min-SNR $-$ ord. & $+13.0$ & {\scriptsize[$+9.2$, $+16.9$]} & {\scriptsize[$+5.9$, $+20.1$]} \\
    Interaction & $-2.6$ & --- & {\scriptsize[$-12.9$, $+8.1$]} \\
    Swap + MS $-$ midpoint + MS & $-0.4$ & {\scriptsize[$-3.3$, $+2.6$]} & {\scriptsize[$-6.0$, $+5.2$]} \\
    Midpoint + MS $-$ Min-SNR & $+7.4$ & {\scriptsize[$+4.6$, $+10.4$]} & {\scriptsize[$+0.7$, $+13.9$]} \\
    Midpoint + MS $-$ LR$\times$2 & $+20.4$ & {\scriptsize[$+16.4$, $+24.4$]} & {\scriptsize[$+13.6$, $+27.3$]} \\
    Swap + MS $-$ recolor + MS & $-0.7$ & {\scriptsize[$-3.7$, $+2.3$]} & {\scriptsize[$-5.2$, $+4.1$]} \\
    Recolor + MS $-$ Min-SNR & $+7.7$ & {\scriptsize[$+4.3$, $+11.1$]} & {\scriptsize[$+1.6$, $+13.9$]} \\
    Recolor + MS $-$ LR$\times$2 & $+20.7$ & {\scriptsize[$+16.8$, $+24.6$]} & {\scriptsize[$+14.6$, $+26.9$]} \\
    \midrule
    \multicolumn{4}{@{}l}{\emph{40,000-update schedule (three seeds per arm, both at LR$\times$2)}} \\
    Swap $-$ LR$\times$2, 20k & $+5.7$ & {\scriptsize[$+2.1$, $+9.2$]} & {\scriptsize[$+0.3$, $+10.9$]} \\
    Swap $-$ LR$\times$2, 40k & $+0.4$ & {\scriptsize[$-2.6$, $+3.4$]} & {\scriptsize[$-4.0$, $+4.8$]} \\
    Swap at 20k $-$ LR$\times$2 at 40k & $-3.3$ & {\scriptsize[$-6.7$, $-0.1$]} & {\scriptsize[$-8.4$, $+1.6$]} \\
    \midrule
    \multicolumn{4}{@{}l}{\emph{12,500 and 17,500 updates}} \\
    Swap $-$ LR$\times$2, 12.5k & $+9.3$ & {\scriptsize[$+5.6$, $+13.1$]} & {\scriptsize[$+3.8$, $+15.0$]} \\
    Swap, 12.5k $-$ LR$\times$2, 17.5k & $-2.4$ & {\scriptsize[$-5.8$, $+0.9$]} & \\
    Swap at LR$\times$2 $-$ LR$\times$2, 17.5k & $+6.7$ & {\scriptsize[$+3.3$, $+10.0$]} & {\scriptsize[$+1.1$, $+12.8$]} \\
    \bottomrule
  \end{tabular}
\end{table}

\paragraph{Other measures.} At 10,000 updates noisy-input swap invents no more patches than ordinary training (3.8\% of outputs against 4.1\%), keeps the actor's identity on more \synclane{} nuisance pairs (64.7\% against 46.7\%), and leaves event detection and single-video fidelity unchanged. After 10,000 updates on \synclane{}, the doubled learning rate answers 81.5\% of outputs and is right on 74.7\% of those (Table~\ref{tab:coverage}); invented patches occur in 4.7\% of its outputs and 2.5\% of the sibling-free recolor's. Extra noise on both cubes' patches collapses \synclane{} switches to 5 and 3 of 50 (seeds 0 and 1). On relay (secondary), switches after 17,500 updates are 49.8, 60.7, 67.6 and 74.9\% for ordinary training, LR$\times$2, noisy-input swap and noisy-input swap at LR$\times$2.

\paragraph{Placement.} After 17,500 updates, noisy-input swap also improves layout placement over ordinary training (39\% against 21\%); the recolor reaches 29.6\% (three seeds, 22.7--40.7\%), noisy-input swap at the doubled learning rate 27.8\% and the doubled learning rate alone 14.9\%, and \synclane{} placement stays at guessing level for every arm. Permuting colors corrupts only a color answer, so whether the rendered sibling teaches a non-color cue is unclear.

\paragraph{Training directly on placement.} The benefit extends beyond color. Trained directly on vertical patch placement, noisy-input swap improves \synclane{} by 19.3 points over LR$\times$2 at 10k updates ([$+6.0$, $+32.7$]; same rate, two seeds). Min-SNR alone gives a similar gain, while the layout contrast is unresolved (secondary; Appendix~\ref{app:heldoutmore}).

 Pre-specified on October 26 and reported whatever it showed: ordinary training, Min-SNR, noisy-input swap and swap + Min-SNR trained on the placement attribute from the released AnyView, all at LR$\times$2 ($8\times10^{-6}$), 256 pixels, two seeds per arm, and scored after 10,000 updates on the final placement selections with the frozen placement scorer (Table~\ref{tab:placementtrained}). Permuting colors cannot corrupt this attribute, so the recolor does not apply; the midpoint control was not run here.

The primary, noisy-input swap minus ordinary on \synclane{} switch accuracy, is $+19.3$ [$+6.0$, $+32.7$] (scenes only: [$+12.7$, $+25.7$]).

Secondary: Min-SNR minus ordinary is $+23.3$ [$+8.7$, $+38.0$] on \synclane{} switches; the layout contrasts are not resolved.

\begin{table}[!htbp]
  \caption{Training directly on placement (\%; 10,000 updates, all at LR$\times$2, two seeds per arm; 150 pairs and 50 squares per link).}
  \label{tab:placementtrained}
  \centering\scriptsize
  \setlength{\tabcolsep}{3pt}
  \begin{tabular}{@{}lcccc@{}}
    \toprule
    & Ordinary & Min-SNR & Noisy-input swap & Swap + Min-SNR \\
    \midrule
    \Synclane{} switch & 37.0 & 60.3 & 56.3 & 67.7 \\
    Layout switch & 70.0 & 75.0 & 75.7 & 78.7 \\
    \Synclane{} all-four & 19.0 & 45.0 & 41.0 & 54.0 \\
    Layout all-four & 59.0 & 61.0 & 67.0 & 77.0 \\
    \bottomrule
  \end{tabular}
\end{table}

\paragraph{Patch weighting.} Patch weighting, which increases the loss weight inside a masked answer region, gives a similar gain, and boxes from a frozen vision-language model can replace simulator masks (first split, two seeds; Appendix~\ref{app:heldoutmore}).

\paragraph{Where the loss weight goes.} Figure~\ref{fig:effweight} gives, for each training arm, the loss weight times the training density of $\sigma$, computed from the arm's definition (descriptive; no training). Min-SNR keeps 26\% of EDM's total weight and concentrates it at low noise: 95\% of it lies below the ordinary crossover $\sigma\approx2.7$ (ordinary training: 62\%), so it does not move loss weight away from where models copy. Patch weighting multiplies the weight inside the answer region by 4.1 and outside by 0.81 without changing its distribution over $\sigma$.

\input{figures/effweight}

\subsection{Seed-level inference and chance}
\paragraph{Seed-level inference for the held-out contrasts.}\label{app:seedlevel}
Treating each training seed as one observation (three per arm; Welch tests on seed means, Holm correction over the four pre-specified contrasts), noisy-input swap minus ordinary training is $+15.3$ points after 10,000 updates ($p=0.013$; Holm 0.053) and $+7.3$ after 17,500 ($p=0.059$; Holm 0.12), and noisy-input swap minus the doubled learning rate is $+9.7$ after 10,000 ($p=0.024$; Holm 0.072) and $+1.4$ after 17,500 ($p=0.41$). With three seeds this test has little power; intervals that resample pairs and seeds jointly exclude zero for the first three contrasts (Section~\ref{sec:experiments}). An independent confirmation uses six fresh seeds per arm on a fresh held-out split (Appendix~\ref{app:confirm}). From validation to the held-out test, the effects shrink from $+17.7$ to $+15.3$ (ordinary, 10,000 updates), $+13.7$ to $+7.3$ (ordinary, 17,500), and $+6.3$ to $+1.4$ (doubled learning rate, 17,500), as expected when the same validation pairs selected the recipe.

\paragraph{Readability-adjusted chance.}
A random-actor reference among readable outputs is $0.25$ times the fraction of the 150 ordinary pairs for which both outputs are answered. On \synclane{} it is 14.5\%, 12.8\%, and 12.8\% for seeds 0--2, beside observed switching of 16.0\%, 19.3\%, and 18.0\%. On layout it spans 17.5--18.5\%, beside 40.7--61.3\% observed. It assumes independent random actor choices in the two readable branches; it is a reference, not a measured control or a test.

\subsection{Coverage and answered accuracy}
Switch counts are end-to-end: an output the scorer cannot read counts as an error. Table~\ref{tab:coverage} separates readability (coverage) from depicted identity where it is readable (answered accuracy). Two-video's \synclane{} failure at 5,000 updates is an identity failure: answered accuracy is 50.0\% for seed~0 and below 60\% for all seeds. The remedies' gains are mostly identity, not readability: every reweighting arm and combination raises \synclane{} answered accuracy above ordinary training's 70.4\% (81.5--91.5\% on the first split; 86.8--94.6\% on the fresh split, against 81.0\% for LR$\times$2), and among the single arms Min-SNR raises coverage the most and answered accuracy the least. ReCamMaster fails on both counts. Actor-region PSNR on the 600 confirmatory outputs is 18.8~dB for Two-video seed~0, 18.5~dB for +InputHead, 19.4~dB for +CF, and 19.6~dB for Two-video-long.

\begin{table}[!htbp]
  \caption{Coverage and accuracy among answered outputs (\%) on ordinary pairs, 300 outputs per link. Two-video and Two-video-long: range over three seeds. $^\dagger$Two seeds.}
  \label{tab:coverage}
  \centering\footnotesize
  \setlength{\tabcolsep}{3pt}
  \begin{tabular}{@{}lcccc@{}}
    \toprule
    & \multicolumn{2}{c}{Layout} & \multicolumn{2}{c}{\Synclane{}} \\
    \cmidrule(lr){2-3}\cmidrule(l){4-5}
    Model & Coverage & Answered & Coverage & Answered \\
    \midrule
    Two-video & 77--84 & 74--89 & 65--74 & 50--60 \\
    Two-video-long & 90--94 & 94--97 & 85--88 & 85--89 \\
    Modular RGB & 92.0 & 94.9 & 75.7 & 83.7 \\
    ReCamMaster, 5k & 49.7 & 59.7 & 37.0 & 41.4 \\
    \multicolumn{5}{@{}l}{\emph{256 pixels, 10,000 updates, mean of three seeds}} \\
    Ordinary & 89.6 & 92.6 & 76.8 & 70.4 \\
    Ordinary, LR$\times$2 & 90.7 & 94.0 & 81.5 & 74.7 \\
    Noisy-input swap & 90.8 & 96.8 & 83.9 & 83.4 \\
    Min-SNR$^\dagger$ & 95.0 & 90.5 & 87.2 & 81.5 \\
    Patch w., simulator & 87.8 & 96.0 & 83.2 & 88.1 \\
    Patch w., MLLM boxes$^\dagger$ & 91.7 & 95.1 & 83.3 & 83.6 \\
    Swap + Min-SNR$^\dagger$ & 94.8 & 98.9 & 90.2 & 91.5 \\
    Swap + patch w.$^\dagger$ & 88.0 & 98.4 & 88.0 & 88.8 \\
    MLLM + Min-SNR$^\dagger$ & 93.8 & 93.8 & 88.8 & 88.2 \\
    \multicolumn{5}{@{}l}{\emph{Fresh split, 10,000 updates, mean of six seeds}} \\
    Ordinary, LR$\times$2 & 88.3 & 95.0 & 82.7 & 81.0 \\
    Noisy-input swap, LR$\times$2 & 86.8 & 97.7 & 87.7 & 89.8 \\
    Min-SNR & 90.9 & 94.7 & 88.4 & 86.8 \\
    Swap + Min-SNR & 92.2 & 99.5 & 92.1 & 94.6 \\
    \bottomrule
  \end{tabular}
\end{table}

\subsection{Harder benchmark variants}
\label{app:difficulty}
To map where trained models break, we built new held-out variants of the benchmark after the main results and evaluated them without retraining on the four arms of Table~\ref{tab:methodheldout}, at 10k (primary) and 17.5k updates (Table~\ref{tab:difficulty}). Each variant's design, gates and analysis were fixed before any of its model output, and every variant is reported whatever it shows. A variant is evaluated only if it passes the link gates (Appendix~\ref{app:linkgates}) and a scorer ceiling: the frozen scorer must be correct on at least 90\% of the true targets and of their 256-pixel autoencoder round trips, and must pass its calibration controls. Each variant is generated on one GPU model (L40S); a variant not complete by a cutoff fixed in advance would not have been analyzed. The 10k and 17.5k checkpoints of the fresh-split LR$\times$2 and noisy-input swap seeds had been deleted before these tests, so these two arms use training seeds 0 and 1 of the same recipe; the Min-SNR arms use two fresh-split seeds at the base rate, as in Table~\ref{tab:methodheldout}. Each arm has two seeds. A switch needs both branch outputs correct, so guessing between the two cubes gives 25\% of pairs, or 15--23\% after accounting for unreadable outputs; an arm that keeps one answer across branches scores below this level. Contrasts use a paired bootstrap over pairs, with the seed-resampled interval second; the primary contrast is swap + Min-SNR minus LR$\times$2 at 10k.

\emph{Relay, classifier-balanced.} Relay fails its gate in the main benchmark: the endpoint matcher reaches 62.0\% on its final pairs and learned position classifiers 71--84\% (Tables~\ref{tab:gatecounts} and~\ref{tab:links}). We built fresh relay pairs with the original generator, kept those on which the order oracle recovers the actor in both branches (a validity filter applied before any output), and selected 150 pairs (75 strike, 75 order) on which both classifiers are right on exactly half the items. The strike family needed the one pre-specified top-up of designs. On the selection, both classifiers score 50.0\%, the median-position, endpoint and start-to-parking matchers 52.7, 55.3 and 47.0\%, and the order oracle 100\%. The scorer is correct on 97.7\% of true targets and 90.3\% of round trips, a narrow pass. Relay is in every arm's training mix, so this is an in-distribution held-out test. Every arm is above guessing at both checkpoints. Swap + Min-SNR minus LR$\times$2 is $+24.0$ [$+16.7$, $+31.0$] [$+14.7$, $+34.0$] at 10k and $+12.3$ [$+6.3$, $+18.3$] [$+3.3$, $+20.7$] at 17.5k. Noisy-input swap minus LR$\times$2 is $+11.3$ [$+5.0$, $+17.7$] [$+1.3$, $+21.3$] and $+5.0$ [$-1.3$, $+11.7$] [$-3.3$, $+13.3$]. Min-SNR minus LR$\times$2 is $+8.7$ [$+2.0$, $+15.3$] [$-4.7$, $+22.0$] and $+5.0$ [$-2.0$, $+11.7$] [$-4.0$, $+13.7$]. As on layout and \synclane{}, the single changes' leads narrow by 17.5k.

\emph{Three cubes.} Scenes with three look-alike cubes. On layout, the scorer is correct on 89.5\% of round-tripped true targets (bar 90\%), mostly abstaining where it cannot detect contact at 256 pixels, so no model was evaluated. On \synclane{}, on a development batch, a position-only endpoint matcher reaches 75.0\%, outside the link gate's 40--60\% range, and the scorer is correct on only 83.7\% of true targets.

\emph{Gate events.} One cube drives through a colored gate while the other passes beside it, an event type absent from training (training uses strike and reach-first events). On \synclane{}, after the one pre-specified top-up of designs, the order oracle answers 271 of 300 items (90.3\%, bar 95\%; correct on all it answers), and the scorer's missing-patch control is false-correct on 5.7\% (bar 5\%), so this link fails. The layout link passes every check: the link oracle is correct on all items, the scorer is correct on 98.0\% of true targets and 97.3\% of round trips, and calibration passes. We decided to evaluate layout alone after the \synclane{} link failed, before any model output. All arms lose most of their trained-event accuracy (Table~\ref{tab:methodheldout}). Swap + Min-SNR minus LR$\times$2 is $+42.7$ [$+36.3$, $+49.3$] [$+32.0$, $+53.0$] at 10k and $+36.0$ [$+29.0$, $+43.0$] [$+28.7$, $+43.7$] at 17.5k. Noisy-input swap minus LR$\times$2 is $+18.3$ [$+12.3$, $+24.3$] [$+6.0$, $+30.7$] and $+18.7$ [$+12.7$, $+24.7$] [$+12.0$, $+25.3$]. Min-SNR minus LR$\times$2 is $+3.7$ [$-0.3$, $+7.7$] [$-2.7$, $+10.7$] and $+6.3$ [$+2.0$, $+11.0$] [$+0.7$, $+12.3$]. Noisy-input swap's seeds differ widely at 10k (34.7 and 20.0\%). Both outputs of a pair are readable on 83--94\% of pairs, so the below-guessing scores of LR$\times$2 and Min-SNR reflect wrong answers, not abstentions. This variant has no human check.

\subsection{ReCamMaster}
\label{app:recam}
\paragraph{Noisy-input swap on ReCamMaster.} Pre-specified on October 26, before any output: noisy-input swap transferred to ReCamMaster's flow matching. On training-square cells, with probability 0.5 at flow noise level $s\ge0.5$ ($\sigma\ge1$), the noisy input is built from the square sibling's target latent with the same noise, while the regression target is the velocity whose implied clean prediction is the cell's own target; ReCamMaster's own recipe otherwise (constant LR $10^{-5}$, 17,000 updates, seeds 0 and 1), held-out final pairs, generated on H100 (ordinary training: H100 and H200). Pooled switches are 48.3 and 55.7\% (layout 62.0\%, \synclane{} 42.0\%) against 33.0 and 32.7\% for ordinary training (46.0\% and 19.7\%): $+19.2$ [$+14.8$, $+23.7$] (primary; seeds resampled [$+12.0$, $+26.7$]), \synclane{} $+22.3$ [$+15.0$, $+29.7$] and layout $+16.0$ [$+11.0$, $+21.3$].

\paragraph{Copy probe on noisy-input swap.} On the same 42 training cells after 17,000 updates (a pre-specified secondary, descriptive), the number of cells that follow Video~1 at $\sigma=0.5,1,2,3.3,5,8$ is 0/24/41/41/41/42 and 0/28/40/40/39/39 for noisy-input swap's two seeds, against 0/0/1/13/29/35 and 0/0/3/15/32/33 for ordinary training. The mean score crosses zero at $\sigma\approx0.97$ and $0.91$, against $4.3$ and $4.5$. Swapping acts only at $\sigma\ge1$, and below that the model still copies.

 Trained for 17,000 updates, the update count of Two-video-long, ReCamMaster switches on 73/150 layout and 26/150 \synclane{} pairs (pooled 33.0\%; +22.7 over 5,000 updates, [+17.3, +28.3]), detects the event in 73.5\% of outputs, and remains 36.7 points below Two-video-long ([$-42.3$, $-31.0$]). On placement it reaches 12/150 and 14/150.

\subsection{Method study on validation}
\label{app:methodstudy}
On validation pairs, which also served to choose the recipe (three seeds per arm), noisy-input swap leads ordinary training by $+17.7$ points at 10,000 updates ([$+8.0$, $+27.0$], bootstrap over pairs and seeds) and LR$\times$2 by $+8.7$ ([$-0.3$, $+17.7$]), and reaches at 12,500 updates the accuracy LR$\times$2 reaches at 17,500; the midpoint variant behaves alike. Figure~\ref{fig:randomgrid} shows held-out pairs drawn at random.

\begin{figure}[!htbp]
  \centering
  \input{figures/emergence}
  \caption{\textbf{Correspondence learning at 576 pixels.} Validation switches along one ordinary-training run (50 pairs per link). Layout improves before \synclane{}; the gray band is a readability-adjusted guessing reference. The 10k dip is specific to this seed. Held-out three-seed results: main-text Table~\ref{tab:core}.}
  \label{fig:emergence}
\end{figure}

The method study compares 256-pixel arms on the 100 validation pairs (50 layout, 50 \synclane{}); each uses the ordinary recipe of Table~\ref{tab:compute} and changes one element.
 \emph{Learning-rate sweep.} Pooled validation switches of 100 for ordinary training at $4$, $8$, $12$ and $16\times10^{-6}$ are 50.7, 59.7, 60.0 and 55.0 after 10,000 updates (three, three, one and two seeds) and 67.0, 74.3 and 70.0 after 17,500 (no 17,500 checkpoint at $16\times10^{-6}$); noisy-input swap at $16\times10^{-6}$ (two seeds) gives 73.0 after 10,000, against 70.3 at $8\times10^{-6}$ (three seeds). Under the rule fixed before the sweep (move an arm to a higher learning rate only if it gains at least 3 switches), no arm changes.

\paragraph{Mechanism variants.}
Table~\ref{tab:mechanismfull} lists matched-update variants: each trains 10,000 updates on the same 18,000-update schedule as the method study, so it is comparable with the method study's 10,000-update checkpoints. The swap share is the fraction of all updates whose noisy input was swapped; at a matched 6\% share, swapping only at $1\le\sigma<5$ reaches 67 and 69 and swapping only at $\sigma\ge20$ 45 and 57. Extra noise on both patches hurts, while the same noise on a background region gives 61 and 51, consistent with (though not a test of) a model that copies whenever the patch looks clean. Continued to 17,500 updates, the low-noise variant reaches 82 and 71 (noisy-input swap: 77--86), so its advantage at 10,000 updates does not grow; on held-out scenes (three seeds, Table~\ref{tab:methodheldoutfull}) it leads the main recipe by $+3.2$ [$-0.1$, $+6.4$]. A constant learning rate after warmup (ordinary training, one seed) reaches 20, 28, 52, 53, 58, 54 and 56 of 100 pairs at 2,500 to 17,500 updates, against 12--31, 34--43, 37--49, 48--54, 50--61, 58--68 and 61--69 for the ordinary seeds.

\begin{table}[!htbp]
  \caption{Mechanism variants: pooled validation switches (of 100) after 10,000 updates of the 18,000-update schedule, seeds 0 / 1 ($^\ast$seeds 0 / 1 / 2). Descriptive: two seeds per row and no correction for the number of variants.}
  \label{tab:mechanismfull}
  \centering\footnotesize
  \setlength{\tabcolsep}{3pt}
  \resizebox{\columnwidth}{!}{\begin{tabular}{@{}lrr@{}}
    \toprule
    Variant & Swap share & Switches \\
    \midrule
    Ordinary (three seeds) & 0\% & 48 / 50 / 54 \\
    Noisy-input swap, $p=0.5$, $\sigma\ge1$ (three seeds) & 29\% & 70 / 64 / 71 \\
    \midrule
    Swap only at $1\le\sigma<5$ & 17\% & 75 / 76 \\
    Swap only at $5\le\sigma<20$ & 9\% & 59 / 60 \\
    Swap only at $\sigma\ge20$ & 3\% & 45 / 48 \\
    Swap on every square sample ($p=1$) & 58\% & 69 / 74 \\
    Sibling-free: own target, colors permuted & 45\% & 77 / 72 \\
    Swap at $1\le\sigma<5$, matched share & 6\% & 67 / 69 \\
    Swap at $5\le\sigma<20$, matched share & 6\% & 55 / 62 \\
    Swap at $\sigma\ge20$, matched share & 6\% & 45 / 57 \\
    \midrule
    Extra noise on both patches & 0\% & 29 / 36 \\
    Same noise on a background region & 0\% & 61 / 51 \\
    Patch loss weight ($1{+}4\times$mask)$^\ast$ & 0\% & 73 / 67 / 64 \\
    Input perturbation~\cite{ning2023inputperturbation} & 0\% & 53 / 56 \\
    \midrule
    Schedule shift, $\sigma\times2$ & 0\% & 53 / 43 \\
    Schedule shift, $\sigma\times4$ & 0\% & 48 / 42 \\
    5$\times$ more pure-noise draws & 0\% & 49$^\ast$ / 54 \\
    Min-SNR-5 weighting & 0\% & 64 / 65 \\
    \bottomrule
  \end{tabular}}
  \\[2pt]{\scriptsize $^\ast$seed 0 is the 18,000-update screen run, evaluated at 10,000.}
\end{table}

\paragraph{Answer corruption versus added noise.}
The task separates changes in generated correspondence from changes in copying. Exploratory validation tests compare noisy-input changes with loss reweighting. Among tested input changes, removing the answer helps most. Schedule shifts, extra pure-noise draws and input perturbation do not help; extra noise on both patches lowers accuracy despite hiding the answer (Appendix Table~\ref{tab:mechanismfull}). Among probed variants, only noisy-input swap and midpoint stop copying at $\sigma=2$; reweighting helps but copying persists.

\paragraph{Benefits at noise levels where models copy.}
Swapping only at $1\le\sigma<5$ gives higher validation scores than the selected recipe. Swapping only at $\sigma\ge20$ does not help, even when both variants swap 6\% of draws (Appendix Table~\ref{tab:mechanismfull}). Noisy-input swap also increases sensitivity to Video~1 at low noise (Appendix Table~\ref{tab:v1dep}). Because the probe resembles noisy-input swap's training inputs, it characterizes learned behavior without identifying why learning is faster.

\subsection{Real multi-camera video}
\label{app:panopticb}

\paragraph{Data and training.} We use windows of 41 high-definition frames (about 4\,s) from social games recorded by four HD cameras in the CMU Panoptic studio~\cite{joo2019panoptic}. We undistort the videos and train and generate at $208\times384$. Each training sample takes one camera as the target and two of the other three as input videos. Training uses 23 haggling sequences and the test 17 other sequences (the SSP test list and 2016 social games). Every arm initializes a three-view model from released AnyView and trains 10,000 updates of an 18,000-update schedule at learning rate $8\times10^{-6}$, two seeds each. Arms: ordinary training and Min-SNR ($\gamma=5$).

\paragraph{Test and scorers.} For each of 454 test items, one person A's garment is recolored (hue $+150^\circ$, chroma $+40$, lightness kept) in the input videos where A is visible, and the model generates the target view from the edited and from the unedited inputs with the same seed. Per frame, a frozen scorer compares A's median garment color in the two generations with the shift the same edit causes in the true target; an item counts as \emph{correct binding} when most decisive frames show A's garment edited and no other person's. The scorer was checked on recolored ground-truth targets before any training output was inspected. \emph{Natural binding}, on the unedited generations, is the share of people whose true garment colors differ by at least 10 CIE76 units from everyone else's and whose generated garment color is nearest their own (329 items).

\paragraph{Results.} After 5,000 updates, the primary checkpoint, correct binding is 23.0\% for ordinary training and 44.6\% with Min-SNR, and natural binding 56.1 and 74.7\%; Min-SNR adds $+21.6$ [$+15.1$, $+27.3$] and $+18.7$ [$+11.9$, $+26.7$] (pre-specified; natural binding on the items scored for every arm in Figure~\ref{fig:panoptic}). Intervals resample sequences as clusters and, per arm, seeds. After 10,000 updates, correct binding is 33.3 and 51.0\% and natural binding (327 items) 62.0 and 79.3\%; Min-SNR adds $+17.7$ [$+13.7$, $+22.5$] and $+17.3$ [$+9.9$, $+25.7$].

\paragraph{Attribute propagation and placement.}
The second test recolors one person's garment in both inputs and checks that only that person changes in the output (Figure~\ref{fig:panoptic}). The scorer reads the person's true location, so success also requires correct placement. A \emph{position-tolerant} score credits the edit wherever the person is rendered. Min-SNR improves this score by 10.9 points at 5k updates ([$+7.1$, $+15.3$]; two seeds), about half its gain under the original score; the rest reflects placement.

\paragraph{Position-tolerant re-scoring.} Generated people are sometimes rendered away from their true position, which both scores above count as failures. A re-scoring, specified before it was computed, separates placement from binding. On the generation from unedited inputs, A is \emph{in place} when A's true-target region shows A (its color is nearest A's true garment color among the item's people). The position-tolerant score takes the change between the generations from edited and unedited inputs (same seed) in the edit's direction, and counts a hit when it lands mostly on A's generated counterpart, identified by garment color, and a leak when it lands on another person.

On VAE round trips of the true videos, A is in place on 99.1\% of items and the tolerant score is correct on 97.3\%. It covers 450 of the 454 items (900 outputs per arm over two seeds; A is identifiable by color on 250 items).

After 5,000 updates, A is in place in 46.2\% of ordinary outputs and 65.9\% with Min-SNR (10,000: 55.6 and 71.9\%); tolerant correct binding is 59.7 and 70.6\% (10,000: 65.7 and 73.0\%), so Min-SNR adds $+10.9$ [$+7.1$, $+15.3$] ($+7.3$ [$+2.9$, $+12.7$] after 10,000), against $+21.7$ and $+18.0$ for the region-based score on the same items; on items where A is identifiable, $+9.6$ [$+3.6$, $+17.2$] and $+7.8$ [$+2.2$, $+15.2$]. Figure~\ref{fig:panoptic} also shows the two sibling-world arms with Min-SNR (described below), which also make the test edit familiar: with and without noisy-input swap, correct binding after 5,000 updates is 70.7 and 60.5\% and tolerant correct binding 76.0 and 72.1\%; swap minus control is $+10.2$ [$+1.8$, $+18.7$] and $+3.9$ [$-0.8$, $+8.6$].

For this propagation task, we train on paired versions with one garment recolored consistently in all views. Noisy-input swap stops copying the planted edit at every tested $\sigma\ge1$, unlike a control trained on the same versions without swapping.

\paragraph{Noisy-input training and the copy probe.} Two further arms train against the copy route. \emph{Sibling worlds}: one person A (at least 2,000 garment pixels in the target and 1,000 in an input) is recolored consistently in all three views (hue rotated by $90^\circ$--$270^\circ$, chroma pushed by 40), and each sample comes from this world or the original with probability 0.5; noisy-input swap then builds the noisy input, with probability 0.5 at $\sigma\ge1$, from the other world's target while regressing to its own, and a control trains on the same worlds without swapping (both also with Min-SNR).

A pre-specified copy probe on 40 held-out episodes plants the recolored garment in the noisy input while the inputs show the original, and reports $s=\langle\hat x_0-(z+z_e)/2,\,d\rangle/\|d\|^2$ ($-0.5$: copies the noisy input; $+0.5$: follows the inputs), averaged over seeds and over $\sigma\in\{1,2,3.3,5\}$.

Noisy-input swap minus its control is $+0.495$ [$+0.415$, $+0.573$] (primary; $+0.40$ on camera triples seen in training, $+0.55$ on unseen ones), and $+0.55$ [$+0.46$, $+0.63$] with Min-SNR, and Min-SNR minus ordinary training $-0.004$ [$-0.019$, $+0.008$].

Ordinary training copies up to $\sigma\approx1.7$ (released AnyView, re-run at the fine-tuning resolution of 384 pixels: 3.75); the noisy-input arms copy at no tested $\sigma\ge1$. The probe's edit is the kind of corruption these arms train against, so it measures the invariance they were trained for, on held-out episodes and cameras.

\paragraph{Evidence split across both videos.} In the tests above, either input can carry the edit to the right person. A further test, pre-specified before any of its output, needs both. Each item takes a 41-frame window, three cameras and two people. Video~1 shows each person in a square crop that follows their garment over time, the two $192\times192$ crops side by side in an order drawn at random per item and on a fixed placeholder camera, so slot order and camera pose carry no scene position; the other selected person's garment pixels inside a crop have their color removed. Video~2 is the real footage of the second camera in grayscale, identical across branches. The target is the third camera. Both people's garments are recolored with two hues $180^\circ$ apart, lightness kept, and the second branch exchanges the two colors in Video~1 and the target. Noisy-input swap builds the noisy input from the other branch with probability 0.5 at $\sigma\ge1$ while regressing to its own.

Position-only matchers that read slot order score 45.9 and 50.5\% on the test set. Garment brightness, visible in both videos, links the people on 84.0\% of eligible items, so the primary set takes, within each sequence, every item on which a brightness-only matcher is wrong and an equal number on which it is right (388 items from 13 of the 17 test sequences, none above 15\%), and training gives the two kinds equal weight (798 windows from 20 training sequences). Faces, trousers, body shape and motion timing remain visible in both videos, so the test measures binding across videos linked mainly by appearance, not same-moment correspondence alone. The other selected person is sometimes visible at the edge of a crop; a matcher that uses this cue is right on 47.8\% of items.

In each output, color blobs are assigned to the person whose true-target position is nearest, within that person's garment width. An item is correct when both people carry their own color in both branches, and \emph{placed} when both people receive a blob in both outputs, whatever its color. On autoencoded true targets the scorer is correct on all 388 items without abstention. This rule replaced a color-assignment rule before any model output, because balancing selects look-alike garments on which color assignment often abstains; since it still credits partly correct placement, placement and accuracy among placed items are reported beside the primary.

All four arms (ordinary training, noisy-input swap, Min-SNR, and noisy-input swap + Min-SNR) use learning rate $8\times10^{-6}$; the 5,000-update checkpoint was fixed on three held-out development sequences by a rule set in advance. Each arm has two seeds, reduced from three before any output when the cluster queue stalled, and all outputs are generated on one GPU model. Correct binding is 15.9\% for ordinary training (seeds: 17.8 and 13.9) and 25.4\% for noisy-input swap (23.5 and 27.3), a gain of $+9.5$ [$+4.4$, $+14.8$] (primary; intervals resample sequences and seeds). Placement does not differ ($-2.3$ [$-13.0$, $+6.2$]; 43.8 and 41.5\% of items placed), and among placed items accuracy rises from 31.8 to 55.9\% ($+24.1$ [$+15.0$, $+33.2$]), so better placement does not explain the gain. The scorer abstains, mostly where too little color is transferred, on 51.9\% for ordinary training and 20.6\% for noisy-input swap. On the 157 items where color assignment applies, the gain is $+19.4$ [$+10.1$, $+30.5$]. It holds without the neighbor cue ($+11.5$ [$+6.7$, $+16.8$]; 249 items) and when the brightness matcher is right ($+8.5$ [$+0.6$, $+15.9$]) or wrong ($+10.6$ [$+3.3$, $+17.6$]; 194 items each). At the same checkpoint, Min-SNR reaches 35.4\% and noisy-input swap + Min-SNR 40.7\%. Min-SNR's gain over ordinary training is $+19.6$ [$+11.5$, $+27.7$]; as with noisy-input swap, it lies among placed items ($+30.0$ [$+20.8$, $+39.1$]) rather than in placement ($+7.1$ [$-2.6$, $+16.9$]). It holds on the natural mix ($+12.5$ [$+1.0$, $+27.5$]; 60 items), without the neighbor cue ($+24.5$ [$+15.2$, $+33.6$]) and when the brightness matcher is wrong ($+19.1$ [$+8.5$, $+28.5$]). Absolute rates remain low.

\subsection{A real-video handoff check}
\label{app:real}

To test whether integration carries over to real video without new labels, we built 200 handoffs from the Argoverse~2 sensor validation set~\cite{wilson2021argoverse}: a vehicle seen by ring camera A and later by camera C, with a same-category look-alike seen by both, and the vehicle never visible to a third camera B. Camera A serves as Video~1, B as Video~2, and C as the target (41 frames at 10\,Hz); the generators run zero-shot. DINOv2~\cite{oquab2024dinov2} crops of the generated target view at the annotated box are compared with Video-1 references of the vehicle and of each look-alike. An item counts only if the same rule picks the right vehicle on the real target frames (51 of 200 items). Among items with usable outputs, the 5,000-update model shows the right vehicle in 16 of 40, the long-trained model in 15 of 40, and the long-trained model given another clip's Video~1 in 12 of 40, all with negative mean margins. We see no identity transfer, although the check is small and the domain gap large.

\subsection{Single-video fidelity}
\label{app:fidelity}
Our three-stream wrapper, fed the single benchmark input twice, reproduces the released AnyView checkpoint on the 176 AnyViewBench episodes: PSNR differs by +0.16~dB (per-episode standard deviation 0.59~dB) and LPIPS by $-0.013$; the reference run matches the published Argoverse and DDAD numbers. With the same harness, no Kubric fine-tune lowers PSNR below the released model on Argoverse, DDAD or DROID; on AssemblyHands, Two-video does (9.89 against 10.15~dB), and LPIPS changes stay within 0.02. Noisy-input swap leaves single-video fidelity unchanged (PSNR 11.92 against 11.89~dB for ordinary training, averaged over the four subsets, 17,500 updates). A rule fixed before the Min-SNR results existed compared swap + Min-SNR with Min-SNR and required PSNR and LPIPS to agree; they disagree, so we claim no fidelity difference between the two. These are single-video measurements, not identity outcomes.

\subsection{Generated examples}
\label{app:qual}
\begin{figure*}[!tp]
  \centering
  \input{figures/random_grid}
  \caption{The first four of eight randomly drawn held-out \synclane{} pairs (256 pixels, 10,000 updates, same starting noise; zoom on the true actor at contact $+4$). Border: frozen scorer's call (green right, red other candidate, orange invented patch, gray abstain; no border on ground truth).}
  \label{fig:randomgrid}
\end{figure*}

The held-out pair of Figure~\ref{fig:teaser} is the first \synclane{} pair, in item order, that the frozen scorer decisively calls right in both branches for noisy-input swap and wrong in at least one for ordinary training (256 pixels, 10,000 updates, seed~0, same starting noise). Figure~\ref{fig:randomgrid} shows the first four of eight held-out \synclane{} pairs drawn at random (fixed seed), without conditioning on any outcome. Noisy-input swap is right where ordinary training abstains or picks the other cube in 3B; both fail in 2; and noisy-input swap draws an invented patch in 3A and 4A, where ordinary training abstains or picks the other cube. In the other four pairs of the draw, noisy-input swap is right where ordinary training fails in 5A, 6A, 6B and 8B, and both fail in 7B.

\oldappsection{Other Interventions and Negative Results}
\label{app:other}
The pre-specified method of this study was the counterfactual residual objective +CF. It failed its compute-matched criterion. Noisy-input swap was developed afterwards, from the diagnosis of Section~\ref{sec:diagnosis}, and is tested under its own protocol (Appendix~\ref{app:confirm}). This section documents +CF and the other interventions we tried.

\subsection{The counterfactual residual objective (+CF)}
\label{app:cf}

+CF, the study's first method, adds 5,000 auxiliary updates, alternating with 5,000 denoising updates, that train the difference between the clean latents predicted for two inputs along one square edge, from a shared pure-noise target, to match the difference between their targets. It was to count as a method gain only if, against the 5,000-update Two-video model, the pooled switch interval excluded zero, all-four accuracy rose, event detection fell by at most five points, the run exceeded the Two-video seed range, and it also beat compute-matched Two-video-long.

It meets the first four ($+14.0$ [$+8.3$, $+19.7$]; 42 against 23 of 100 squares; 87.0 against 85.2\% event detection; 52.7\% against 30.0--38.7\%) and fails the last: Two-video-long at 17,000 updates is better by 17.0 points ([$-23.0$, $-10.7$]; \synclane{} $-26.0$ [$-35.3$, $-16.7$]) and by 25 all-four squares ([$-34$, $-16$]).

The pre-specified plan stated that a failure would make the paper a benchmark and failure analysis, not a method paper. Its two additional training seeds were not run.

\subsection{Zero-shot generators and shuffled inputs}
\label{app:inputhead}
Zero-shot AnyView, NoPo4D and LiveStre4m do not render this camera separation (at most 1 of 88 switches on a development pilot), which is a renderability limit and not an identity result. Shuffling Video~2 collapses switching, which shows dependence on the second input and not correct identity use: the Two-video-long model with shuffled Video~2 switches on 2/50 layout and 1/50 \synclane{} pairs.

\subsection{Summary of other interventions}
\paragraph{Other interventions.} Results are pooled switches of the 100 validation pairs unless marked held-out; validation searches are development evidence, since the same pairs were used for selection. At 576 pixels, +CF (Appendix~\ref{app:cf}) gains $+14.0$ held-out over Two-video seed~0 ($+17.0$ over its seed mean) but trails compute-matched Two-video-long by $-17.0$ [$-23.0$, $-10.7$]; +InputHead, an event head trained on separate steps, gives $+2.3$ [$-4.8$, $+10.0$] held-out against the three-seed mean. On the 5,000-update recipe, Video-1 dropout with guidance $w\in\{1,2,3,5\}$ gives 34, 39, 42 and 46 against 44 (seed~0; best $+2.0$ [$-7.0$, $+11.0$]), and motion-weighted denoising a mean of 46.0 against 38.0 over three seeds (seed~0 $+8.0$ [$-1.0$, $+17.0$]). An oracle evidence sketch at the target camera as a fourth stream reaches 74 after 2,000 more updates against 42 for 2,000 ordinary updates, matching Two-video-long ($+2.0$ [$-9.0$, $+13.0$]) with privileged input; a sketch perceived by the modular pipeline reaches 62 ($-10.0$ [$-21.0$, $+1.0$] against Two-video-long). At inference, on the 17,000-update model (72 unguided), starting the sampler at $\sigma=1000$ gives 68 and weight extrapolation away from the 5,000-update checkpoint 74 ($+2$ [$-8$, $+12$]; worse from earlier checkpoints); autoguidance with the 7,500-update checkpoint gives 51--54 against 54 unguided. A label-free learned track association for the modular reference reaches 65 of 100 \synclane{} associations against 77 for the fixed rule. Zero-shot, a ProcTHOR contents task fails because the outputs miss the steep target camera on all 10 pairs, and Argoverse~2 handoffs show the right vehicle in 16/40 (5k), 15/40 (17k) and 12/40 (17k, wrong Video~1; Appendix~\ref{app:real}). The 256-pixel training variants are in Table~\ref{tab:mechanismfull}.

\oldappsection{Protocol Deviations}
\label{app:deviations}
Table~\ref{tab:deviations} lists every change to the pre-specified protocol, when it was made relative to the results it could affect, and why. Two pre-specified secondary analyses are omitted from this supplement for length, and some secondary contrasts are summarized; all will be released with the protocol log. The omitted analyses are sampler splicing, which locates each method's gain within the sampling trajectory, and an edit-strength test of the Panoptic propagation task.

\begin{table*}[!tp]
  \caption{Protocol deviations.}
  \label{tab:deviations}
  \centering\scriptsize
  \renewcommand{\arraystretch}{1.0}
  \begin{tabular}{@{}>{\raggedright\arraybackslash}p{0.33\textwidth}>{\raggedright\arraybackslash}p{0.29\textwidth}>{\raggedright\arraybackslash}p{0.32\textwidth}@{}}
    \toprule
    Change & When & Why \\
    \midrule
    Human checks completed by one or two raters, although at least three were specified & After rating began & Rater time; agreement between raters is reported for one part only \\
    \addlinespace[1pt]
    Midpoint and sibling-free recolor added to held-out evaluation as controls, outside the four pre-specified contrasts & After those contrasts were fixed & To compare noisy-input swap with simpler ways of removing the answer \\
    \addlinespace[1pt]
    Relay excluded from confirmatory pooling; reported as secondary and analyzed per item & After its final selections failed the link gate; before any model output & Endpoint matcher at 62\%; rebalancing would spend its independence as a check \\
    \addlinespace[1pt]
    Training set of 1,742 instead of 1,800 bases (842 squares) & Before training & Layout design pool exhausted; squares with truncated target videos removed \\
    \addlinespace[1pt]
    Placement scorer repaired (no silhouette erosion) and recalibrated on fresh pairs & After its first calibration failed on \synclane{}; before any placement output was inspected & Erosion removed the lower patch band in small target views \\
    \addlinespace[1pt]
    Compute matching on one device scale with a median-based conversion factor; 16,500 updates added as a sensitivity checkpoint & After training moved between GPU models; before any +CF or Two-video-long output was scored & Mixed hardware; a slowdown episode distorted mean-based timing \\
    \addlinespace[1pt]
    +CF trained with one seed instead of three & Before the extra seeds started & Compute; their launch was conditional on the training check, which failed \\
    \addlinespace[1pt]
    Modular reference repaired (placement-only fallback) and re-evaluated on the same placement selection & After the first version's final placement run & Unassociated layout tracks; color-attribute records unchanged \\
    \addlinespace[1pt]
    Human scorer check limited to the first 100 items of each shuffled part (200 of 416 per rater) & After the rating set was built & Rater time; the random order keeps each block a category-stratified sample \\
    \addlinespace[1pt]
    Time-shift test reduced to the delay, its gate to ``no correct same-moment evidence'' & After the gate, before any generation & The gate showed that a shift contradicts the same-moment evidence rather than removing it, and that advancing Video~1 hides most layout patches \\
    \addlinespace[1pt]
    Time-synchronous matcher and per-item relay analysis (shortcut-incongruent items) added & After the confirmatory runs & To identify the evidence that links \synclane{}, and to analyze relay despite its gate failure \\
    \addlinespace[1pt]
    Primary method changed from +CF to noisy-input swap & After +CF failed its compute-matched criterion; recipe selected on validation and frozen before any held-out result of the 256-pixel runs & Diagnosis of Section~\ref{sec:diagnosis}; protocol in Appendix~\ref{app:confirm} \\
    \addlinespace[1pt]
    Method study at 256 instead of 576 pixels, evaluated at 17,500 instead of 17,000 updates & Before the method study & 256 pixels passed a tokenizer round-trip gate and trains about 7$\times$ faster per update; checkpoints saved every 2,500 updates \\
    \addlinespace[1pt]
    Secondary arm (noisy-input swap at LR$\times$2) also evaluated at 10,000 updates and extended to three seeds & After the primary held-out results & Its single seed was the strongest validation curve \\
    \addlinespace[1pt]
    Low-noise noisy-input swap ($1\le\sigma<5$) given its own held-out check & After the primary held-out results; selected on validation (mechanism table) & Reported as a post-hoc variant, not as the pre-specified method \\
    \addlinespace[1pt]
    Seed-resampling intervals added to the primary contrasts & After the primary results & Pre-specified intervals resample pairs only \\
    \addlinespace[1pt]
    Real-video split-evidence test: nearest-person scoring instead of color assignment; two seeds instead of three; generation on one GPU model other than the planned one & Before any model output & Color assignment abstains on the look-alike garments that brightness balancing selects; the cluster queue stalled \\
    \bottomrule
  \end{tabular}
\end{table*}

\oldappsection{Further Limitations}
\label{app:positioning}
\label{app:morelimits}
\emph{Method.} Synthetic siblings require simulator renders; the sibling-free recolor relies on the palette's symmetry and transfers less to patch placement than noisy-input swap, and adding Min-SNR reduces noisy-input swap's transfer to placement. The recipe was selected on validation after +CF failed.

\emph{Analysis.} Relay fails its validity check, and static classifiers identify the actor on 71--84\% of relay and 79--87\% of \synclane{} items, so the validity checks exclude only the specified rules. On fresh sets where two classifiers are at chance, all arms remain above guessing and the benefit of longer training replicates; there, noisy-input swap's lead over LR$\times$2 is not resolved once seeds are resampled, while adding Min-SNR gives a resolved gain over noisy-input swap. The time-synchronous matcher was introduced after the confirmatory runs, and the handoff readout uses one seed.

\emph{Learning trajectory.} At 17,500 updates, noisy-input swap at LR$\times$2 exceeds ordinary training at the same rate (78.9 versus 72.2\%, a secondary arm; fresh split 77.6 versus 72.4\%); on a 40k schedule the difference is small (80.8 versus 80.3\%) and equivalence is inconclusive. Whether noisy-input swap simply accelerates ordinary training's trajectory is unresolved.

\end{document}

%% file: preamble.tex
\usepackage{microtype}

%% file: figures/teaser.tex
\setlength{\tabcolsep}{0.8pt}
\newcommand{\rlab}[2]{\raisebox{-.5\height}{\rotatebox{90}{\parbox{.118\textwidth}{\centering\textbf{#1}\\[-1pt]\scriptsize #2}}}}
\newcommand{\gap}{\hspace{5pt}}
\begin{tabular}{@{}c@{\hspace{3pt}}c@{\gap}c@{\gap}c@{\gap}ccc@{}}
  & \textbf{Video 1} & \textbf{Video 2} & \textbf{Target view} & \multicolumn{3}{c}{\textbf{Acting cube (zoom)}} \\[-1pt]
  & {\scriptsize patches visible} & {\scriptsize which cube acts} & {\scriptsize ground truth} & {\scriptsize ground truth} & {\scriptsize ordinary} & {\scriptsize noisy-input swap} \\[2pt]
  \rlab{A}{yellow-patch cube acts} & \raisebox{-.5\height}{\includegraphics[width=.118\textwidth]{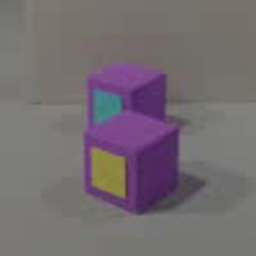}} & \raisebox{-.5\height}{\includegraphics[width=.118\textwidth]{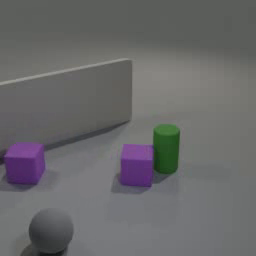}} & \raisebox{-.5\height}{\includegraphics[width=.118\textwidth]{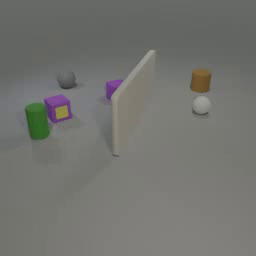}} & \raisebox{-.5\height}{\includegraphics[width=.118\textwidth]{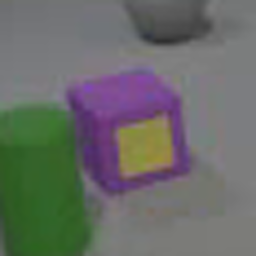}} & \raisebox{-.5\height}{\includegraphics[width=.118\textwidth]{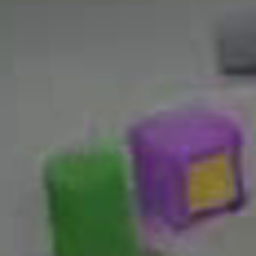}} & \raisebox{-.5\height}{\includegraphics[width=.118\textwidth]{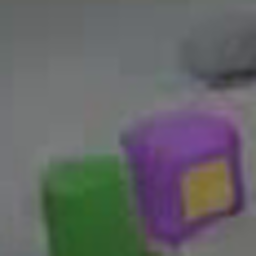}} \\[2pt]
  \rlab{B}{cyan-patch cube acts} & \raisebox{-.5\height}{\includegraphics[width=.118\textwidth]{figures/teaser_heldout/v1.png}} & \raisebox{-.5\height}{\includegraphics[width=.118\textwidth]{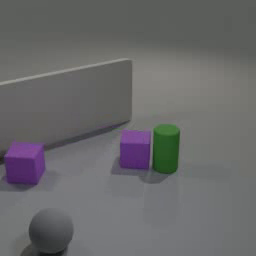}} & \raisebox{-.5\height}{\includegraphics[width=.118\textwidth]{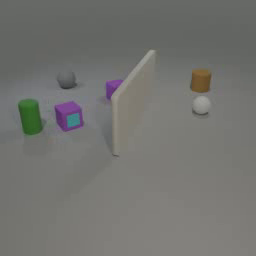}} & \raisebox{-.5\height}{\includegraphics[width=.118\textwidth]{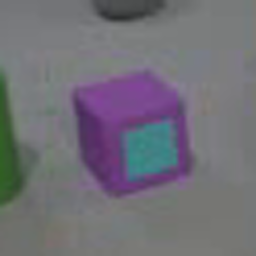}} & \raisebox{-.5\height}{\includegraphics[width=.118\textwidth]{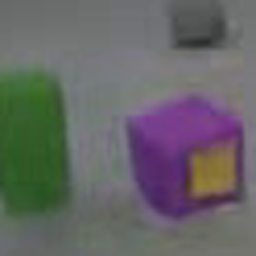}} & \raisebox{-.5\height}{\includegraphics[width=.118\textwidth]{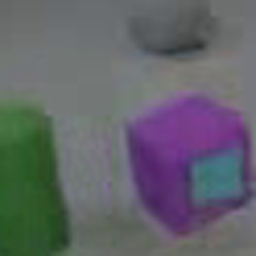}} \\
\end{tabular}

%% file: figures/teaser_mechanism.tex
\begin{tikzpicture}[>=latex, font=\scriptsize,
  img/.style={inner sep=0pt, draw=gray!50, line width=0.3pt},
  lab/.style={font=\scriptsize, align=center, inner sep=1pt},
  den/.style={draw=black!70, rounded corners=2pt, fill=gray!8, minimum width=1.0cm, minimum height=0.8cm, align=center},
  ttl/.style={anchor=west, font=\footnotesize, inner sep=0pt}]
\def\imw{1.55cm}
\foreach \xs/\ttl/\yimg/\ylab/\note/\notecol/\frame in {%
  0/{\textbf{Ordinary training:} $y$ already shows the answer}/cell_s1_zoom/{(true patch visible)}/{copying $y$ lowers the loss; Video~1 is not needed}/black!75/gray!50,%
  7.6/{\textbf{\textcolor{xswap}{Noisy-input swap:}} $y$ from the Video-1 sibling}/sibling_s1_zoom/{(sibling's patch)}/{copying $y$ raises the loss; the patch must come from Video~1}/xswap!85!black/xswap} {
  \begin{scope}[xshift=\xs cm]
    \node[ttl] at (-0.78,1.45) {\ttl};
    \node[img, draw=\frame, line width=0.7pt] (y) at (0,0) {\def\ycell{cell_s1_zoom}\edef\ycur{\yimg}\ifx\ycur\ycell\includegraphics[width=\imw]{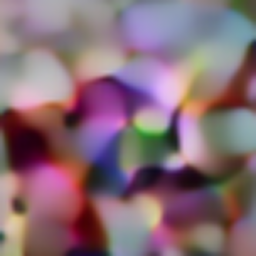}\else\includegraphics[width=\imw]{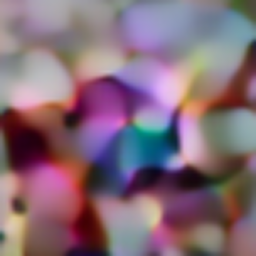}\fi};
    \draw[black!75, dashed, line width=0.45pt] ([xshift=0.11cm,yshift=-0.10cm]y.center) circle (0.21cm);
    \node[lab, below=1pt of y] {noisy input $y$\\\ylab};
    \node[den] (d) at (2.25,0) {$D_\theta$};
    \node[lab, text=gray!70!black] (c) at (2.25,1.0) {Video 1, Video 2};
    \draw[->, gray!70] (c.south) -- (d.north);
    \node[img] (z) at (4.5,0) {\includegraphics[width=\imw]{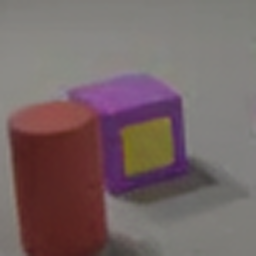}};
    \node[lab, below=1pt of z] {regression target $z$\\(true patch)};
    \draw[->, line width=0.6pt] (y.east) -- (d.west);
    \draw[->, line width=0.6pt] (d.east) -- (z.west);
    \node[lab, text=\notecol, text width=2.5cm] at (2.25,-0.95) {\note};
  \end{scope}
}
\draw[gray!40, line width=0.4pt] (6.55,-1.35) -- (6.55,1.55);
\end{tikzpicture}

%% file: figures/lane_scene.tex
\begin{tikzpicture}[font=\scriptsize, x=1cm, y=1cm, >=Latex,
  cam/.style={draw=black!75, fill=white, thick, minimum width=0.5cm, minimum height=0.28cm, inner sep=0pt},
  cube/.style={draw=black!65, fill={rgb,255:red,165;green,95;blue,200}, minimum size=0.32cm, inner sep=0pt},
  bigcube/.style={draw=black!65, fill={rgb,255:red,165;green,95;blue,200}, minimum size=0.52cm, inner sep=0pt},
  lab/.style={black!80},
  ttl/.style={font=\footnotesize\bfseries, anchor=base west, inner sep=0pt}]
  \newcommand{\patch}[2]{\fill[#2] ($(#1)+(-0.09,-0.15)$) rectangle ($(#1)+(0.09,-0.05)$);}
  \newcommand{\bigpatch}[2]{\fill[#2] ($(#1)+(-0.15,-0.23)$) rectangle ($(#1)+(0.15,-0.06)$);}
  \node[ttl] at (0,6.62) {(a) \Synclane{} scene (top-down)};
  \fill[black!4] (0,0) rectangle (10.4,6.4);
  \fill[black!50] (1.0,2.75) rectangle (7.0,3.0);
  \node[anchor=north west, lab] at (1.0,2.73) {wall};
  \node[anchor=south west, lab] at (0.08,0.05) {front of the wall};
  \node[anchor=south west, lab, inner sep=1pt] at (0.08,3.03) {behind the wall};
  \draw[xtiming, thick, rounded corners] (1.5,3.3) rectangle (9.0,6.3);
  \node[xtiming!80!black, anchor=north east, align=right] at (8.95,6.25) {Video 2 sees:\\wall end + behind};
  \draw[black!70, thick, rounded corners] (4.2,0.3) rectangle (8.6,3.5);
  \node[lab, anchor=south west] at (4.25,0.33) {Video 1 sees: patches, departures};
  \node[cube] (c1) at (5.4,1.05) {}; \patch{c1}{cyan}
  \node[cube] (c2) at (5.4,1.8) {};  \patch{c2}{yellow!85!black}
  \draw[->, very thick, black!70, rounded corners=7pt] (5.75,1.42) -- (8.0,1.42) -- (8.0,4.05) -- (6.6,4.05);
  \node[anchor=west, lab, align=left] at (8.65,1.42) {one lane,\\single file};
  \fill[xtiming, opacity=0.22] (8.0,3.4) ellipse (0.5cm and 0.38cm);
  \draw[xtiming!80!black, thick] (8.0,3.4) ellipse (0.5cm and 0.38cm);
  \draw[xtiming!80!black, thick] (8.5,3.4) -- (9.12,3.4);
  \node[anchor=west, xtiming!75!black, align=left, font=\scriptsize\bfseries, inner sep=1pt] at (9.12,3.4) {both videos\\see the cubes\\here at the\\same moment};
  \node[cube, draw=black!60, dashed, fill=white] (p1) at (4.3,4.6) {};
  \node[cube, draw=black!60, dashed, fill=white] (p2) at (5.4,4.85) {};
  \draw[->, thick, black!60, dashed] (6.55,4.08) -- (p1.east);
  \draw[->, thick, black!60, dashed] (6.55,4.08) -- (p2.south east);
  \node[anchor=south, lab] at (4.85,5.1) {parking spots (random order)};
  \fill[red!65!black] (2.4,4.9) circle (0.17);
  \draw[->, thick, red!65!black] (p1.west) -- (2.62,4.86);
  \node[anchor=south, red!65!black, font=\scriptsize\bfseries] at (2.4,5.1) {event: strike};
  \node[cam] (v1cam) at (6.3,0.0) {};
  \node[anchor=west, lab] at (v1cam.east) {Video-1 camera};
  \draw[->, black!70, thick] (6.3,0.14) -- (6.3,0.42);
  \node[cam] (v2cam) at (5.2,6.72) {};
  \node[anchor=west, lab] at (v2cam.east) {Video-2 camera};
  \draw[->, black!70, thick] (5.2,6.58) -- (5.2,6.18);
  \node[cam] (tcam) at (0.6,4.75) {};
  \node[anchor=north, lab] at (tcam.south) {target camera};
  \draw[->, black!70, thick] (0.88,4.78) -- (2.18,4.88);
  \node[ttl] at (10.8,6.62) {(b) Layout};
  \node[anchor=east, lab] at (12.15,5.92) {Video 1};
  \node[anchor=east, lab] at (12.15,4.97) {Video 2};
  \draw[black!65, thick, rounded corners=1.5pt, fill=black!4] (12.25,5.57) rectangle (14.65,6.27);
  \draw[black!65, thick, rounded corners=1.5pt, fill=black!4] (12.25,4.62) rectangle (14.65,5.32);
  \node[cube] (l1) at (12.85,5.97) {}; \patch{l1}{cyan}
  \node[cube] (l2) at (13.85,5.97) {}; \patch{l2}{yellow!85!black}
  \node[cube] (m1) at (12.85,4.97) {};
  \node[cube] (m2) at (13.85,4.97) {};
  \fill[red!65!black] (14.38,4.97) circle (0.11);
  \draw[->, thick, red!65!black] (m2.east) -- (14.25,4.97);
  \draw[xlayout, thick, dashed] (l1.south) -- (m1.north);
  \draw[xlayout, thick, dashed] (l2.south) -- (m2.north);
  \node[anchor=west, align=left, xlayout!85!black] at (14.75,5.45) {same arrangement\\in both views:\\identity linked\\by position};
  \node[ttl] at (10.8,4.12) {(c) One scene: a square of 4 cells};
  \node[anchor=east, lab] at (12.8,3.62) {Video-2 actor:};
  \node[lab] at (13.4,3.62) {$b{=}0$}; \node[lab] at (14.5,3.62) {$b{=}1$};
  \node[anchor=east, lab] at (12.8,2.95) {Video-1 cues $a$};
  \node[anchor=east, lab] at (12.8,1.85) {Video-1 cues $\bar a$};
  \foreach \cx/\cy/\pc in {13.4/2.95/cyan, 14.5/2.95/yellow!85!black, 13.4/1.85/yellow!85!black, 14.5/1.85/cyan} {
    \draw[black!55, fill=white] (\cx-0.46,\cy-0.46) rectangle (\cx+0.46,\cy+0.46);
    \node[bigcube] (q) at (\cx,\cy+0.05) {}; \bigpatch{q}{\pc}
  }
  \draw[cvprblue, thick, rounded corners=2pt] (12.87,1.32) rectangle (15.03,2.38);
  \node[anchor=north, cvprblue, align=center] at (13.95,1.27) {pair: one row (same Video 1)};
  \draw[<->, thick, xtiming!80!black] (15.08,2.95) to[out=0, in=0, looseness=1.4] (15.08,1.85);
  \node[anchor=west, xtiming!80!black, align=left] at (15.5,2.4) {Video-1\\siblings};
  \foreach \cx in {16.15,16.95} {
    \draw[black!55, fill=white] (\cx-0.36,0.84) rectangle (\cx+0.36,1.56);
    \node[cube] (q) at (\cx,1.23) {}; \patch{q}{cyan}
  }
  \node[lab] at (16.15,1.76) {$V^2$}; \node[lab] at (16.95,1.76) {$V^{2\prime}$};
  \node[anchor=north, align=center, lab] at (16.55,0.78) {nuisance pair:\\same actor,\\Video 2 re-timed};
\end{tikzpicture}

%% file: figures/sigma_leak.tex
\newlength{\slw}\setlength{\slw}{0.088\textwidth}%
\begin{figure*}[t]
\centering
\begin{minipage}[b]{0.012\textwidth}\centering
  \rotatebox{90}{\scriptsize\makebox[\slw]{sibling}\makebox[\slw]{cell}}\\[-1pt]
  {\scriptsize\strut}\end{minipage}\hspace{2pt}%
\begin{minipage}[b]{\slw}\centering
  \IfFileExists{figures/leak_vis/panel_clean.png}{\includegraphics[width=\linewidth]{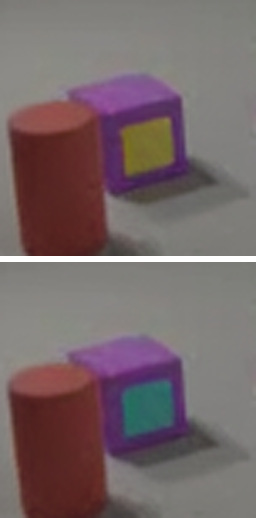}}%
  {\fcolorbox{gray!50}{gray!15}{\rule{0pt}{1.9\linewidth}\rule{0.92\linewidth}{0pt}}}\\[-1pt]
  {\scriptsize clean $z$}\end{minipage}\hspace{1pt}%
\begin{minipage}[b]{\slw}\centering
  \IfFileExists{figures/leak_vis/panel_s0p5.png}{\includegraphics[width=\linewidth]{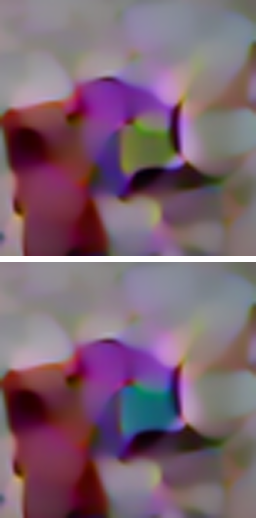}}%
  {\fcolorbox{gray!50}{gray!15}{\rule{0pt}{1.9\linewidth}\rule{0.92\linewidth}{0pt}}}\\[-1pt]
  {\scriptsize $\sigma{=}0.5$}\end{minipage}\hspace{1pt}%
\begin{minipage}[b]{\slw}\centering
  \IfFileExists{figures/leak_vis/panel_s1.png}{\includegraphics[width=\linewidth]{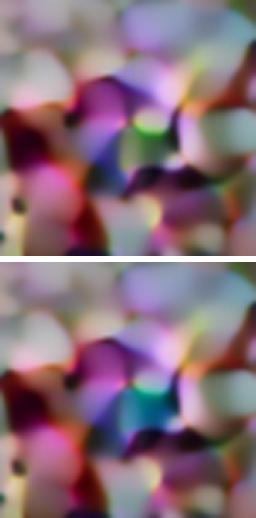}}%
  {\fcolorbox{gray!50}{gray!15}{\rule{0pt}{1.9\linewidth}\rule{0.92\linewidth}{0pt}}}\\[-1pt]
  {\scriptsize $\sigma{=}1$}\end{minipage}\hspace{1pt}%
\begin{minipage}[b]{\slw}\centering
  \IfFileExists{figures/leak_vis/panel_s3.png}{\includegraphics[width=\linewidth]{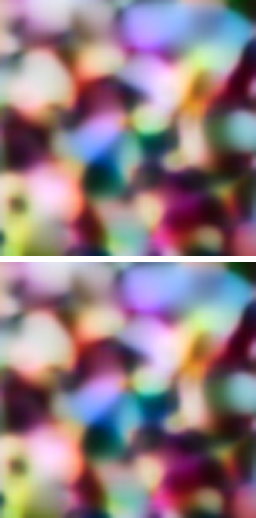}}%
  {\fcolorbox{gray!50}{gray!15}{\rule{0pt}{1.9\linewidth}\rule{0.92\linewidth}{0pt}}}\\[-1pt]
  {\scriptsize $\sigma{=}3$}\end{minipage}\hspace{1pt}%
\begin{minipage}[b]{\slw}\centering
  \IfFileExists{figures/leak_vis/panel_s10.png}{\includegraphics[width=\linewidth]{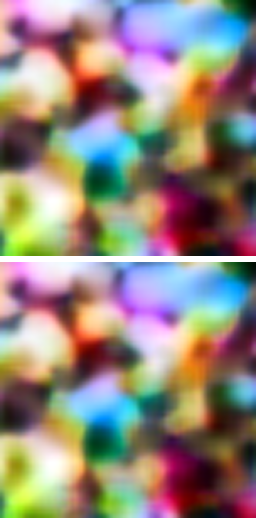}}%
  {\fcolorbox{gray!50}{gray!15}{\rule{0pt}{1.9\linewidth}\rule{0.92\linewidth}{0pt}}}\\[-1pt]
  {\scriptsize $\sigma{=}10$}\end{minipage}\hfill
\begin{minipage}[b]{0.5\textwidth}\centering
\begin{tikzpicture}
\begin{axis}[xmode=log, log basis x=10, width=\linewidth, height=4.1cm, xmin=0.3, xmax=300, ymin=0, ymax=100,
  xtick={0.5,1,2,5,10,20,50,100,200}, xticklabels={0.5,1,2,5,10,20,50,100,200},
  ytick={0,25,50,75,100}, yticklabel={\pgfmathprintnumber{\tick}\%},
  label style={font=\footnotesize}, tick label style={font=\scriptsize},
  axis line style={gray!60}, tick style={gray!60}, ymajorgrids, grid style={gray!20},
  axis x line*=bottom, axis y line*=left, clip=false,
  xlabel={noise level $\sigma$ of the noisy input}, ylabel={cells following Video~1},
  legend style={font=\scriptsize, draw=none, fill=none, row sep=-2pt, inner sep=1pt, at={(0.0,1.0)}, anchor=north west},
  legend cell align=left]
\addplot[domain=-1.204:5.704, samples=150, draw=none, fill=black!9, forget plot] ({exp(x)}, {45*exp(-((x-1.19895)^2)/2)}) \closedcycle;
\node[font=\scriptsize, text=black!55, anchor=south west, inner sep=1pt] at (axis cs:0.32,1) {training $\sigma$};
\draw[densely dashed, black!60] (axis cs:158,0) -- (axis cs:158,100) node[above, font=\scriptsize, text=black!70, inner sep=1pt] {$\sigma^\ast$ (576\,px)};
\addplot[color=xord!70, dashed, line width=0.8pt, mark=o, mark size=1.2pt] coordinates {(2,0) (3.3,0) (5,4.8) (8,14.3) (13,26.2) (20,40.5) (40,50) (80,54.8)};
\addlegendentry{released AnyView}
\addplot[color=black!35, line width=0.9pt, mark=*, mark size=1.2pt] coordinates {(2,0) (3.3,2.4) (5,14.3) (8,38.1) (13,59.5) (20,73.8) (40,83.3) (80,83.3)};
\addlegendentry{Two-video, 5k}
\addplot[color=black!65, line width=0.9pt, mark=*, mark size=1.2pt] coordinates {(2,0) (3.3,9.5) (5,33.3) (8,69.0) (13,83.3) (20,92.9) (40,100) (80,88.1)};
\addlegendentry{long, 10k}
\addplot[color=black, line width=1.1pt, mark=square*, mark size=1.3pt] coordinates {(2,0) (3.3,21.4) (5,45.2) (8,78.6) (13,90.5) (20,97.6) (40,100) (80,100)};
\addlegendentry{long, 17k}
\end{axis}
\end{tikzpicture}
\end{minipage}
\caption{\textbf{Noisy-input copying.} Left: cell and sibling with the same noise $\epsilon$, decoded at 256 pixels (actor zooms). Right: predictions following Video~1 despite the sibling's patch in the noisy input (42 training cells, 576 pixels). Released AnyView has no Video-1 input: a chance reference. Shading: main 95\% of training draws. Dashed $\sigma^\ast\approx160$ ($\approx54$ at 256 pixels): noise where an ideal observer knowing both targets falls to 60\% accuracy.}
\label{fig:leak}
\end{figure*}

%% file: figures/random_strip_fresh.tex
\setlength{\tabcolsep}{0.5pt}
\newcommand{\rk}[1]{{\scriptsize #1}}
\begin{tabular}{@{}c@{\hspace{2pt}}ccccc@{\hspace{5pt}}ccccc@{}}
  & \multicolumn{5}{c}{\textbf{Branch A}} & \multicolumn{5}{c}{\textbf{Branch B}} \\[-1pt]
  & {\scriptsize truth} & {\scriptsize LR$\times$2} & {\scriptsize Swap} & {\scriptsize Min-SNR} & {\scriptsize Swap + MS} & {\scriptsize truth} & {\scriptsize LR$\times$2} & {\scriptsize Swap} & {\scriptsize Min-SNR} & {\scriptsize Swap + MS} \\[1pt]
  \raisebox{.4\height}{1} & \includegraphics[width=.091\textwidth]{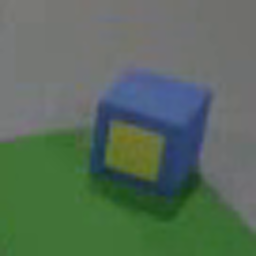} & \includegraphics[width=.091\textwidth]{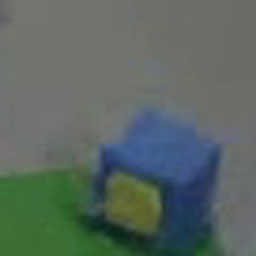} & \includegraphics[width=.091\textwidth]{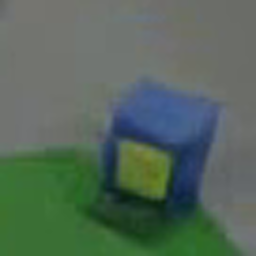} & \includegraphics[width=.091\textwidth]{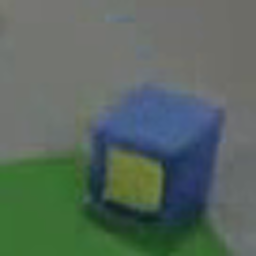} & \includegraphics[width=.091\textwidth]{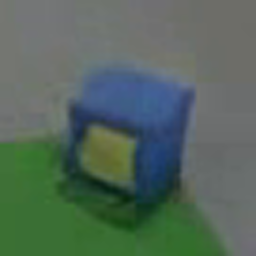} & \includegraphics[width=.091\textwidth]{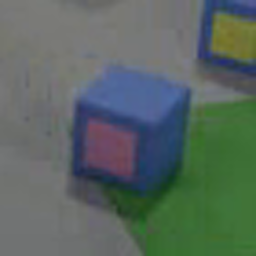} & \includegraphics[width=.091\textwidth]{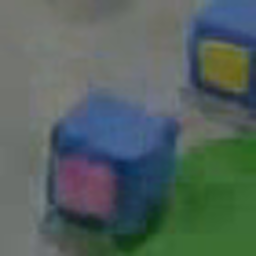} & \includegraphics[width=.091\textwidth]{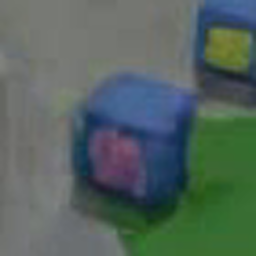} & \includegraphics[width=.091\textwidth]{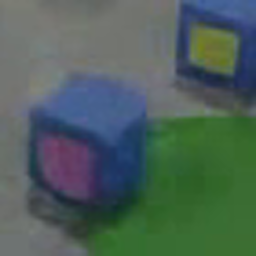} & \includegraphics[width=.091\textwidth]{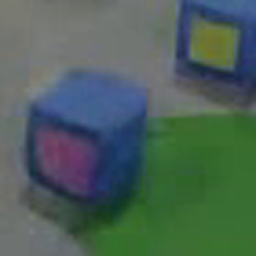} \\[-2pt]
  &  & \rk{\checkmark} & \rk{\checkmark} & \rk{\checkmark} & \rk{\checkmark} &  & \rk{\checkmark} & \rk{\checkmark} & \rk{\checkmark} & \rk{\checkmark} \\[1pt]
  \raisebox{.4\height}{2} & \includegraphics[width=.091\textwidth]{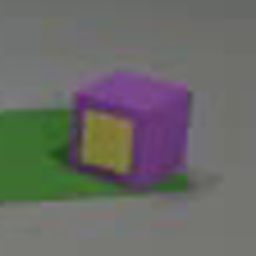} & \includegraphics[width=.091\textwidth]{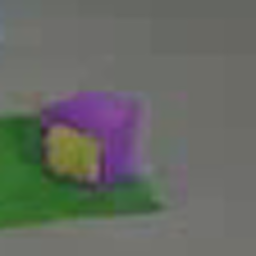} & \includegraphics[width=.091\textwidth]{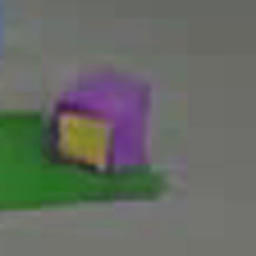} & \includegraphics[width=.091\textwidth]{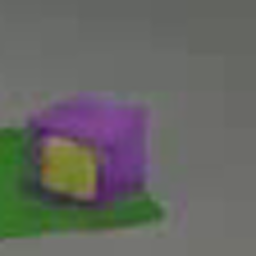} & \includegraphics[width=.091\textwidth]{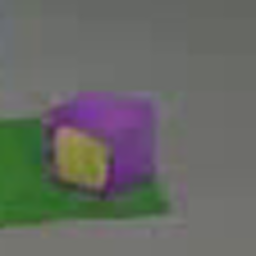} & \includegraphics[width=.091\textwidth]{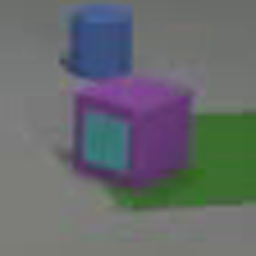} & \includegraphics[width=.091\textwidth]{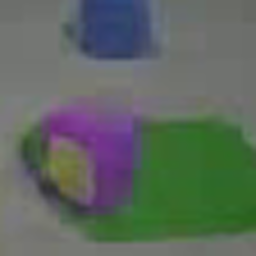} & \includegraphics[width=.091\textwidth]{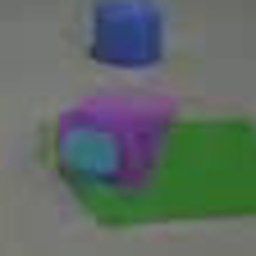} & \includegraphics[width=.091\textwidth]{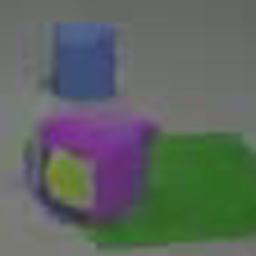} & \includegraphics[width=.091\textwidth]{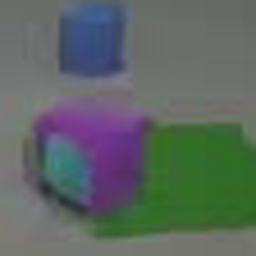} \\[-2pt]
  &  & \rk{\checkmark} & \rk{\checkmark} & \rk{\checkmark} & \rk{\checkmark} &  & \rk{--} & \rk{\checkmark} & \rk{$\times$} & \rk{\checkmark} \\[1pt]
  \raisebox{.4\height}{3} & \includegraphics[width=.091\textwidth]{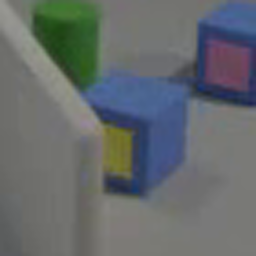} & \includegraphics[width=.091\textwidth]{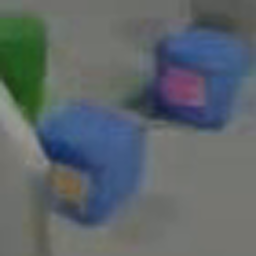} & \includegraphics[width=.091\textwidth]{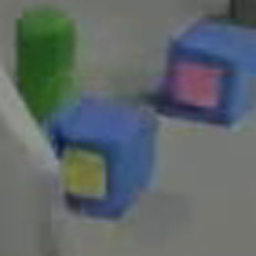} & \includegraphics[width=.091\textwidth]{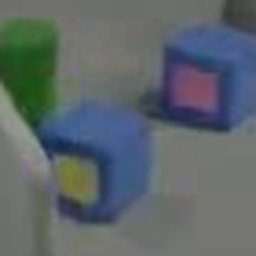} & \includegraphics[width=.091\textwidth]{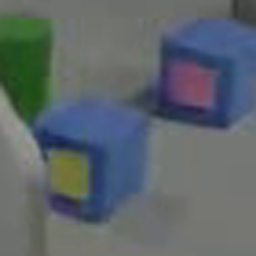} & \includegraphics[width=.091\textwidth]{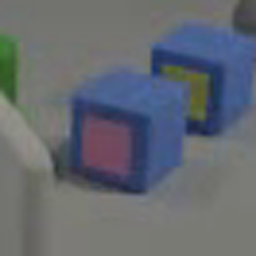} & \includegraphics[width=.091\textwidth]{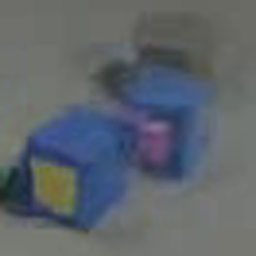} & \includegraphics[width=.091\textwidth]{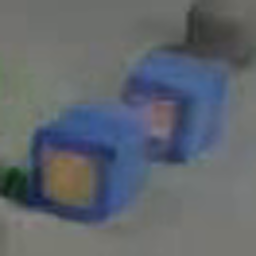} & \includegraphics[width=.091\textwidth]{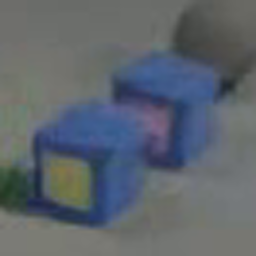} & \includegraphics[width=.091\textwidth]{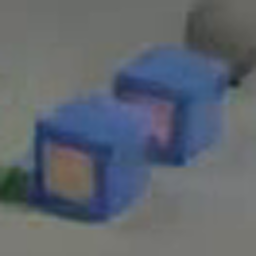} \\[-2pt]
  &  & \rk{\checkmark} & \rk{\checkmark} & \rk{\checkmark} & \rk{\checkmark} &  & \rk{$\times$} & \rk{$\times$} & \rk{--} & \rk{$\times$} \\[1pt]
\end{tabular}

%% file: figures/forest.tex
\begin{tikzpicture}[font=\scriptsize]
  \def\xa{3.3}\def\xs{0.12}\def\vmin{-6}\def\vmax{34}\def\dy{0.39}\def\yb{11.5}
  \newcommand{\px}[1]{\xa+\xs*((#1)-(\vmin))}
  \newcommand{\fg}[2]{\node[anchor=base west, inner sep=0, font=\scriptsize\bfseries] at (0,{-\dy*#1-0.05}) {#2};}
  \newcommand{\fr}[6]{%
    \node[anchor=base west, inner sep=0] at (0.15,{-\dy*#1-0.05}) {#2};
    \draw[black!75, line width=0.9pt] ({\px{#4}},{-\dy*#1}) -- ({\px{#5}},{-\dy*#1});
    \node[circle, inner sep=0, minimum size=4.4pt, draw=black, fill=#6, line width=0.8pt] at ({\px{#3}},{-\dy*#1}) {};}
  \node[circle, inner sep=0, minimum size=4.4pt, draw=black, fill=black, line width=0.8pt] at ({\px{2}},0) {};
  \node[anchor=base west, inner sep=0] at ({\px{3.2}},-0.05) {primary};
  \node[circle, inner sep=0, minimum size=4.4pt, draw=black, fill=white, line width=0.8pt] at ({\px{12}},0) {};
  \node[anchor=base west, inner sep=0] at ({\px{13.2}},-0.05) {secondary};
  \fill[gray!9] (-0.05,{-\dy*4.75}) rectangle ({\px{\vmax}},{-\dy*7.65});
  \foreach \v in {10,20,30} \draw[gray!25] ({\px{\v}},{-\dy*1.2}) -- ({\px{\v}},{-\dy*\yb});
  \draw[gray!65, dashed] ({\px{0}},{-\dy*1.2}) -- ({\px{0}},{-\dy*\yb});
  \draw[gray!60] ({\px{\vmin}},{-\dy*\yb}) -- ({\px{\vmax}},{-\dy*\yb});
  \foreach \v in {0,10,20,30} {
    \draw[gray!60] ({\px{\v}},{-\dy*\yb}) -- ++(0,-0.06);
    \node[anchor=north, inner sep=1.5pt] at ({\px{\v}},{-\dy*\yb-0.06}) {$\v$};}
  \node[anchor=north, font=\footnotesize] at ({\px{14}},{-\dy*\yb-0.38}) {difference (points)};
  \fg{1.6}{Early checkpoint}
  \fr{2.6}{256 px, fresh split, 10k}{9.9}{2.9}{16.7}{black}
  \fr{3.6}{576 px, 10k}{22.9}{13.9}{32.6}{black}
  \fg{5.2}{Late checkpoint}
  \fr{6.2}{256 px, fresh split, 17.5k}{5.2}{1.4}{9.1}{white}
  \fr{7.2}{256 px, 40k schedule, 40k}{0.4}{-4.0}{4.8}{black}
  \fg{8.8}{Adding Min-SNR to noisy-input swap}
  \fr{9.8}{256 px, fresh split, 10k}{11.5}{5.3}{18.1}{black}
  \fr{10.8}{256 px, fresh split, 17.5k}{5.0}{1.6}{8.4}{black}
\end{tikzpicture}

%% file: figures/panoptic_split.tex
\begin{tikzpicture}
\begin{axis}[
  width=\linewidth, height=0.36\linewidth,
  ybar=0pt, bar width=22pt, bar shift=0pt,
  xmin=0.4, xmax=4.6, ymin=0, ymax=100, enlarge y limits={upper,value=0.02},
  xtick={1,2,3,4}, xticklabels={ordinary,swap,Min-SNR,swap + MS},
  ytick={0,25,50,75,100}, ylabel={\% (5k)},
  label style={font=\footnotesize}, tick label style={font=\scriptsize},
  x tick label style={font=\footnotesize, text height=1.5ex, text depth=.25ex},
  axis line style={gray!60}, tick style={gray!60}, ymajorgrids, grid style={gray!20},
  major x tick style={draw=none},
  axis x line*=bottom, axis y line*=left, clip=false,
  nodes near coords, nodes near coords style={font=\scriptsize, /pgf/number format/.cd, fixed, fixed zerofill, precision=1},
]
\addplot[fill=xord, draw=none] coordinates {(1,15.9)};
\addplot[fill=xswap!45, draw=none] coordinates {(2,25.4)};
\addplot[fill=xreweight, draw=none] coordinates {(3,35.4)};
\addplot[fill=xswap, draw=none] coordinates {(4,40.7)};
\end{axis}
\end{tikzpicture}

%% file: figures/panoptic_bars.tex
\begin{tikzpicture}
\begin{axis}[
  width=\linewidth, height=0.6\linewidth,
  ybar=1.5pt, bar width=10pt,
  xmin=0.45, xmax=3.55, ymin=0, ymax=100, enlarge y limits={upper,value=0.02},
  xtick={1,2,3}, xticklabels={correct binding,position-tolerant,natural},
  ytick={0,25,50,75,100}, ylabel={\% (5k updates)},
  label style={font=\footnotesize}, tick label style={font=\scriptsize},
  x tick label style={font=\footnotesize},
  axis line style={gray!60}, tick style={gray!60}, ymajorgrids, grid style={gray!20},
  major x tick style={draw=none},
  axis x line*=bottom, axis y line*=left, clip=false,
  nodes near coords, nodes near coords style={font=\scriptsize, rotate=90, anchor=west, inner sep=1.5pt, /pgf/number format/.cd, fixed, fixed zerofill, precision=1},
  legend style={font=\scriptsize, draw=none, fill=none, at={(0.5,1.03)}, anchor=south, column sep=4pt, cells={anchor=west}},
  legend columns=2, legend image code/.code={\draw[#1, draw=none] (0cm,-0.08cm) rectangle (0.24cm,0.12cm);},
]
\addplot[fill=xord, draw=none] coordinates {(1,23.0) (2,59.7) (3,56.1)};
\addlegendentry{ordinary}
\addplot[fill=xreweight, draw=none] coordinates {(1,44.6) (2,70.6) (3,74.7)};
\addlegendentry{Min-SNR}
\addplot[fill=xreweight!40, draw=none] coordinates {(1,60.5) (2,72.1) (3,74.4)};
\addlegendentry{Min-SNR$^\dagger$ (control)}
\addplot[fill=xswap, draw=none] coordinates {(1,70.7) (2,76.0) (3,79.1)};
\addlegendentry{swap + Min-SNR$^\dagger$}
\end{axis}
\end{tikzpicture}

%% file: figures/probe_style.tex
\pgfplotsset{probeaxis/.style={
  width=\linewidth, xmin=1.8, xmax=90, ymin=0, ymax=100,
  xtick={2,5,10,20,40,80}, xticklabels={2,5,10,20,40,80},
  yticklabel={\pgfmathprintnumber{\tick}\%},
  label style={font=\footnotesize}, tick label style={font=\scriptsize},
  axis line style={gray!60}, tick style={gray!60}, ymajorgrids, grid style={gray!20},
  axis x line*=bottom, axis y line*=left,
  title style={font=\footnotesize, at={(0,1)}, anchor=south west, yshift=-2pt, xshift=-14pt},
  legend style={font=\scriptsize, draw=none, fill=none, row sep=-2pt, inner sep=1pt}, legend cell align=left}}

%% file: figures/decision_probe.tex
\begin{tikzpicture}
\begin{axis}[xmode=log, log basis x=10, probeaxis, height=0.5\linewidth,
  ytick={0,50,100},
  title={Decision probe on training squares},
  xlabel={starting noise level $\sigma_s$}, ylabel={correct | answered},
  legend style={at={(0.02,0.97)}, anchor=north west},
  error bars/y dir=both, error bars/y explicit,
]
\addplot[color=black!35, line width=0.9pt, mark=*, mark size=1.2pt,
  error bars/error bar style={line width=0.6pt, black!35}, error bars/error mark options={rotate=90, mark size=1.5pt, black!35}]
  coordinates {(10,9.0) +=(0,9.2) -=(0,4.8)  (20,27.0) +=(0,12.0) -=(0,9.4)  (40,41.7) +=(0,12.6) -=(0,11.6)  (80,54.7) +=(0,11.6) -=(0,12.1)};
\addlegendentry{5k}
\addplot[color=black, line width=1.1pt, mark=square*, mark size=1.3pt,
  error bars/error bar style={line width=0.6pt, black}, error bars/error mark options={rotate=90, mark size=1.5pt, black}]
  coordinates {(10,53.2) +=(0,11.9) -=(0,12.2)  (20,83.6) +=(0,7.0) -=(0,10.7)  (40,90.5) +=(0,4.8) -=(0,8.8)  (80,95.7) +=(0,2.8) -=(0,7.7)};
\addlegendentry{long, 17k}
\end{axis}
\end{tikzpicture}

%% file: figures/effweight.tex
\begin{figure}[!htbp]
\centering
\begin{tikzpicture}
\begin{axis}[xmode=log, log basis x=10, width=\linewidth, height=4.4cm, ymin=0, xmin=0.05, xmax=300, ymax=0.75,
  xtick={0.1,1,10,100}, xticklabels={0.1,1,10,100},
  label style={font=\footnotesize}, tick label style={font=\scriptsize},
  axis line style={gray!60}, tick style={gray!60}, ymajorgrids, grid style={gray!20},
  axis x line*=bottom, axis y line*=left,
  xlabel={noise level $\sigma$}, ylabel={share of loss weight per unit $\ln\sigma$},
  legend style={font=\scriptsize, draw=none, fill=none, row sep=-2pt, inner sep=1pt, at={(1.0,1.0)}, anchor=north east},
  legend cell align=left]
\fill[black!7] (axis cs:0.05,0) rectangle (axis cs:2.7,0.75);
\node[font=\scriptsize, text=black!55, anchor=north west, inner sep=1pt] at (axis cs:0.055,0.74) {ordinary copies};
\draw[densely dashed, black!50] (axis cs:1,0) -- (axis cs:1,0.75);
\node[font=\scriptsize, text=black!55, anchor=north west, inner sep=1pt] at (axis cs:1.05,0.53) {swap acts $\rightarrow$};
\addplot[color=xord, line width=1.0pt] table[col sep=comma, x=sigma, y expr=\thisrow{ordinary}/1.000] {figures/data/effweight_crossswitch.csv};
\addlegendentry{ordinary}
\addplot[color=xlr, densely dashed, line width=0.8pt] table[col sep=comma, x=sigma, y expr=\thisrow{shift_x2}/0.712] {figures/data/effweight_crossswitch.csv};
\addlegendentry{$\sigma$ shift $\times2$}
\addplot[color=xlr, dotted, line width=0.9pt] table[col sep=comma, x=sigma, y expr=\thisrow{shift_x4}/0.638] {figures/data/effweight_crossswitch.csv};
\addlegendentry{$\sigma$ shift $\times4$}
\addplot[color=xord!60, dashdotted, line width=0.8pt] table[col sep=comma, x=sigma, y expr=\thisrow{pure_noise_5x}/0.919] {figures/data/effweight_crossswitch.csv};
\addlegendentry{$5\times$ pure noise}
\addplot[color=black, line width=1.1pt] table[col sep=comma, x=sigma, y expr=\thisrow{minsnr5}/0.262] {figures/data/effweight_crossswitch.csv};
\addlegendentry{Min-SNR}
\end{axis}
\end{tikzpicture}
\caption{\textbf{Where each arm puts its loss weight} (CrossSwitch, AnyView's EDM objective). Loss weight times training density per unit $\ln\sigma$, each arm normalized to its own total, since the optimizer is largely insensitive to the loss scale. Shaded: noise levels at which ordinary models copy (crossover $\sigma\approx2.7$ at 256 pixels after 10,000 updates). Min-SNR removes 74\% of EDM's weight, most of it above $\sigma=1$, so that 69\% of what remains lies below $\sigma=1$ (ordinary training: 37\%) and 95\% below the crossover (62\%); in the band $1\le\sigma<5$, where noisy-input swap acts, its share falls from 40 to 30\%. The noise shifts move weight the other way: 27 and 8\% below the crossover. Patch weighting has ordinary training's shape. The log-uniform tail of draws on $[200,10^5]$ lies mostly off the axis.}
\label{fig:effweight}
\end{figure}

%% file: figures/emergence.tex
\begin{tikzpicture}
\begin{axis}[
  width=0.95\linewidth, height=0.52\linewidth,
  xmin=1.5, xmax=18.6, ymin=0, ymax=90,
  xtick={2.5,5,7.5,10,12.5,15,18}, xticklabels={2.5,5,7.5,10,12.5,15,18},
  ytick={0,25,50,75},
  xlabel={training updates (thousands)}, ylabel={switch accuracy (\%)},
  label style={font=\small}, tick label style={font=\footnotesize},
  axis line style={gray!60}, tick style={gray!60},
  ymajorgrids, grid style={gray!15},
  axis x line*=bottom, axis y line*=left,
  clip=false,
  legend style={at={(0.02,0.98)}, anchor=north west, draw=none, fill=white, fill opacity=0.85, text opacity=1,
                font=\small, row sep=-1pt, inner sep=2pt},
  legend cell align=left,
]
\fill[black!7] (axis cs:1.5,13) rectangle (axis cs:18.6,19);
\node[anchor=west, font=\footnotesize, text=black!55] at (axis cs:11.8,16) {guessing};
\addplot[color=xlayout, line width=1.3pt, mark=*, mark size=2pt, mark options={draw=white, line width=0.5pt}]
  coordinates {(2.5,30) (5,52) (7.5,56) (10,48) (12.5,78) (15,70) (17,78) (18,78)};
\addlegendentry{layout}
\addplot[color=xtiming, line width=1.3pt, mark=square*, mark size=1.9pt, mark options={draw=white, line width=0.5pt}]
  coordinates {(2.5,10) (5,20) (7.5,52) (10,28) (12.5,40) (15,64) (17,66) (18,66)};
\addlegendentry{\synclane{}}
\end{axis}
\end{tikzpicture}

%% file: figures/random_grid.tex
\setlength{\tabcolsep}{0.6pt}
\begin{tabular}{@{}c@{\hspace{3pt}}ccc@{\hspace{6pt}}ccc@{}}
  & \multicolumn{3}{c}{\textbf{Branch A}} & \multicolumn{3}{c}{\textbf{Branch B}} \\[-1pt]
  & {\scriptsize truth} & {\scriptsize ordinary} & {\scriptsize noisy-input swap} & {\scriptsize truth} & {\scriptsize ordinary} & {\scriptsize noisy-input swap} \\[1pt]
  1 & \raisebox{-.5\height}{\includegraphics[width=.118\textwidth]{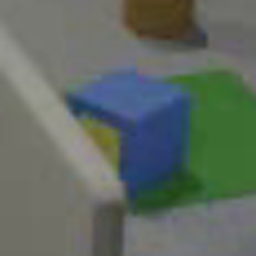}} & \raisebox{-.5\height}{\includegraphics[width=.118\textwidth]{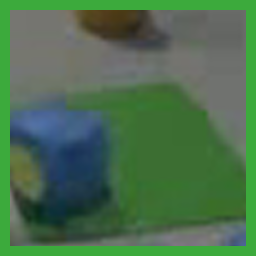}} & \raisebox{-.5\height}{\includegraphics[width=.118\textwidth]{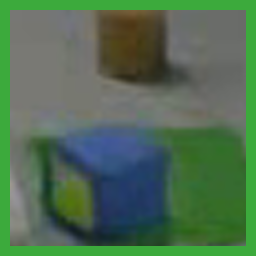}} & \raisebox{-.5\height}{\includegraphics[width=.118\textwidth]{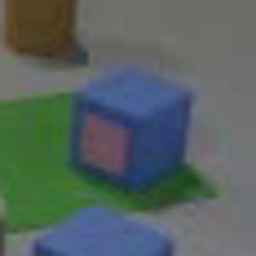}} & \raisebox{-.5\height}{\includegraphics[width=.118\textwidth]{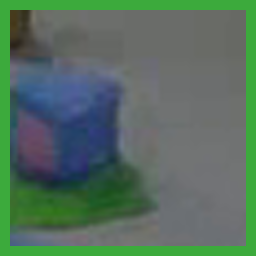}} & \raisebox{-.5\height}{\includegraphics[width=.118\textwidth]{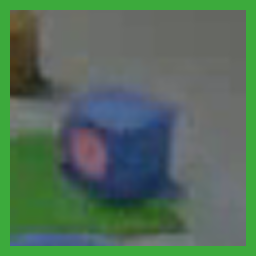}} \\[1pt]
  2 & \raisebox{-.5\height}{\includegraphics[width=.118\textwidth]{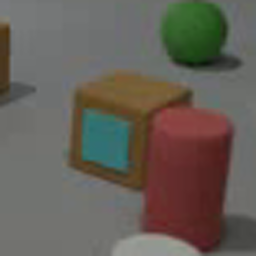}} & \raisebox{-.5\height}{\includegraphics[width=.118\textwidth]{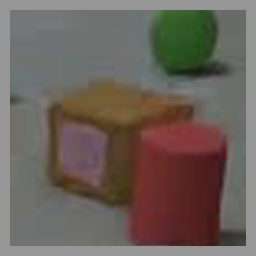}} & \raisebox{-.5\height}{\includegraphics[width=.118\textwidth]{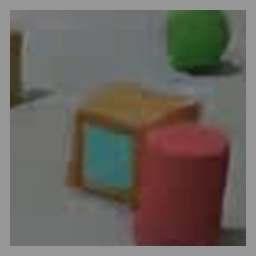}} & \raisebox{-.5\height}{\includegraphics[width=.118\textwidth]{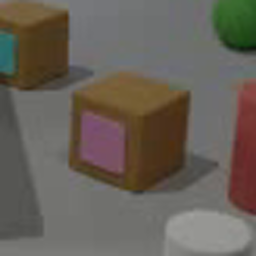}} & \raisebox{-.5\height}{\includegraphics[width=.118\textwidth]{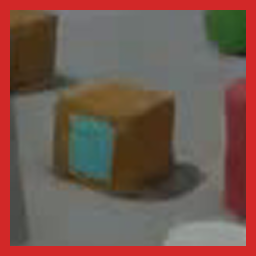}} & \raisebox{-.5\height}{\includegraphics[width=.118\textwidth]{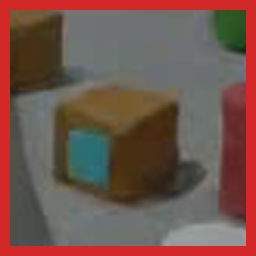}} \\[1pt]
  3 & \raisebox{-.5\height}{\includegraphics[width=.118\textwidth]{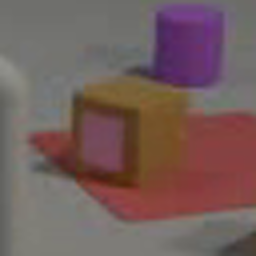}} & \raisebox{-.5\height}{\includegraphics[width=.118\textwidth]{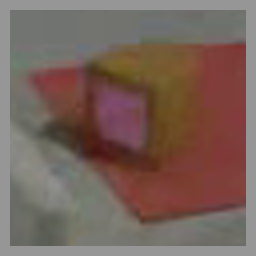}} & \raisebox{-.5\height}{\includegraphics[width=.118\textwidth]{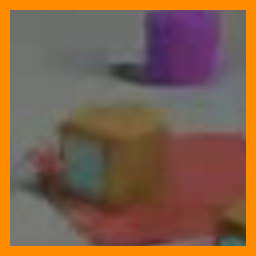}} & \raisebox{-.5\height}{\includegraphics[width=.118\textwidth]{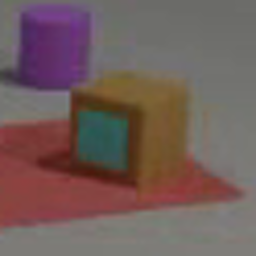}} & \raisebox{-.5\height}{\includegraphics[width=.118\textwidth]{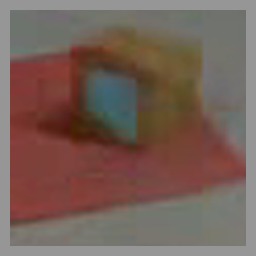}} & \raisebox{-.5\height}{\includegraphics[width=.118\textwidth]{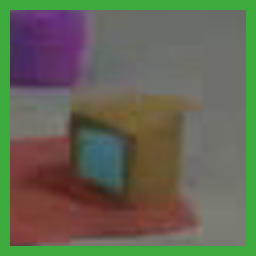}} \\[1pt]
  4 & \raisebox{-.5\height}{\includegraphics[width=.118\textwidth]{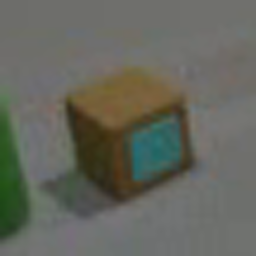}} & \raisebox{-.5\height}{\includegraphics[width=.118\textwidth]{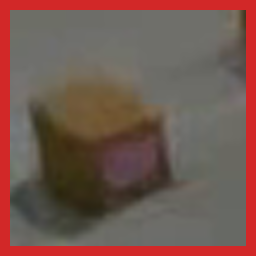}} & \raisebox{-.5\height}{\includegraphics[width=.118\textwidth]{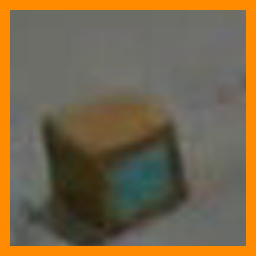}} & \raisebox{-.5\height}{\includegraphics[width=.118\textwidth]{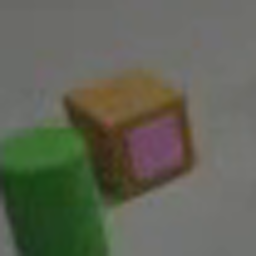}} & \raisebox{-.5\height}{\includegraphics[width=.118\textwidth]{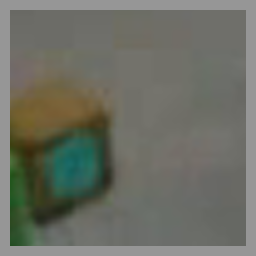}} & \raisebox{-.5\height}{\includegraphics[width=.118\textwidth]{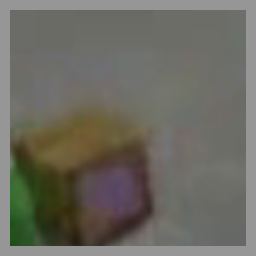}} \\[1pt]
\end{tabular}